%% file: paper.tex
\pdfoutput=1
\documentclass[11pt,letterpaper]{article}
\usepackage[T1]{fontenc}
\usepackage[utf8]{inputenc}
\usepackage{lmodern}
\usepackage[margin=1in]{geometry}
\usepackage{microtype}
\usepackage{graphicx}
\usepackage{booktabs}
\usepackage{array}
\usepackage{tabularx}
\usepackage{adjustbox}
\usepackage{xurl}
\usepackage{caption}
\usepackage{subcaption}
\usepackage{xcolor}
\usepackage{placeins}
\usepackage{float}
\usepackage{enumitem}
\usepackage{amsmath,amssymb,amsthm}
\usepackage{dsfont}
\usepackage{bm}
\usepackage{tikz}
\usetikzlibrary{positioning,arrows.meta,calc}
\usepackage[backend=biber,style=authoryear,maxcitenames=2,uniquename=false,dashed=false]{biblatex}
\appto{\bibsetup}{\raggedright}
\usepackage[colorlinks=true,linkcolor=blue!50!black,citecolor=blue!50!black,urlcolor=blue!50!black]{hyperref}
\setlist{itemsep=2pt,topsep=3pt}

\newtheorem{proposition}{Proposition}
\newtheorem{assumption}{Assumption}
\newcommand{\E}{\mathbb{E}}
\newcommand{\Var}{\mathrm{Var}}
\newcommand{\ind}[1]{\mathds{1}[#1]}
\newcommand{\hT}{\widehat{T}}
\graphicspath{{figures/}}
\input{numbers_csai.tex}


\input{latex_report/numbers_validation.tex}

\newcommand{\csAdjudicatorSentence}{The three readers adjudicated them together, by discussion, and were shown nothing beyond the workbook: no machine label and none of this paper's estimates. Every one of the decisions was unanimous, and neither the codebook nor any of its rules changed during the adjudication.}

\title{\textbf{Counting Under Disagreement:}\\[3pt]\large Dollar-Weighted Measurement, Selective Machine Labelling and\\Budgeted Human Review from Disagreeing AI Readers of Administrative Text}
\author{George Koomullil\\\small Ascendr, PBC\\\small \texttt{george@ascendr.ai}}
\date{}

\begin{document}
\maketitle

\begin{abstract}
\noindent
We study AI-assisted measurement when one text label applies to many administrative records, the target is a
monetary total, and the human readings that anchor the machines disagree. The testbed is \csNTexts{} grant-purpose
texts from US tax filings (\csNRows{} records, \$\csFrameB{}B), labelled by a rule-based classifier and three
language models, with a designed sample of \csNSample{} texts read independently by three people under one
codebook, who then adjudicated the \csNSplit{} texts they split on. We evaluate calibrated prediction, design-based correction, selective machine labelling and adaptive
human review on it. We state exact conditions for a fixed-predictor difference estimator and for a cross-fitted
predictor whose folds are assigned to the whole frame independently of the sample, under Poisson sampling, and assess the
implemented fixed-size procedures by simulation: \csNPredictors{} predictors whose own totals span
\$\csPluginAllMinB{}--\csPluginAllMaxB{}B agree to within \$\csPpiAllSpreadB{}B once rectified, and
stratification on the machines' votes leaves the predictor little scope to improve the national total. Two rules
that accept machine labels at a stated dollar-weighted error, weighted conformal p-values under a dollar measure
and Learn-then-Test on the design estimate of the dollar error, are evaluated empirically on nested held-out
folds and on re-draws of the design with synthetic truths; their error control in deployment is not established,
and the count-based rule accepts no text under a dollar-drawn sample. In a replay that counts every reference
label a policy queries, its pilot included, a two-phase review policy whose calibrator is fitted on its own first
phase reduces the RMSE of the total by \csAdTwoPhaseVsClassicHundredPct{} at 100 queries relative to
dollar-proportional sampling; a query reveals the recorded reference label, and the readings and the adjudication
that produced it are not costed. The reduction depends on the penalty of that calibrator: it is
\csRidgeTunedHundredPct{} when the policy chooses the penalty on its own first-phase labels
(\csRerunRidgeTunedHundredMin{}--\csRerunRidgeTunedHundredMax{} with an individual reader or the readers' majority
as the reference), \csRidgeModerateHundredMinPct{}--\csRidgeModerateHundredMaxPct{} over moderate fixed
penalties, and close to nothing under the strongest one tried. The pure Neyman allocation fails with an uncalibrated predictor and under-covers without a floor on the inclusion
probabilities. Independent readers shift the estimated total by \$\csAltReadersSpreadPpiB{}B,
\csAltSpreadOverWidth{} times the sampling interval; the adjudicated reading, the reference of this version, puts
it at \$\csPpiB{}B, \$\csAdjMinusOnePpiB{}B above the first reader's labels alone; and institution-level
intervals under-cover at every size. The reference-dependent results are re-run under each reader; the simulations
on synthetic truths are held fixed. The results identify both
the benefits and the limits of using learned predictions to decide what a person should verify.
\end{abstract}


\clearpage

\input{sections/intro}
\input{sections/related}
\input{sections/problem}
\input{sections/testbed}
\input{sections/aggregation}
\input{sections/ppi}
\input{sections/selective}
\input{sections/acquisition}
\input{sections/invariance}
\input{sections/downstream}
\input{sections/discussion}
\input{sections/release}

\clearpage
\printbibliography

\clearpage
\appendix
\input{sections/appendix}

\end{document}

%% file: numbers_csai.tex
\newcommand{\csAblAtomsDollarHi}{+0.005}
\newcommand{\csAblAtomsDollarLL}{0.081}
\newcommand{\csAblAtomsDollarLo}{-0.006}
\newcommand{\csAblDsLL}{0.155}

\newcommand{\csAblFullMaxDollar}{+0.0015}
\newcommand{\csAblFullMinDollar}{+0.0004}

\newcommand{\csAblFullSonnetLL}{0.172}
\newcommand{\csAblFullWorst}{GPT-5~mini}

\newcommand{\csAblOtherBelow}{0.002}

\newcommand{\csAblTextDollarHi}{+0.011}
\newcommand{\csAblTextDollarLL}{0.087}
\newcommand{\csAblTextDollarLo}{-0.000}

\newcommand{\csAcqReps}{1,000}

\newcommand{\csAdBatchAvgMixedFiftyCov}{0.88}

\newcommand{\csAdBatchAvgMixedHundredRmse}{1.25}

\newcommand{\csAdBatchAvgVsClassicHundred}{1.89}

\newcommand{\csAdBatchAvgVsClassicTwoHundred}{2.04}

\newcommand{\csAdBatchLastMixedHundredRmse}{1.96}

\newcommand{\csAdBestPhiFifty}{50\%}
\newcommand{\csAdBestPhiHundred}{25\%}

\newcommand{\csAdBestPhiTwoHundred}{25\%}

\newcommand{\csAdClassicHundredRmse}{2.35}

\newcommand{\csAdClassicRatioHundred}{2.7}

\newcommand{\csAdClassicRatioTwoHundred}{2.4}

\newcommand{\csAdClassicTwoHundredRmse}{1.38}
\newcommand{\csAdCostBudgetHundred}{705}

\newcommand{\csAdCostMixedCostHundredRmse}{0.86}
\newcommand{\csAdCostMixedCostHundredTexts}{140}

\newcommand{\csAdCostMixedCostTwoHundredRmse}{0.48}

\newcommand{\csAdCostMixedHundredRmse}{0.96}
\newcommand{\csAdCostMixedHundredTexts}{94}

\newcommand{\csAdCostMixedTwoHundredRmse}{0.57}

\newcommand{\csAdCostPpsHundredRmse}{1.26}

\newcommand{\csAdCovMcSe}{0.014}
\newcommand{\csAdFeatures}{58}

\newcommand{\csAdFiftyMixedHundredRmse}{1.44}

\newcommand{\csAdK}{4}

\newcommand{\csAdMultiHundredMmMaxYearRel}{9.9\%}
\newcommand{\csAdMultiHundredMmTotalRel}{6.0\%}

\newcommand{\csAdMultiHundredTotMaxRatio}{8.14}
\newcommand{\csAdMultiHundredTotMaxYearRel}{10.9\%}
\newcommand{\csAdMultiHundredTotTotalRel}{5.6\%}
\newcommand{\csAdMultiHundredTotWorstYear}{2024}

\newcommand{\csAdMultiTwoHundredMmMaxYearRel}{6.0\%}
\newcommand{\csAdMultiTwoHundredMmTotalRel}{3.5\%}

\newcommand{\csAdMultiTwoHundredNote}{, where the worst year does not gain}

\newcommand{\csAdMultiTwoHundredTotMaxYearRel}{5.9\%}
\newcommand{\csAdMultiTwoHundredTotTotalRel}{3.1\%}

\newcommand{\csAdNoPilotRatioHundred}{1.6}

\newcommand{\csAdNoPilotRatioTwoHundred}{1.2}

\newcommand{\csAdRefMixedHundredRmse}{0.87}

\newcommand{\csAdRefMixedTwoHundredRmse}{0.56}
\newcommand{\csAdReps}{300}

\newcommand{\csAdSeventyFiveMixedHundredRmse}{1.79}

\newcommand{\csAdTwentyFiveMixedHundredCov}{0.94}

\newcommand{\csAdTwentyFiveMixedHundredRmse}{1.36}

\newcommand{\csAdTwentyFiveMixedTwoHundredRmse}{0.68}

\newcommand{\csAdTwoPhaseVsClassicHundred}{1.73}

\newcommand{\csAdTwoPhaseVsClassicHundredPct}{42\%}
\newcommand{\csAdTwoPhaseVsClassicHundredSe}{0.10}

\newcommand{\csAdTwoPhaseVsClassicTwoHundred}{2.03}
\newcommand{\csAdTwoPhaseVsClassicTwoHundredPct}{51\%}

\newcommand{\csAdWordsMean}{10.0}
\newcommand{\csAdWordsMedian}{6}

\newcommand{\csAdjMajSame}{99}
\newcommand{\csAdjMajToNotR}{7}

\newcommand{\csAdjMajToR}{46}

\newcommand{\csAdjMinusOnePpiB}{2.20}

\newcommand{\csAdjOneNToR}{22}

\newcommand{\csAdjOneSaysR}{84}
\newcommand{\csAdjOneSaysRAdjR}{46}
\newcommand{\csAdjOneUToR}{34}

\newcommand{\csAdjSameOneAllTexts}{90\%}

\newcommand{\csAdjSameOneN}{85}

\newcommand{\csAdjSameThreeAllTexts}{93\%}

\newcommand{\csAdjSameThreeN}{109}

\newcommand{\csAdjSameTwoAllTexts}{85\%}

\newcommand{\csAdjSameTwoN}{57}

\newcommand{\csAdjSplitMajRN}{68}
\newcommand{\csAdjSplitMajRWd}{47\%}

\newcommand{\csAdjSplitRWd}{67\%}

\newcommand{\csAdjStatusN}{17}
\newcommand{\csAdjStatusR}{107}

\newcommand{\csAdjStatusU}{28}
\newcommand{\csAdjTexts}{152}

\newcommand{\csAdjTwoSayR}{68}
\newcommand{\csAdjTwoSayRAdjR}{61}

\newcommand{\csAgAllKappaStatus}{0.39}

\newcommand{\csAgBinaryTextsMax}{87\%}
\newcommand{\csAgBinaryTextsMin}{82\%}

\newcommand{\csAgKappaBinaryMax}{0.64}
\newcommand{\csAgKappaBinaryMin}{0.51}
\newcommand{\csAgKappaStatusMax}{0.48}
\newcommand{\csAgKappaStatusMin}{0.34}

\newcommand{\csAgTwoThreeBinaryTexts}{82\%}
\newcommand{\csAgTwoThreeBinaryW}{94\%}

\newcommand{\csAllRPpiB}{7.30}

\newcommand{\csAltReadersSpreadPpiB}{3.93}

\newcommand{\csAltSpreadOverWidth}{2.7}

\newcommand{\csAnyRPpiB}{16.58}

\newcommand{\csBBoot}{1,000}

\newcommand{\csBiasARefit}{+0.11\%}

\newcommand{\csBiasBRefit}{+0.01\%}

\newcommand{\csBiasCRefit}{+0.10\%}

\newcommand{\csBiasRefitMaxRatio}{0.8}
\newcommand{\csCfinal}{0.03}
\newcommand{\csClassicB}{13.45}

\newcommand{\csClassicHiB}{14.2}
\newcommand{\csClassicLoB}{12.7}

\newcommand{\csConMajB}{1.90}

\newcommand{\csConMajHiB}{+2.71}
\newcommand{\csConMajLoB}{+1.11}

\newcommand{\csConOneB}{2.16}
\newcommand{\csConOneGain}{56}
\newcommand{\csConOneGainB}{2.74}
\newcommand{\csConOneGainN}{22}
\newcommand{\csConOneGainU}{34}
\newcommand{\csConOneHiB}{+2.99}
\newcommand{\csConOneLoB}{+1.34}
\newcommand{\csConOneLoss}{11}
\newcommand{\csConOneLossB}{0.59}

\newcommand{\csConRoundingNote}{1.04 $-$ 1.59}
\newcommand{\csConThreeAbsB}{0.54}

\newcommand{\csConThreeGain}{22}

\newcommand{\csConThreeHiB}{+0.33}
\newcommand{\csConThreeLoB}{-1.43}
\newcommand{\csConThreeLoss}{21}

\newcommand{\csConThreeSign}{$-$}

\newcommand{\csConTwoB}{3.39}

\newcommand{\csConTwoHiB}{+4.53}
\newcommand{\csConTwoLoB}{+2.25}

\newcommand{\csContrastFlashB}{37.9}
\newcommand{\csContrastFlashN}{84}
\newcommand{\csContrastFlashRows}{393,760}
\newcommand{\csContrastFlashSame}{98\%}
\newcommand{\csContrastFlashSameR}{100\%}

\newcommand{\csContrastMiniSame}{98\%}

\newcommand{\csCovAClassic}{0.980}

\newcommand{\csCovARefit}{0.973}

\newcommand{\csCovBRefit}{0.945}

\newcommand{\csCovCRefit}{0.980}

\newcommand{\csCovCalActiveFiveHundred}{0.90}

\newcommand{\csCovCalActiveHundred}{0.95}

\newcommand{\csCovDsActiveHundred}{0.90}
\newcommand{\csCovDsActiveMax}{0.92}
\newcommand{\csCovDsActiveMin}{0.46}

\newcommand{\csCovMCSE}{0.011}
\newcommand{\csDelta}{0.1}
\newcommand{\csDepAdjCorrTextsMax}{0.45}
\newcommand{\csDepAdjCorrTextsMin}{0.38}

\newcommand{\csDepAdjCorrWdMax}{0.66}
\newcommand{\csDepAdjCorrWdMin}{0.57}

\newcommand{\csDepOneCorrWMax}{0.90}
\newcommand{\csDepOneCorrWMin}{0.81}
\newcommand{\csDepOneCorrWNoTopMax}{0.77}
\newcommand{\csDepOneCorrWNoTopMin}{0.54}

\newcommand{\csDisagreeAUCHuman}{0.58}
\newcommand{\csDisagreeAUCLLM}{0.64}
\newcommand{\csDriftCalAbsMax}{0.7}
\newcommand{\csDriftCalMax}{+0.7}
\newcommand{\csDriftCalMin}{-0.1}

\newcommand{\csDriftRulesTwenty}{-0.9}
\newcommand{\csDriftRulesTwentyFour}{-0.2}
\newcommand{\csDriftSingleAbsMax}{1.9}
\newcommand{\csDriftSingleStepMax}{1.3}

\newcommand{\csDryVersionChangedMax}{10\%}
\newcommand{\csDryVersionChangedMin}{5\%}

\newcommand{\csDsAUCdollar}{0.990}

\newcommand{\csDsGoldFreeB}{14.0}

\newcommand{\csDsGoldFreeVsRef}{inside}
\newcommand{\csDsHeldoutOne}{-2.230}
\newcommand{\csDsHeldoutTwo}{-2.212}

\newcommand{\csDsPriorR}{17\%}

\newcommand{\csDsmixGoldFreeB}{15.0}
\newcommand{\csDsmixGoldFreeVsRef}{above}
\newcommand{\csDsmixHardShare}{20\%}

\newcommand{\csEpsHard}{10^{-6}}

\newcommand{\csEssGainCal}{1.07}
\newcommand{\csExCorrFDR}{0.0000}
\newcommand{\csExCorrPle}{0.0000}
\newcommand{\csExReps}{100,000}
\newcommand{\csExUncorrFDR}{0.1239}
\newcommand{\csExUncorrPle}{0.0909}
\newcommand{\csExcapB}{102.2}

\newcommand{\csExpRTotalB}{13.8}

\newcommand{\csExplicitB}{13.0}
\newcommand{\csFedGrowth}{40\%}

\newcommand{\csFrameB}{126.4}
\newcommand{\csGbmAUCU}{0.79}

\newcommand{\csGenericDisagreedDollars}{21\%}

\newcommand{\csGenericHumanDisagreedDollars}{22\%}
\newcommand{\csGrowthClassic}{37\%}
\newcommand{\csGrowthClassicHi}{60\%}
\newcommand{\csGrowthClassicLo}{20\%}
\newcommand{\csGrowthFrame}{35\%}

\newcommand{\csGrowthPpiClip}{35\%}

\newcommand{\csGrowthPpiOne}{34\%}
\newcommand{\csGrowthPpiOneHi}{53\%}
\newcommand{\csGrowthPpiOneLo}{18\%}
\newcommand{\csGrowthPpiPooled}{35\%}

\newcommand{\csGrowthReadersMax}{31\%}
\newcommand{\csGrowthReadersMin}{24\%}
\newcommand{\csHumanVsMajDollars}{16\%}
\newcommand{\csHumanVsMajTexts}{41\%}
\newcommand{\csHybFiveAccepted}{71}
\newcommand{\csHybFiveAcceptedDollars}{48\%}
\newcommand{\csHybFiveAcceptedErr}{0.002}
\newcommand{\csHybFiveAllAccepted}{none}
\newcommand{\csHybFiveBiasB}{-0.035}
\newcommand{\csHybFiveBiasRmsB}{0.070}
\newcommand{\csHybFiveCovFifty}{0.91}

\newcommand{\csHybFiveRemaining}{147}
\newcommand{\csHybFiveRemainingMax}{218}
\newcommand{\csHybFiveRemainingMin}{77}
\newcommand{\csHybFiveRmseFifty}{0.26}

\newcommand{\csHybFiveShortFrom}{100}
\newcommand{\csHybSplits}{30}

\newcommand{\csHybTenAcceptedDollars}{98\%}
\newcommand{\csHybTenAcceptedErr}{0.026}
\newcommand{\csHybTenAllAccepted}{3}

\newcommand{\csHybTenBiasRmsB}{0.188}

\newcommand{\csHybTenCovMax}{0.08}

\newcommand{\csHybTenCovZeroFrom}{75}

\newcommand{\csHybTenRemainingMax}{76}
\newcommand{\csHybTenRemainingMin}{0}

\newcommand{\csHybTenShortFrom}{25}
\newcommand{\csInvCalls}{9,618}
\newcommand{\csInvCostUSD}{14}
\newcommand{\csInvTokensIn}{13.9}

\newcommand{\csIsoLLdollar}{0.081}

\newcommand{\csIsoVoteLLdollar}{0.106}

\newcommand{\csLLMDisagreeDollars}{11\%}
\newcommand{\csLLMDisagreeTexts}{20\%}
\newcommand{\csLambdaBFix}{0.17}
\newcommand{\csLambdaBRefit}{1.10}
\newcommand{\csLambdaStar}{0.75}

\newcommand{\csLambdaStarRules}{0.15}
\newcommand{\csLambdaYearMax}{1.50}
\newcommand{\csLambdaYearMin}{0.25}

\newcommand{\csLpCalBothAP}{0.085}

\newcommand{\csLpInsts}{1,025}

\newcommand{\csLpRulesBothAP}{0.073}

\newcommand{\csLpRulesCommonAP}{0.065}
\newcommand{\csLpRulesCommonAUC}{0.962}

\newcommand{\csLpRulesCommonPat}{3.9\%}

\newcommand{\csLpRulesNTest}{4,771,647}

\newcommand{\csLpRulesPosTest}{3,875}

\newcommand{\csLpRulesPrev}{0.08\%}

\newcommand{\csLttBootFiveDollarShare}{96\%}

\newcommand{\csLttBootFiveFdr}{0.021}
\newcommand{\csLttBootFiveFdrHi}{0.051}
\newcommand{\csLttBootFiveFdrLo}{0.003}

\newcommand{\csLttFiveDollarShare}{80\%}

\newcommand{\csLttFiveExceed}{0\%}
\newcommand{\csLttFiveFdr}{0.006}
\newcommand{\csLttFiveFdrHi}{0.024}
\newcommand{\csLttFiveFdrLo}{0.000}
\newcommand{\csLttFiveFdrMax}{0.036}

\newcommand{\csLttFiveTextShare}{80\%}

\newcommand{\csLttFiveThr}{0.92}

\newcommand{\csLttTenDollarShare}{100\%}

\newcommand{\csLttTenExceed}{0\%}
\newcommand{\csLttTenFdr}{0.031}
\newcommand{\csLttTenFdrHi}{0.064}
\newcommand{\csLttTenFdrLo}{0.008}

\newcommand{\csMachMajorityAdjT}{90\%}

\newcommand{\csMachMajorityOneT}{84\%}

\newcommand{\csMachSonnetAdjT}{89\%}

\newcommand{\csMachSonnetOneT}{85\%}

\newcommand{\csMachSonnetThreeT}{86\%}

\newcommand{\csMajFlashHuman}{64\%}
\newcommand{\csMajFlashHumanDollars}{85\%}

\newcommand{\csMajMiniHuman}{62\%}

\newcommand{\csMajRPpiB}{11.59}
\newcommand{\csMajRPpiHiB}{12.4}
\newcommand{\csMajRPpiLoB}{10.7}

\newcommand{\csMajSonnetHuman}{59\%}

\newcommand{\csMajVsOneB}{0.26}

\newcommand{\csNEffAll}{335}
\newcommand{\csNEligible}{113,142}
\newcommand{\csNFeatures}{58}
\newcommand{\csNGrid}{40}
\newcommand{\csNLLMPatterns}{10}
\newcommand{\csNNotSplit}{499}
\newcommand{\csNNullBinary}{53}

\newcommand{\csNNullBinaryOne}{67}

\newcommand{\csNPredictors}{8}
\newcommand{\csNReaderThree}{651}
\newcommand{\csNReaderTwo}{149}
\newcommand{\csNReaderTwoFull}{651}
\newcommand{\csNReaderTwoRest}{502}
\newcommand{\csNRecheck}{150}
\newcommand{\csNRows}{799,675}
\newcommand{\csNSample}{651}
\newcommand{\csNSplit}{152}
\newcommand{\csNStatusOnlySplit}{173}
\newcommand{\csNTextFeatures}{16}
\newcommand{\csNTexts}{121,070}

\newcommand{\csNUnsampledPiZero}{391}

\newcommand{\csNYearMax}{391}
\newcommand{\csNYearMin}{306}
\newcommand{\csNaiveFiveDollarShare}{71\%}

\newcommand{\csNaiveFiveFdr}{0.005}

\newcommand{\csNestedModalC}{0.03}
\newcommand{\csNestedShareSameC}{74\%}
\newcommand{\csNoCtxFlashHuman}{62\%}

\newcommand{\csNoCtxFlashSameDollars}{97\%}
\newcommand{\csNoCtxFlashSameRDollars}{99\%}
\newcommand{\csNoCtxFlashSameTexts}{79\%}

\newcommand{\csNoCtxMiniSameTexts}{94\%}

\newcommand{\csOccFirstThreeStatus}{71\%}

\newcommand{\csOccFirstTwoKappa}{0.24}
\newcommand{\csOccFirstTwoN}{149}

\newcommand{\csOccFirstTwoStatus}{56\%}
\newcommand{\csOccFirstTwoU}{30\%}

\newcommand{\csOccRestThreeStatus}{70\%}

\newcommand{\csOccRestTwoKappa}{0.40}
\newcommand{\csOccRestTwoN}{502}

\newcommand{\csOccRestTwoStatus}{65\%}
\newcommand{\csOccRestTwoU}{23\%}
\newcommand{\csOrigFlashHuman}{56\%}
\newcommand{\csOrigFlashHumanDollars}{84\%}
\newcommand{\csOrigMiniHuman}{67\%}
\newcommand{\csOrigMiniHumanDollars}{85\%}

\newcommand{\csOrigSonnetHumanDollars}{83\%}
\newcommand{\csOutvotedOneModel}{72\%}

\newcommand{\csOutvotedOneModelWd}{78\%}
\newcommand{\csOutvotedOneN}{36}
\newcommand{\csOutvotedThreeModel}{56\%}

\newcommand{\csOutvotedThreeModelWd}{62\%}
\newcommand{\csOutvotedThreeN}{50}
\newcommand{\csOutvotedTwoModel}{74\%}

\newcommand{\csOutvotedTwoModelWd}{80\%}
\newcommand{\csOutvotedTwoN}{66}
\newcommand{\csPFedClassic}{0.38}

\newcommand{\csPFedPpiOne}{0.26}

\newcommand{\csPFedShiftPpiOne}{0.25}

\newcommand{\csPZeroPpiOne}{$<$0.001}

\newcommand{\csPairFlashDollar}{-0.014}
\newcommand{\csPairFlashDollarHi}{-0.004}
\newcommand{\csPairFlashDollarLo}{-0.026}

\newcommand{\csPairIsoAuc}{+0.004}
\newcommand{\csPairIsoAucHi}{+0.009}
\newcommand{\csPairIsoAucLo}{+0.001}
\newcommand{\csPairIsoDollar}{+0.000}
\newcommand{\csPairIsoDollarHi}{+0.009}
\newcommand{\csPairIsoDollarLo}{-0.009}
\newcommand{\csPairIsoDollarOne}{-0.021}
\newcommand{\csPairIsoDollarOneHi}{-0.006}
\newcommand{\csPairIsoDollarOneLo}{-0.038}

\newcommand{\csPairMiniDollar}{-0.007}
\newcommand{\csPairMiniDollarHi}{+0.000}
\newcommand{\csPairMiniDollarLo}{-0.015}

\newcommand{\csPairRatioHundred}{2.1}

\newcommand{\csPairRatioSdHundred}{0.3}

\newcommand{\csPairShareHundred}{100\%}

\newcommand{\csPairSonnetDollar}{-0.009}
\newcommand{\csPairSonnetDollarHi}{+0.001}
\newcommand{\csPairSonnetDollarLo}{-0.019}

\newcommand{\csPairTextOnlyDollar}{-0.055}
\newcommand{\csPairTextOnlyDollarHi}{-0.034}
\newcommand{\csPairTextOnlyDollarLo}{-0.082}

\newcommand{\csPanelAUCU}{0.62}
\newcommand{\csPanelAUCdollar}{0.991}

\newcommand{\csPanelLLdollar}{0.082}
\newcommand{\csPanelLLdollarHi}{0.118}
\newcommand{\csPanelLLdollarLo}{0.055}
\newcommand{\csPanelLLtext}{0.16}
\newcommand{\csPanelLLtextHi}{0.24}
\newcommand{\csPanelLLtextLo}{0.10}

\newcommand{\csParaFlashAllSameDollars}{95\%}
\newcommand{\csParaFlashAllSameTexts}{73\%}
\newcommand{\csParaFlashHumanMax}{71\%}
\newcommand{\csParaFlashHumanMin}{58\%}
\newcommand{\csParaFlashNPrompts}{5}
\newcommand{\csParaFlashSameDollarsMax}{98\%}
\newcommand{\csParaFlashSameDollarsMin}{96\%}

\newcommand{\csParaFlashSameTextsMax}{97\%}
\newcommand{\csParaFlashSameTextsMin}{81\%}

\newcommand{\csParaMiniAllSameTexts}{82\%}

\newcommand{\csParaMiniNPrompts}{3}

\newcommand{\csParaMiniSameTextsMax}{91\%}
\newcommand{\csParaMiniSameTextsMin}{86\%}

\newcommand{\csParaSonnetAllSameTexts}{91\%}

\newcommand{\csParaSonnetNPrompts}{3}
\newcommand{\csParaSonnetSameDollarsMax}{97\%}
\newcommand{\csParaSonnetSameDollarsMin}{95\%}

\newcommand{\csParaSonnetSameTextsMax}{95\%}
\newcommand{\csParaSonnetSameTextsMin}{94\%}

\newcommand{\csParaVFlashSameDollarsRange}{96\%--99\%}

\newcommand{\csParaVFlashSameTextsRange}{81\%--97\%}
\newcommand{\csParaVMinSameRDollars}{99\%}

\newcommand{\csParaVMiniSameDollarsRange}{96\%}

\newcommand{\csParaVMiniSameTextsRange}{83\%--93\%}

\newcommand{\csParaVSonnetSameDollarsRange}{95\%--97\%}

\newcommand{\csParaVSonnetSameTextsRange}{94\%--98\%}

\newcommand{\csPlugBFix}{+11\%}

\newcommand{\csPluginAllMaxB}{15.0}
\newcommand{\csPluginAllMinB}{12.5}
\newcommand{\csPluginCalB}{14.15}

\newcommand{\csPpiAllMaxB}{13.8}
\newcommand{\csPpiAllMinB}{13.4}
\newcommand{\csPpiAllSpreadB}{0.4}
\newcommand{\csPpiB}{13.49}
\newcommand{\csPpiBoot}{2,000}
\newcommand{\csPpiHiB}{14.2}
\newcommand{\csPpiLoB}{12.8}
\newcommand{\csPpiStarB}{13.48}
\newcommand{\csPpiStarHiB}{14.2}
\newcommand{\csPpiStarLoB}{12.8}

\newcommand{\csPpiWidthB}{1.43}
\newcommand{\csPpsDrawsZeroDomain}{39}
\newcommand{\csPpsMaxDev}{0.005}
\newcommand{\csProtActiveFifty}{0.34}
\newcommand{\csProtActiveHundred}{0.16}

\newcommand{\csProtClassicHundred}{1.08}

\newcommand{\csProtCovActiveFifty}{0.94}
\newcommand{\csProtCovActiveHundred}{0.94}
\newcommand{\csProtCovActiveHundredFifty}{0.91}

\newcommand{\csProtCovMixedHundred}{0.94}
\newcommand{\csProtCovMixedHundredFifty}{0.93}

\newcommand{\csProtCovMixedUnifHundred}{0.94}
\newcommand{\csProtCovMixedUnifHundredFifty}{0.93}

\newcommand{\csProtMixedHundred}{0.21}

\newcommand{\csProtMixedUnifHundred}{0.20}

\newcommand{\csProtPpsHundred}{0.32}

\newcommand{\csProtRatioFifty}{1.6}
\newcommand{\csProtRatioHundred}{2.0}

\newcommand{\csProtReps}{300}
\newcommand{\csProtSize}{218}
\newcommand{\csProtSplits}{30}

\newcommand{\csProtUnifHundred}{0.27}

\newcommand{\csRGOneSe}{0.95}
\newcommand{\csRGOneSp}{0.97}

\newcommand{\csRGOneTotalB}{11.15}
\newcommand{\csRGPooledShare}{13.7\%}
\newcommand{\csRGPostShare}{13.2\%}
\newcommand{\csRGSe}{0.87}
\newcommand{\csRGSp}{0.95}

\newcommand{\csRGThreeSe}{0.89}
\newcommand{\csRGThreeSp}{0.99}

\newcommand{\csRGThreeTotalB}{13.89}
\newcommand{\csRGTotalB}{9.76}
\newcommand{\csRShareOne}{21\%}
\newcommand{\csRShareThree}{28\%}
\newcommand{\csRShareTwo}{19\%}
\newcommand{\csRShareWdOne}{11.5\%}
\newcommand{\csRShareWdThree}{14.2\%}
\newcommand{\csRShareWdTwo}{10.2\%}

\newcommand{\csReaderOneAltClassicB}{11.29}

\newcommand{\csReaderOneAltPpiB}{11.34}
\newcommand{\csReaderOneAltPpiHiB}{12.2}
\newcommand{\csReaderOneAltPpiLoB}{10.4}

\newcommand{\csReaderOneUIsThreeN}{55\%}
\newcommand{\csReaderOneUIsThreeR}{29\%}

\newcommand{\csReaderThreePpiB}{14.04}
\newcommand{\csReaderThreePpiHiB}{15.1}
\newcommand{\csReaderThreePpiLoB}{13.0}

\newcommand{\csReaderTotalMaxB}{15.0}
\newcommand{\csReaderTotalMinB}{12.5}

\newcommand{\csReaderTotalRulesB}{13.0}

\newcommand{\csReaderTwoPpiB}{10.10}
\newcommand{\csReaderTwoPpiHiB}{11.1}
\newcommand{\csReaderTwoPpiLoB}{9.1}

\newcommand{\csReadersBoot}{1,000}

\newcommand{\csRecheckAltPpiB}{10.90}
\newcommand{\csRecheckAltPpiHiB}{11.8}
\newcommand{\csRecheckAltPpiLoB}{10.0}

\newcommand{\csRegAPpsSixFiftyOne}{2.9}

\newcommand{\csRegASrsSixFiftyOne}{7.0}

\newcommand{\csRegAStratMiniSixFiftyOne}{1.0}
\newcommand{\csRegAStratSixFiftyOne}{1.1}

\newcommand{\csRegBSrsSixFiftyOne}{12.5}

\newcommand{\csRegBStratSixFiftyOne}{0.5}

\newcommand{\csRegDraws}{200}
\newcommand{\csRegSeMax}{8.03}
\newcommand{\csRegSeMin}{0.01}

\newcommand{\csRelPanelDollarSentence}{By dollars, the interval of the calibrated panel covers the diagonal in all eight bins that are drawn.}

\newcommand{\csRelPanelTextNeffMax}{23}
\newcommand{\csRelPanelTextNeffMin}{2.0}

\newcommand{\csRelVoteTopObsDollars}{95\%}
\newcommand{\csRelVoteTopObsDollarsOne}{80\%}
\newcommand{\csRelVoteTopObsTexts}{93\%}
\newcommand{\csRelVoteTopObsTextsOne}{37\%}

\newcommand{\csRepeatFlashHumanDollars}{84\%}
\newcommand{\csRepeatFlashSameDollars}{98\%}
\newcommand{\csRepeatFlashSameRDollars}{99\%}
\newcommand{\csRepeatFlashSameTexts}{98\%}
\newcommand{\csRepeatMiniHuman}{74\%}
\newcommand{\csRepeatMiniHumanDollars}{86\%}
\newcommand{\csRepeatMiniSameDollars}{96\%}
\newcommand{\csRepeatMiniSameRDollars}{99\%}
\newcommand{\csRepeatMiniSameTexts}{91\%}

\newcommand{\csRepeatSonnetHumanDollars}{83\%}
\newcommand{\csRepeatSonnetSameDollars}{97\%}
\newcommand{\csRepeatSonnetSameRDollars}{99\%}
\newcommand{\csRepeatSonnetSameTexts}{95\%}

\newcommand{\csRerunAblFullMax}{+0.0055}
\newcommand{\csRerunAblFullMin}{-0.0025}
\newcommand{\csRerunAcqCovMax}{94\%}
\newcommand{\csRerunAcqCovMin}{92\%}
\newcommand{\csRerunAcqRatioMax}{1.74}
\newcommand{\csRerunAcqRatioMin}{1.19}
\newcommand{\csRerunAdBatchAvgHundredMax}{1.54}
\newcommand{\csRerunAdBatchAvgHundredMin}{1.19}

\newcommand{\csRerunAdClassicHundredMax}{2.66}
\newcommand{\csRerunAdClassicHundredMin}{1.60}
\newcommand{\csRerunAdClassicTwoHundredMax}{2.77}
\newcommand{\csRerunAdClassicTwoHundredMin}{1.40}
\newcommand{\csRerunAdNoPilotHundredMax}{1.59}
\newcommand{\csRerunAdNoPilotHundredMin}{1.32}
\newcommand{\csRerunAdNoPilotTwoHundredMax}{1.59}
\newcommand{\csRerunAdNoPilotTwoHundredMin}{1.19}
\newcommand{\csRerunAdRankingHolds}{every}
\newcommand{\csRerunAdjAUC}{0.965}
\newcommand{\csRerunAdjAcqCov}{94\%}
\newcommand{\csRerunAdjAcqRatio}{1.50}

\newcommand{\csRerunAdjDLL}{0.082}

\newcommand{\csRerunAdjEss}{1.07}
\newcommand{\csRerunAdjGrowth}{33\%}

\newcommand{\csRerunAdjGrowthP}{0.20}

\newcommand{\csRerunAdjLL}{0.156}
\newcommand{\csRerunAdjLam}{0.75}
\newcommand{\csRerunAdjLtt}{79.8\%}

\newcommand{\csRerunAdjLttTen}{100.0\%}
\newcommand{\csRerunAdjLttTenErr}{3.1\%}

\newcommand{\csRerunDriftCalAbsMax}{1.5}

\newcommand{\csRerunDriftCalMax}{+1.5}
\newcommand{\csRerunDriftCalMin}{-1.3}
\newcommand{\csRerunEssMax}{1.23}
\newcommand{\csRerunEssMin}{1.01}
\newcommand{\csRerunGrowthMax}{39\%}
\newcommand{\csRerunGrowthMin}{23\%}
\newcommand{\csRerunGrowthPMax}{0.46}
\newcommand{\csRerunGrowthPMin}{0.04}

\newcommand{\csRerunInvFlashHumanWdMax}{83\%}
\newcommand{\csRerunInvFlashHumanWdMin}{79\%}

\newcommand{\csRerunInvMiniHumanWdMax}{84\%}
\newcommand{\csRerunInvMiniHumanWdMin}{78\%}

\newcommand{\csRerunInvSonnetHumanWdMax}{82\%}
\newcommand{\csRerunInvSonnetHumanWdMin}{76\%}

\newcommand{\csRerunLamMax}{0.85}
\newcommand{\csRerunLamMin}{0.60}
\newcommand{\csRerunLttEmptyRefs}{reader~2 and reader~3}

\newcommand{\csRerunLttTenErrMax}{2.6\%}
\newcommand{\csRerunLttTenMax}{97.2\%}
\newcommand{\csRerunLttTenMin}{85.1\%}
\newcommand{\csRerunMajAUC}{0.946}

\newcommand{\csRerunMajDLL}{0.079}

\newcommand{\csRerunMajLL}{0.243}

\newcommand{\csRerunMajLtt}{85.7\%}

\newcommand{\csRerunMajPpiB}{11.56}
\newcommand{\csRerunMajPpiHiB}{12.3}
\newcommand{\csRerunMajPpiLoB}{10.8}
\newcommand{\csRerunMajSaKappa}{0.77}

\newcommand{\csRerunOneAUC}{0.784}

\newcommand{\csRerunOneClassicB}{11.29}
\newcommand{\csRerunOneClassicHiB}{12.2}
\newcommand{\csRerunOneClassicLoB}{10.3}

\newcommand{\csRerunOneLL}{0.397}

\newcommand{\csRerunOneLtt}{79.9\%}

\newcommand{\csRerunOnePpiB}{11.29}
\newcommand{\csRerunOnePpiHiB}{12.1}
\newcommand{\csRerunOnePpiLoB}{10.4}
\newcommand{\csRerunOneSaKappa}{1.85}

\newcommand{\csRerunRidgeMainHundredMax}{40\%}
\newcommand{\csRerunRidgeMainHundredMin}{20\%}
\newcommand{\csRerunRidgeMainTwoHundredMax}{43\%}
\newcommand{\csRerunRidgeMainTwoHundredMin}{21\%}
\newcommand{\csRerunRidgeTunedHundredMax}{36\%}
\newcommand{\csRerunRidgeTunedHundredMin}{9\%}
\newcommand{\csRerunRidgeTunedTwoHundredMax}{40\%}
\newcommand{\csRerunRidgeTunedTwoHundredMin}{22\%}
\newcommand{\csRerunRidgeTwoMainHundred}{20\%}
\newcommand{\csRerunRidgeTwoTunedHundred}{9\%}

\newcommand{\csRerunSaMovesMax}{6}
\newcommand{\csRerunSaMovesMin}{1}
\newcommand{\csRerunSaTopCalMaxB}{0.51}
\newcommand{\csRerunSaTopCalMinB}{0.36}

\newcommand{\csRerunSaTopDirectMaxB}{0.66}
\newcommand{\csRerunSaTopDirectMinB}{0.47}

\newcommand{\csRerunThreeAUC}{0.938}

\newcommand{\csRerunThreeLL}{0.223}

\newcommand{\csRerunThreePpiB}{14.00}
\newcommand{\csRerunThreePpiHiB}{15.0}
\newcommand{\csRerunThreePpiLoB}{13.0}

\newcommand{\csRerunTwoAUC}{0.957}

\newcommand{\csRerunTwoLL}{0.213}

\newcommand{\csRerunTwoPpiB}{10.12}
\newcommand{\csRerunTwoPpiHiB}{11.1}
\newcommand{\csRerunTwoPpiLoB}{9.1}
\newcommand{\csRerunTwoSaKappa}{0.63}

\newcommand{\csRerunVsFixedMaxB}{0.05}
\newcommand{\csRetuneBias}{+0.37\%}
\newcommand{\csRetuneBiasMinusClassic}{-0.01\%}
\newcommand{\csRetuneBiasSE}{0.28\%}
\newcommand{\csRetuneClassicBias}{+0.38\%}
\newcommand{\csRetuneClassicCov}{0.970}
\newcommand{\csRetuneCov}{0.970}

\newcommand{\csRetuneDraws}{100}

\newcommand{\csRetuneShareSameC}{32\%}

\newcommand{\csRidgeCMain}{0.1}

\newcommand{\csRidgeClassicHundredRmse}{2.37}

\newcommand{\csRidgeCovMax}{0.96}
\newcommand{\csRidgeCovMin}{0.92}
\newcommand{\csRidgeGrid}{0.003, 0.01, 0.03, 0.1, 0.3, 1}

\newcommand{\csRidgeMainHundredPct}{43\%}
\newcommand{\csRidgeMainHundredRmse}{1.36}

\newcommand{\csRidgeMainTwoHundredPct}{48\%}

\newcommand{\csRidgeModerateHundredMaxPct}{45\%}
\newcommand{\csRidgeModerateHundredMinPct}{39\%}
\newcommand{\csRidgeModerateTwoHundredMaxPct}{48\%}
\newcommand{\csRidgeModerateTwoHundredMinPct}{45\%}
\newcommand{\csRidgeReps}{1,000}

\newcommand{\csRidgeStrongHundredPct}{27\%}

\newcommand{\csRidgeTunedHundredPct}{42\%}

\newcommand{\csRidgeTunedTwoHundredPct}{47\%}

\newcommand{\csRidgeVeryStrongHundredPct}{4\%}

\newcommand{\csRmseCalActiveHundred}{0.80}

\newcommand{\csRmseCalMixedHundred}{0.87}

\newcommand{\csRmseCalPpsHundred}{1.26}

\newcommand{\csRmseDsActiveHundred}{2.39}

\newcommand{\csRmseDsMixedHundred}{1.41}

\newcommand{\csRmseDsPpsHundred}{1.31}

\newcommand{\csSUFifty}{0.485}
\newcommand{\csSUFive}{0.026}
\newcommand{\csSUTen}{0.063}
\newcommand{\csSUTwenty}{0.161}
\newcommand{\csSUn}{1060}
\newcommand{\csSaAHundredCalRel}{4\%}

\newcommand{\csSaAHundredCompBeatsCal}{48\%}
\newcommand{\csSaAHundredCompBeatsDirect}{100\%}
\newcommand{\csSaAHundredCompCov}{0.61}

\newcommand{\csSaAHundredCompRel}{4\%}

\newcommand{\csSaAHundredDirectCov}{0.90}

\newcommand{\csSaAHundredDirectRel}{17\%}

\newcommand{\csSaAHundredPps}{47.5}

\newcommand{\csSaAOneCalRel}{16\%}

\newcommand{\csSaAOneCompCov}{0.23}

\newcommand{\csSaAOneCompRel}{17\%}

\newcommand{\csSaAOneDirectCov}{0.35}
\newcommand{\csSaAOneDirectCovFive}{0.41}

\newcommand{\csSaASmallCalRel}{62\%}

\newcommand{\csSaASmallCompCov}{0.18}

\newcommand{\csSaASmallCompRel}{66\%}

\newcommand{\csSaASmallDirectCov}{0.24}

\newcommand{\csSaAZeroN}{1803}

\newcommand{\csSaBHundredCalRel}{6\%}

\newcommand{\csSaBHundredCompCov}{0.84}

\newcommand{\csSaBHundredCompRel}{5\%}

\newcommand{\csSaBHundredDirectCov}{0.93}

\newcommand{\csSaBHundredDirectRel}{11\%}

\newcommand{\csSaBHundredRulesRel}{3\%}

\newcommand{\csSaBOneCalRel}{10\%}

\newcommand{\csSaBOneCompRel}{11\%}

\newcommand{\csSaBZeroN}{2182}

\newcommand{\csSaCHundredCompCov}{0.64}

\newcommand{\csSaCHundredDirectCov}{0.82}

\newcommand{\csSaCZeroN}{573}

\newcommand{\csSaDraws}{200}

\newcommand{\csSaGammaTopTwenty}{0.26}
\newcommand{\csSaHundredDirectCovMax}{0.93}
\newcommand{\csSaHundredDirectCovMin}{0.82}
\newcommand{\csSaInstOneCalB}{0.49}
\newcommand{\csSaInstOneCompB}{0.52}
\newcommand{\csSaInstOneCompSeB}{0.06}
\newcommand{\csSaInstOneDirectB}{0.62}
\newcommand{\csSaInstOneDirectSeB}{0.13}
\newcommand{\csSaInstOneDollarsB}{5.41}
\newcommand{\csSaInstOneGamma}{0.23}
\newcommand{\csSaInstOneN}{138}
\newcommand{\csSaInstOneName}{Johns Hopkins University}
\newcommand{\csSaInstOneRulesB}{0.41}

\newcommand{\csSaInstTwoCalB}{0.69}

\newcommand{\csSaInstTwoDirectB}{0.56}

\newcommand{\csSaInstTwoName}{Stanford University}

\newcommand{\csSaKappaReal}{0.49}
\newcommand{\csSaMedianSampled}{10}
\newcommand{\csSaNInst}{3,204}
\newcommand{\csSaShareWithSample}{96\%}
\newcommand{\csSaTopTwentyDirectMoves}{2}

\newcommand{\csSampleAdjN}{372}
\newcommand{\csSampleAdjR}{181}
\newcommand{\csSampleAdjU}{98}

\newcommand{\csSampleN}{389}
\newcommand{\csSampleR}{136}
\newcommand{\csSampleThreeN}{367}
\newcommand{\csSampleThreeR}{180}
\newcommand{\csSampleThreeU}{104}
\newcommand{\csSampleTwoN}{364}
\newcommand{\csSampleTwoR}{126}
\newcommand{\csSampleTwoU}{161}
\newcommand{\csSampleU}{126}
\newcommand{\csSensMajAUC}{0.98}

\newcommand{\csSensMajLL}{0.16}
\newcommand{\csSensReaderOneAUC}{0.80}

\newcommand{\csSensReaderOneLL}{0.51}
\newcommand{\csSensReaderThreeAUC}{0.96}

\newcommand{\csSensReaderThreeLL}{0.19}
\newcommand{\csSensReaderTwoAUC}{0.97}

\newcommand{\csSensReaderTwoLL}{0.16}
\newcommand{\csSensRecheckAUC}{0.77}

\newcommand{\csSensRecheckLL}{0.39}
\newcommand{\csSensRefAUC}{0.97}

\newcommand{\csSensRefLL}{0.16}

\newcommand{\csSimDraws}{400}
\newcommand{\csSimOneCfinal}{0.1}
\newcommand{\csSimOneCovBRefit}{0.930}
\newcommand{\csSimOneCovBShortSE}{1.6}
\newcommand{\csSimOneWidthBRefit}{0.08}

\newcommand{\csSplitDollars}{9.3\%}

\newcommand{\csStressALttFive}{0.0071}

\newcommand{\csStressALttFiveViol}{0.000}

\newcommand{\csStressAWbhTen}{0.0001}

\newcommand{\csStressDLttTenGivenNonempty}{0.0607}

\newcommand{\csStressDLttTenPAny}{0.18}

\newcommand{\csStressDraws}{200}

\newcommand{\csStressELttTenGivenNonempty}{0.0307}

\newcommand{\csStressELttTenPAny}{0.12}

\newcommand{\csStressLttViolMax}{0.000}
\newcommand{\csStressLttViolUpMax}{0.015}

\newcommand{\csStressSUAFive}{0.013}
\newcommand{\csStressSUATen}{0.052}
\newcommand{\csStressSUATwenty}{0.152}
\newcommand{\csStressSUNullFive}{0.047}
\newcommand{\csStressSUNullTen}{0.096}
\newcommand{\csStressSUNullTwenty}{0.196}
\newcommand{\csStressWbhViolMax}{0.000}
\newcommand{\csStressWbhViolUpMax}{0.015}
\newcommand{\csSumYearsClassicB}{13.47}

\newcommand{\csTextMeasureMaxCert}{0}
\newcommand{\csTextMinP}{0.0003}

\newcommand{\csTextShareBoundBelowTwenty}{33\%}

\newcommand{\csTextsExcap}{113,749}
\newcommand{\csTextsZeroDomain}{7,321}

\newcommand{\csThreeLttTenDollarShare}{56\%}
\newcommand{\csThreeLttTenEmpty}{62\%}
\newcommand{\csThreeLttTenFdr}{0.011}
\newcommand{\csThreeLttTenFdrNonEmpty}{0.028}

\newcommand{\csThreeLttTwentyDollarShare}{96\%}

\newcommand{\csThreeWbhTenEmpty}{86\%}
\newcommand{\csThreeWbhTenFdr}{0.009}
\newcommand{\csThreeWbhTenFdrNonEmpty}{0.067}

\newcommand{\csTruthCB}{27.33}
\newcommand{\csTruthCExpectedB}{21.5}
\newcommand{\csTruthCFlippedTop}{2}
\newcommand{\csTruthCFlippedTopB}{9.9}
\newcommand{\csTstarB}{13.85}

\newcommand{\csTwoUnitFixedBias}{+0.500}

\newcommand{\csTwoUnitPoissonBias}{+0.000}
\newcommand{\csTwoUnitSwitchFixedBias}{+0.250}
\newcommand{\csTwoUnitSwitchPoissonBias}{-0.125}
\newcommand{\csTwoUnitSwitchPoissonExp}{0.875}

\newcommand{\csUShareOne}{19\%}
\newcommand{\csUShareThree}{16\%}
\newcommand{\csUShareTwo}{25\%}

\newcommand{\csUnanBinaryTexts}{77\%}

\newcommand{\csUnanBinaryWd}{91\%}

\newcommand{\csUnanErrDollarBin}{1.8\%}

\newcommand{\csUnanErrTextBin}{2.8\%}

\newcommand{\csUnanShareDollar}{89\%}
\newcommand{\csUnanShareText}{66\%}
\newcommand{\csUnanStatusTexts}{50\%}

\newcommand{\csUnclearPpiB}{7.4}
\newcommand{\csUnclearPpiHiB}{8.8}
\newcommand{\csUnclearPpiLoB}{6.0}

\newcommand{\csUnnestedLttFiveFdr}{0.006}

\newcommand{\csWbhTenDollarShareOne}{79\%}

\newcommand{\csWbhTwentyDollarShare}{77\%}
\newcommand{\csWbhTwentyEmpty}{0\%}

\newcommand{\csWbhTwentyFdr}{0.011}
\newcommand{\csWbhTwentyFdrHi}{0.036}
\newcommand{\csWbhTwentyFdrLo}{0.000}

\newcommand{\csWbhTwentyPowerHeld}{46\%}
\newcommand{\csWbhTwentyTextShare}{78\%}

\newcommand{\csWidthBRefit}{0.15}
\newcommand{\csWidthCutCal}{3\%}
\newcommand{\csWidthRatioCal}{0.97}
\newcommand{\csWidthRatioRules}{1.35}
\newcommand{\csWithinFlashDollars}{2.5\%}
\newcommand{\csWithinFlashDollarsHi}{3.7\%}
\newcommand{\csWithinFlashDollarsLo}{1.5\%}
\newcommand{\csWithinFlashTexts}{2.7\%}
\newcommand{\csWithinFlashTextsHi}{6.1\%}
\newcommand{\csWithinFlashTextsLo}{1.2\%}
\newcommand{\csWithinMiniDollars}{2.3\%}

\newcommand{\csWithinMiniTexts}{0.7\%}

\newcommand{\csWithinRepFlashDollars}{1.2\%}

\newcommand{\csWithinRepFlashTexts}{2.7\%}

\newcommand{\csWithinRepMiniDollars}{2.4\%}

\newcommand{\csWithinRepMiniTexts}{1.4\%}

\newcommand{\csWithinRepSonnetDollars}{1.5\%}

\newcommand{\csWithinRepSonnetTexts}{1.4\%}

\newcommand{\csWithinSonnetDollars}{2.3\%}

\newcommand{\csWithinSonnetTexts}{3.4\%}

\newcommand{\csWithinVFlashDollars}{2.4\%}

\newcommand{\csWithinVMiniDollars}{2.8\%}

\newcommand{\csWithinVSonnetDollars}{1.9\%}

\newcommand{\csWtAdjAnyN}{150}
\newcommand{\csWtAdjIndepTexts}{0.2\%}

\newcommand{\csWtAdjN}{25}
\newcommand{\csWtAdjOfAnyTexts}{17\%}

\newcommand{\csWtAdjTexts}{3.8\%}

\newcommand{\csWtAdjWd}{1.6\%}

\newcommand{\csWtNMax}{70}
\newcommand{\csWtNMin}{42}

\newcommand{\csWtOneN}{59}

\newcommand{\csWtOneOthersBoth}{20}

\newcommand{\csWtOneOthersSide}{47}

\newcommand{\csWtOneTopTwo}{78\%}
\newcommand{\csWtOneTopTwoOthers}{both}
\newcommand{\csWtOneTopTwoWeightMin}{5,500}
\newcommand{\csWtOneW}{16.8\%}
\newcommand{\csWtOneWNoTopTwo}{4.3\%}

\newcommand{\csWtTextsMax}{10.8\%}
\newcommand{\csWtTextsMin}{6.5\%}

%% file: latex_report/numbers_validation.tex
\newcommand{\valBigClusterB}{13.9}
\newcommand{\valBigClusterRows}{45,082}

\newcommand{\valNTakeAll}{71}

\newcommand{\valRangeHi}{15}

\newcommand{\valRangeLo}{11}

\newcommand{\valSelfSame}{75\%}

%% file: sections/intro.tex
\section{Introduction}\label{sec:intro}

Every year US private foundations and grant-making charities report each grant they make on a tax return (Form
990-PF, or Schedule~I of Form 990): the recipient, the amount, and a short purpose written by the filer, such as
``General support'', ``Cancer research'' or ``Toward the new engineering building''. The form has no box for
research, so anyone who wants to know how much private money reaches universities for research has to read the
purposes and decide. For tax years 2020--2024 that is \csNRows{} grant rows worth \$\csFrameB{}B to US colleges
and universities; many rows repeat the same purpose text word for word, so there are \csNTexts{} distinct texts
to read, too many for a person and routine for a language model. The companion report had four machine readers, a
rule-based classifier and three language models, label every text under a written codebook, and a person label a
sample of \csNSample{} texts under the same codebook. The readers disagree. The dollars each calls research run
from \$\csReaderTotalMinB{}B to \$\csReaderTotalMaxB{}B; the three language models give different labels on
\csLLMDisagreeTexts{} of texts carrying \csLLMDisagreeDollars{} of the dollars; and three people who read the
sample give the same research-or-not answer on \csUnanBinaryTexts{} of texts (\csUnanBinaryWd{} of dollars). The
companion reported the result as a spread, between \$\valRangeLo{}B and \$\valRangeHi{}B depending on who reads.
That is an honest statement, but an office cannot act on it. This paper is about the three decisions an office has
to take instead: which machine labels to accept, and at what error; which texts a person should read next, with a
limited budget; and what total to report, with what interval, when the answer depends on who did the reading.

A growing share of what is counted about institutions begins as free text. A grant is a line on a tax return with
a purpose written by the filer; a procurement record has a description; a permit has a reason. Before any of these
can be added up by category, something has to read the text and decide what category it names. Large language
models now do this reading at a scale and price that make population-wide classification routine
\parencite{gilardi2023,ziems2024}, and the natural next step, taken in the testbed of this paper, is to have several
models read everything and a person read a sample. That step exposes the problem we study: \emph{the readers
disagree}, with one another and with the person, and a total built from any one reading inherits that reading's
bias. The office that owns the records is left with the three decisions above, each to be taken under
uncertainty about the reading itself.

Three features of administrative text make these decisions harder than the standard annotation setting. First, the
unit that is read is not the unit that is counted. The same purpose text (``General support'', ``Research'',
``For the grantee's exempt purposes'') appears on thousands of records; in our testbed one text covers
\valBigClusterRows{} rows and \$\valBigClusterB{}B. That text is the single word ``Education'', and the three
readers and the four machine readers all label it not research. A label is a decision about a \emph{shared evidence unit}, and
its error propagates to every record that carries it. Second, the estimand is a dollar sum, not an accuracy, so an
error on a rare large text matters more than many errors on small ones, and any guarantee must be stated in dollars.
Third, the only reference is a human reading of a designed probability sample, drawn by dollars within strata
defined by the machine readers' own votes, so the sample is neither uniform nor independent of the predictions.

The companion report \parencite{koomullil2026companion} reported this situation as a spread: depending
on which reading of the purpose texts one adopts, between \$\valRangeLo{}B and \$\valRangeHi{}B of private grants to
US universities carried a stated research purpose in 2020--2024. The tools that turn a spread into a decision exist
in the statistics and machine-learning literatures of the last three years: prediction-powered inference and its
stratified, design-based, cross-fitted and active variants \parencite{angelopoulos2023ppi,fisch2024stratppi,
egami2023dsl,zrnic2024cross,zrnic2024active}, calibrated aggregation of noisy annotators
\parencite{dawidskene1979,whitehill2009glad}, selection with error control from conformal p-values
\parencite{jincandes2023,jincandes2023weighted} and risk control over a family of decision rules
\parencite{angelopoulos2021ltt}. None of them was stated for a dollar estimand over shared evidence units under a
PPS-within-strata design with take-all units, and transferring them carelessly produces guarantees that fail
simple checks. This paper makes the transfer carefully, states what each guarantee assumes and which results are
proved, which approximate and which only evaluated, checks each on the committed design with synthetic truths, and
reports what the methods do and do not change about the testbed's numbers.

\paragraph{Contributions.}
\begin{enumerate}
\item \textbf{Valid totals from shared evidence units} (Sections~\ref{sec:problem}, \ref{sec:ppi}). We specialise
  design-based prediction-powered estimation \parencite{egami2023dsl,fisch2024stratppi} to text-level labels with
  record-level dollars under the committed design, and separate what is exact (a predictor fixed or trained
  independently of the sample; a predictor cross-fitted on folds assigned to the whole frame independently of the sample, under
  Poisson sampling) from what is approximate (cross-fitting under fixed-size sampling, the Rao--Wu interval, the
  tuned $\lambda^*$, the by-year and growth ratios), with a simulation that refits the calibrator inside every
  re-draw and an exact two-unit enumeration that shows why the construction matters. On the testbed, \csNPredictors{} predictors whose plug-in totals span
  \$\csPluginAllMinB{}--\csPluginAllMaxB{}B give rectified estimates within \$\csPpiAllSpreadB{}B of one another;
  the calibrated panel cuts the classic interval by \csWidthCutCal. A regime analysis on the real frame shows why
  the gain is small: the vote-pattern stratification is itself model-assisted estimation, and the same predictor
  reduces the variance by a factor \csRegASrsSixFiftyOne{} under a simple random sample of texts.
\item \textbf{Selective machine labelling at a stated dollar-weighted error, evaluated} (Section~\ref{sec:selective}).
  Two rules decide which machine labels may be accepted at a stated dollar-weighted error. The first builds weighted
  conformal p-values \parencite{tibshirani2019} under the \emph{dollar measure}, in which a test unit is a random
  dollar of the unsampled population and the likelihood-ratio weight of every unit is the dollars it stands for,
  $d_i/\pi_i$; super-uniformity is proved for an idealised draw model and checked, and the weighted
  Benjamini--Hochberg step is checked on the committed design because PRDS is not guaranteed for weighted conformal
  p-values. The second applies Learn-then-Test \parencite{angelopoulos2021ltt} to the design estimate of the dollar
  error of a confidence threshold. Its p-values substitute an effective sample size into a bound for equal-weight
  independent draws, which has no finite-sample justification under unequal weights, so the rule is reported as
  evaluated, not certified: fitted on the whole sample it accepts \csLttFiveDollarShare{} of the unsampled dollars at
  $\alpha = 0.05$, its realised error on nested held-out folds is \csLttFiveFdr, and in \csStressDraws{} re-draws of the design under four
  synthetic truths, one with a true error just above the target and one with the wrong labels on the largest
  candidate texts, no accepted set's realised error exceeded its target (one-sided 95\% upper bound
  \csStressLttViolUpMax{} on each cell's violation probability, against the \csDelta{} a valid rule would allow;
  a check on this design, not a guarantee). The same
  constructions under the text measure accept nothing, and we state the power bound that explains why for this
  family of rules.
\item \textbf{Budgeted human review as active inference, replayed honestly} (Section~\ref{sec:acquisition}). The
  review design that minimises the variance of the dollar total under Poisson sampling samples each text with
  probability proportional to its population dollar mass times its predicted residual. Replayed on a third of the
  sample whose labels enter none of the calibrator's coefficients (the calibrator is fitted on the other two thirds;
  its ridge strength alone is a constant inherited from tuning on the full sample), it reduces the
  RMSE of the total by \csPairRatioHundred$\times$ at 100 texts after a separate pilot relative to
  dollar-proportional sampling with a calibrated predictor, under-covers when its inclusion probabilities approach
  zero, and is worse than dollar sampling with an uncalibrated predictor; a mixture design with a floor on every
  inclusion probability improves coverage and reduces the loss under miscalibration without removing it. The
  result nearest to practice charges every reference label a policy uses, its pilot included (one unit of budget
  is one recorded reference label revealed to the policy; the three readings and the adjudication that produced
  that label are not costed, so the replay does not estimate a saving in reading time): a two-phase design that
  fits the calibrator on its own first phase has
  an RMSE \csAdTwoPhaseVsClassicHundredPct{} below dollar sampling's at a common total budget of 100 texts (ratio
  \csAdTwoPhaseVsClassicHundred; \csRidgeTunedHundredPct{} when the policy chooses its calibrator's penalty on its
  own first-phase labels, and less under a strong fixed penalty, Appendix Table~\ref{tab:adaptive_ridge}), with conditional inclusion probabilities that are fixed before each wave's labels
  are revealed and an estimator we write out. A hybrid estimator (accept at a stated error, read the rest by the
  active design) is kept as a negative result: accepting labels can invalidate an otherwise useful interval.
\end{enumerate}
Supporting analyses measure the dependence among the language-model readers (all three differ from the reference on
\csWtAdjTexts{} of the sampled texts against \csWtAdjIndepTexts{} if independent), the sensitivity of the readings to
instruction wording, context, run and codebook version, the agreement of three human readers and the
reference-dependent results re-run under each of them, a noisy-reference correction, and a small-area composite estimator for one
institution's research dollars, whose point estimate improves on the direct estimate in the simulated regimes and
on the calibrated plug-in in some size classes, and whose intervals under-cover at every size. The reference is
a human reading: three readers labelled the sample independently and then adjudicated together the \csNSplit{}
texts on which they had split, and that adjudicated reading is the reference of this version of the paper (the
first version used the first reader's labels alone). Section~\ref{sec:discussion} states what the methods assume
and what the change of reference moved, and Section~\ref{sec:ppi} re-runs the reference-dependent results under
each reader: the totals, the calibration, the growth test, the acceptance yields, the review replays, the
sensitivity scores and the institution point estimates. The forecast, the regime analysis, the stress tests and
the coverage simulation rest on synthetic truths or machine labels and are held fixed
(Appendix Table~\ref{tab:reference_rerun_sections}).

%% file: sections/related.tex
\section{Related work}\label{sec:related}

\paragraph{Valid inference from machine labels.}
Prediction-powered inference \parencite{angelopoulos2023ppi} estimates a population quantity from predictions on
every unit plus a labelled sample used to \emph{rectify} the prediction's bias; PPI++ \parencite{angelopoulos2023ppipp}
tunes how much of the prediction to use; cross-prediction \parencite{zrnic2024cross} fits the predictor on the
labelled data by cross-fitting; active statistical inference \parencite{zrnic2024active} chooses which units to
label with probabilities driven by the predictor's uncertainty. Two lines bring the sampling design in. Stratified
PPI \parencite{fisch2024stratppi} allocates and tunes within strata; design-based supervised learning
\parencite{egami2023dsl} combines surrogate labels with a gold subset drawn with known, non-uniform probabilities,
the social-science setting closest to ours. \textcite{mozer2026difference} and \textcite{denishaddou2026ppe} place
the estimator in the survey tradition of difference and model-assisted estimators \parencite{sarndal1992}, and
\textcite{klugerbates2026multiwave} give the multiwave two-phase theory that adaptive label collection needs;
\textcite{yeyoga2026allocation} allocate human labels across several LLM-assisted estimates. Earlier post-prediction
corrections are \textcite{wang2020postpred} and \textcite{miao2023popinf}; \textcite{gligoric2025} apply
confidence-driven selection to LLM annotations. What this paper adds is not the difference estimator but its
specialisation: the labelled unit is a text shared by many records, the estimand is a dollar total over those
records, the design is systematic PPS by dollars within strata that are functions of the predictions themselves,
with take-all and certainty units, and the predictor is cross-fitted on the same sample, on folds fixed over the
frame independently of the sample. We state which of the usual properties survive each of those features
(Proposition~\ref{prop:ppi}), show by exact enumeration that the cross-fitting construction and not only the
exclusion of a unit's own label decides exactness, and measure, by a regime analysis, how much of the gain the
stratification itself already takes.

\paragraph{Aggregating disagreeing readers.}
\textcite{dawidskene1979} model each annotator by a confusion matrix and infer the latent class by EM; GLAD
\parencite{whitehill2009glad} adds item difficulty, \textcite{paun2018annotation} compare the Bayesian variants,
and MACE \parencite{hovy2013mace} adds spamming behaviour. Panels of LLM judges are advocated by
\textcite{verga2024juries} and calibrated rather than pruned by \textcite{li2026calibrate};
\textcite{chen2026judge} derive the efficient estimator for a noisy LLM judge with a gold subset, and
\textcite{egamishin2026dmm} identify the error rates of three or more machine readers whose errors are
independent given the latent label and the unit's features, without any gold labels; \textcite{chen2026judge}'s
gold subset and this identification are the two routes to a truth that three disagreeing readers do not supply
on their own. Our aggregation is a stacked \parencite{wolpert1992stacking} cross-fitted calibrator on the
readers' outputs, an item-difficulty Dawid--Skene variant and text cues, anchored to the human reference; its
contribution is empirical: the machine readers differ from the reference together far more often than
independent errors would, a pooled association that does not test conditional independence
(Section~\ref{sec:aggregation}),
whether the homogeneous gold-free estimate recovers the reference total depends on the reading it is held against
(it is inside the rectified interval under the adjudicated reference and outside it under one reader's labels),
and on dollar log-loss no resolved improvement of the panel over the best single language model with text cues,
or over the unsupervised posterior recalibrated on the sample, was detected at this sample size.

\paragraph{Selective labelling, conformal selection and risk control.}
\textcite{jincandes2023} show that Benjamini--Hochberg \parencite{bh1995} on conformal p-values selects units with
FDR control, with PRDS established by \textcite{bates2023outliers}; \textcite{huang2025selective} frame selective
labelling the same way under i.i.d.\ sampling. Weighted conformal inference \parencite{tibshirani2019} extends
exchangeability to covariate shift with known likelihood ratios, and \textcite{jincandes2023weighted} show that
weighted conformal p-values need not be PRDS and give a selection procedure that is. Learn-then-Test
\parencite{angelopoulos2021ltt} and conformal risk control \parencite{angelopoulos2024crc} control a monotone risk
over a family of thresholds. The weighted FDR is from \textcite{bh1997weighted} and \textcite{genovese2006weighted}.
Our contribution is the choice of measure and the derivation of the weights from a sampling design: under the
dollar measure the likelihood ratio between the law of a PPS draw and the law of a random unsampled dollar is
the dollars a text stands for, which supports an idealised covariate-shift construction (it does not make the
actual fixed-size draws exchangeable with the unsampled texts), makes dollar-weighted error control natural and
leaves count control powerless; the conformal p-values are proved super-uniform under that idealised draw model,
the Learn-then-Test rule is evaluated and not certified because its effective-sample-size substitution has no
finite-sample justification under unequal weights, and both rules are checked on the committed design with
synthetic truths, including a truth whose expected error sits just above the target.

\paragraph{Annotation validity and measurement.}
Inter-prompt reliability has been framed as a measurement problem for LLM-based social-science labelling
\parencite{liu2026interprompt}; context sensitivity and drift are standing concerns \parencite{ziems2024}. We
measure instruction, context and version sensitivity on real administrative text with a designed sample and
dollar weights, and report them as sensitivity rather than as a general invariance.

\paragraph{The testbed's domain.}
\textcite{shekhtman2024} mapped philanthropic support of science from Form 990 data and predicted future ties with
Adamic--Adar on a static split; SOURCE \parencite{tan2026source} tags 990 grants to universities by topic. The
companion report \parencite{koomullil2026companion} built the database, the codebook, the four machine readings and
the human sample used here, and reported their disagreement as a spread. This paper's subject is the inference,
not the funding; the domain findings are stated only where the methods change them.

%% file: sections/problem.tex
\section{Measurement from shared evidence units}\label{sec:problem}

\subsection{Setting}
A population of \emph{records} $r \in \mathcal{R}$ (grant rows) each carries an amount $a_r \ge 0$ and a text
$t(r) \in \mathcal{T}$. The texts are the \emph{shared evidence units}: $|\mathcal{T}| = N \ll |\mathcal{R}|$, and the
bridge $t(\cdot)$ is many-to-one. A \emph{reading} is a function of the text alone, so a label assigned to text $i$
applies to every record $r$ with $t(r)=i$. Each text $i$ has a reference label $y_i \in \{R, N, U\}$ (the purpose
names research, names something else, or does not settle the question) given by human readers under a frozen codebook
(for a text outside the sample, $y_i$ is the label the same protocol would give it: three independent readings
under the codebook and, where they split on research or not, the readers' adjudication; the estimand is defined by
that protocol, not by what the money was spent on), a
vector of $K$ machine readings $\ell_i = (\ell_{i1}, \dots, \ell_{iK})$, and features $x_i$ (the readings, the
text, its dollars). Two amounts attach to a text. Its \emph{sampling size} $a_i$ is the sum of all external dollars
on its records, paid or committed, the size measure of the human sample's design. For a domain of records
$\mathcal{D} \subseteq \mathcal{R}$ (here: paid grants, tax years 2020--2024, research-capital rows excluded) its
\emph{dollar mass} is
\begin{equation}
d_i = \sum_{r \in \mathcal{D}:\, t(r) = i} a_r , \qquad d_{i,y} = \sum_{r \in \mathcal{D}:\, t(r)=i,\ \mathrm{year}(r) = y} a_r ,
\end{equation}
and the estimand is
\begin{equation}
T = \sum_{i=1}^{N} d_i\, \ind{y_i = R}, \qquad T_y = \sum_i d_{i,y}\,\ind{y_i = R} .
\label{eq:estimand}
\end{equation}
$T$ is the dollars whose stated purpose is research under the codebook. It is a count of reported amounts by
reference label; it is not an estimate of research expenditure, and the probabilities below are about the reference
judgement, not about how money was spent. Table~\ref{tab:notation} collects the notation.

\begin{table}[tbp]\centering\small
\caption{Notation. Sampling probabilities are $\pi$; reader confusion matrices in Section~\ref{sec:aggregation}
are written $\Pi_k$ to keep the two apart.}\label{tab:notation}
\begin{tabular}{lp{0.72\textwidth}}
\toprule
$i = 1..N$; $S$, $n$ & texts (shared evidence units); the human sample and its size \\
$a_i$; $d_i$; $d_{i,y}$ & sampling size (all external dollars); domain dollars; domain dollars in tax year $y$ \\
$y_i \in \{R,N,U\}$; $\ell_{ik}$; $x_i$ & reference label; reading of machine reader $k$; features \\
$s(i)$; $D_s$, $A_s$; $n_s$ & vote-pattern stratum; its domain dollars and sampling size; its PPS draws \\
$\pi_i$; $w_i = 1/\pi_i$; $I_i$ & inclusion probability; design weight; inclusion indicator \\
$f_i$; $e_i = \ind{y_i = R} - \lambda f_i$ & a prediction of $\ind{y_i=R}$; the residual at $\lambda$ \\
$m_i = w_i d_i$ & dollars text $i$ stands for in the population (its likelihood-ratio weight under the dollar measure) \\
$\hat y_i$, $c_i$, $s_i = 1 - c_i$ & machine label, its confidence, its score \\
$q_i$, $b$ & second-phase (review) inclusion probability; review budget in texts \\
\bottomrule
\end{tabular}
\end{table}

\subsection{The human sample}
Human labels exist on a sample $S \subset \{1,\dots,N\}$ drawn by a known design: texts are grouped into strata
$s(i)$ defined by the three LLM readers' vote pattern (six strata, from ``all three say research'' to ``all three
say not research''); within each stratum a fixed number $n_s$ of texts is drawn by systematic PPS on $a_i$ in a
random order, texts whose size makes their inclusion probability exceed one are taken with certainty, and the texts
of the \valNTakeAll{} largest research grants and commitments are taken in full. Every text has a known inclusion
probability $\pi_i \in [0, 1]$ (zero for the \csNUnsampledPiZero{} texts without positive sampling size) and the
sampled texts carry design weights $w_i = 1/\pi_i$. For a PPS draw $\pi_i = n_s a_i / A_s^{\circ}$, with
$A_s^{\circ}$ the sampling size of the stratum's non-certainty texts, so $w_i a_i = A_s^{\circ}/n_s$ is constant
within a stratum, whereas $w_i d_i$ is not: the domain is narrower than the sampling frame on
\csPpsDrawsZeroDomain{} PPS draws with no domain dollars at all. Table~\ref{tab:reconciliation} reconciles the two
amounts by stratum. Because $s(i)$ is a function of the machine readings and the PPS size is $a_i$, the design
depends on $x_i$ but, given $x_i$, not on $y_i$ (Assumption~\ref{as:ign}).

\begin{table}[tbp]\centering\small
\caption{Frame and sample by stratum: texts, sampling size $A_s$ and domain dollars $D_s$ (\$B), texts with no
domain dollars, sample composition, the inclusion-probability range of the PPS draws, and the dollars each PPS draw
stands for under the sampling size ($w a$, constant) and under the domain ($w d$, mean).}\label{tab:reconciliation}
\adjustbox{max width=\textwidth}{\input{tables/t_reconciliation}}
\end{table}

\subsection{Three tasks and the data roles}
\begin{figure}[tbp]\centering
\begin{tikzpicture}[font=\scriptsize, node distance=4mm and 6mm, box/.style={draw, rounded corners=2pt, align=center, inner sep=3pt, minimum height=7mm, text width=40mm}, arr/.style={-{Latex[length=1.6mm]}, thick}]
\node[box, fill=gray!8] (pop) {Population: $N$ texts\\$K=4$ machine readings, $d_i$, $a_i$};
\node[box, right=of pop, fill=gray!8] (samp) {Human sample $S$ ($n$ texts)\\stratified PPS by $a_i$, take-all, $\pi_i$};
\node[box, below=of pop, fill=blue!6] (ds) {Dawid--Skene, DS-mix\\(unsupervised, all $N$)};
\node[box, below=of samp, fill=blue!6] (cal) {Calibrator (10 frame folds, fitted on $S$)\\$f_i$ from the model of text $i$'s fold, sampled or not};
\node[box, below=of cal, fill=green!7] (t2) {T2 accept rules\\nulls and risks from $S$};
\node[box, left=of t2, fill=green!7] (t1) {T1 dw-PPI\\rectify $\sum d_i f_i$ on $S$};
\node[box, right=of t2, fill=green!7] (t3) {T3 review policies\\replay on $S$};
\node[box, below=of t1, fill=yellow!10] (c1) {check: \csSimDraws{} re-draws of\\the design, calibrator refit};
\node[box, below=of t2, fill=yellow!10] (c2) {check: nested held-out folds of $S$;\\re-draws with synthetic truths};
\node[box, below=of t3, fill=yellow!10] (c3) {check: calibrator on $\tfrac23 S$,\\policies on the other $\tfrac13$};
\draw[arr] (pop) -- (ds); \draw[arr] (pop) -- (samp); \draw[arr] (samp) -- (cal); \draw[arr] (ds) -- (cal);
\draw[arr] (cal) -- (t1); \draw[arr] (cal) -- (t2); \draw[arr] (cal) -- (t3);
\draw[arr] (t1) -- (c1); \draw[arr] (t2) -- (c2); \draw[arr] (t3) -- (c3);
\end{tikzpicture}
\caption{Data flow and label use. The human labels enter the calibrator (cross-fitted), each task's rectifier or
rule, and each task's check; no text's own label enters its own probability.}\label{fig:overview}
\end{figure}
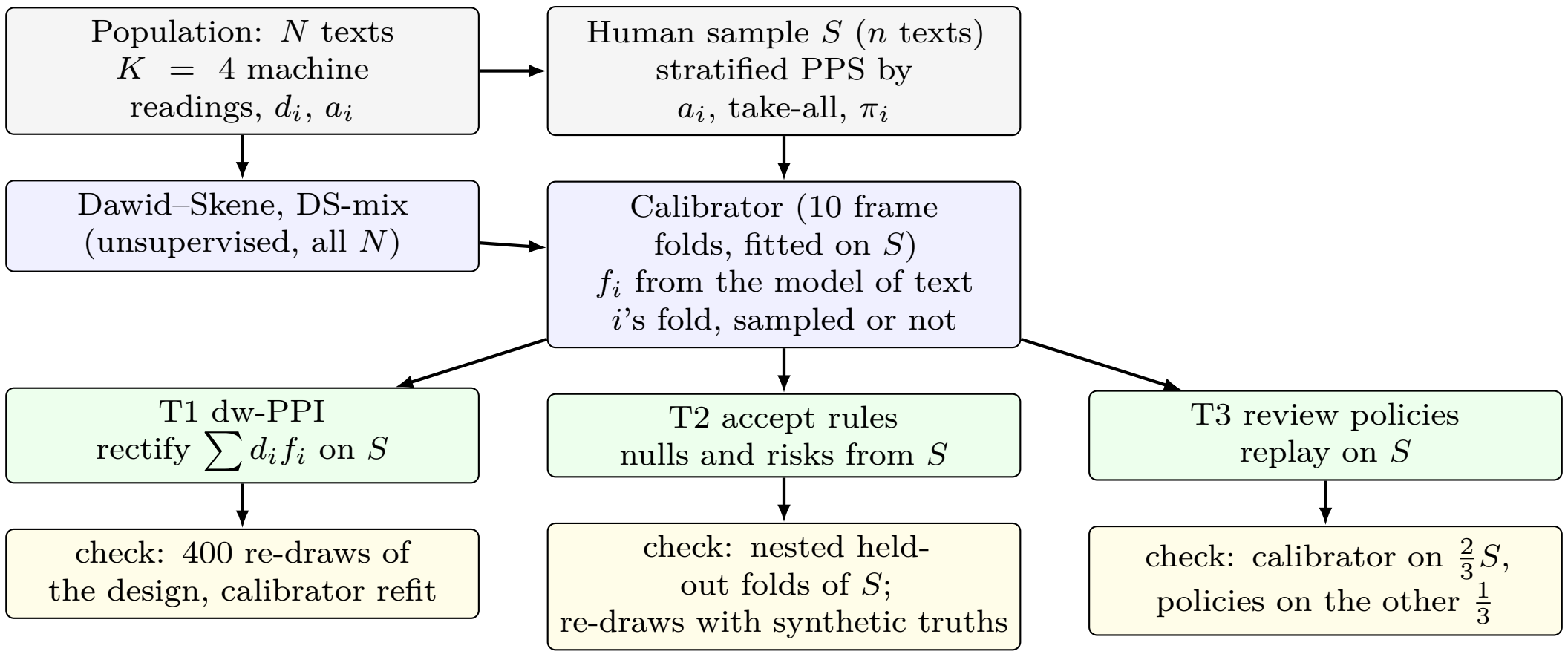

\begin{description}[leftmargin=1.2em]
\item[T1, valid estimation.] An estimator of $T$ (and of $T_y$, of growth $T_{2024}/T_{2020}-1$) that uses all $N$
  machine readings and the $n = |S|$ human labels and is design-unbiased for $T$ whatever the readings' bias:
  exactly for a fixed predictor and, under Poisson sampling, for the cross-fitted construction of
  Section~\ref{sec:ppi}; approximately under the committed fixed-size design, for the Hájek form and for the
  by-year and growth ratios, with an interval whose coverage is checked by simulation.
\item[T2, selective labelling.] A rule that decides, for each unlabelled text, whether its machine label may be
  accepted, whose aim is that the share of accepted \emph{dollars} carrying a wrong label stays below a chosen
  $\alpha$ (in expectation, or with a stated probability), and that reports how many texts and dollars remain for
  a person; Section~\ref{sec:selective} says for which rule and under which experiment that aim is proved, and for
  which it is only measured.
\item[T3, budgeted acquisition.] Given a budget of $b$ texts a person can read, which texts to read so that the
  estimator of $T$ built from the revealed labels and the machine readings has the smallest error, while staying
  valid.
\end{description}
Each task uses a per-text probability $f_i \approx P(y_i = R \mid x_i)$, which Section~\ref{sec:aggregation}
builds first; the validity of T1 does not need it to be calibrated, its efficiency and the rules and designs of T2
and T3 do. Figure~\ref{fig:overview} shows which labels enter which fit, and Table~\ref{tab:predspec} states, for
every task, which labels train and tune the predictor and which model predicts each text: the calibrator is
cross-fitted on folds assigned to the whole frame by a seeded random permutation within each vote stratum, made
without reference to sample membership or labels and held fixed (every sampled text's probability is out-of-fold,
and every unsampled text is predicted by the model of its own fold); the estimators of T1 use that one prediction
vector; the rules of T2 are calibrated on $S$ and evaluated on held-out folds of $S$, with the calibrator refitted
and its strength re-chosen inside each split, and on fixed-score re-draws of the design; the policies of T3 are
replayed on a third of $S$ whose labels enter none of the calibrator's coefficients, or with every label charged;
in both replays the ridge strength alone is a constant taken from tuning on the full sample
(Section~\ref{sec:adaptive}).

\begin{table}[tbp]\centering\small
\caption{Prediction specification by task. Features are the same \csNFeatures{} label-free features throughout
(the machine readings, the unsupervised posteriors fitted on all $N$ readings, and the text cues); the ridge
multinomial logistic calibrator is the model in every case. The feature scaler is fitted on the whole frame for T1
and T2, where it is label-free and fixed before any sample is drawn, and on the texts the policy has read for the
T3 replays with every label charged (Section~\ref{sec:acquisition}), where the scaler and the regression
coefficients use only the labels the policy has bought. The ridge strength of the T3 replays is the one exception:
it is a constant taken from the first version's tuning on the full sample, and Appendix
Table~\ref{tab:adaptive_ridge} repeats the headline design with the strength chosen on the policy's own
first-phase labels.}\label{tab:predspec}
\begin{tabular}{p{0.2\textwidth}p{0.36\textwidth}p{0.36\textwidth}}
\toprule
Task & training and tuning labels & prediction of a text \\
\midrule
T1 estimation (Section~\ref{sec:ppi}) & the sampled texts outside the text's frame fold; ridge strength chosen by
inner 5-fold cross-validation within those texts & the model of the text's own frame fold, whether or not the
text was sampled \\
T2 held-out evaluation (Section~\ref{sec:selective}) & in each of 100 splits, the four calibration folds of $S$
only; strength re-chosen on them; nulls and risk estimates from them & calibration texts: cross-fitted within the
calibration folds; held-out fold and unsampled population: one model fitted on all calibration folds \\
T2 stress tests (Appendix~\ref{app:stress}) & none refitted: the T1 calibrator's scores are held fixed
(fixed-score tests) & as T1 \\
T3 protected replay (Section~\ref{sec:acquisition}) & two thirds of $S$ by stratum; strength a constant, $C = \csRidgeCMain$, not tuned in the replay &
one model fitted on the two thirds scores the replay third \\
T3 designs with every label charged (Section~\ref{sec:adaptive}) & only the labels the policy has bought so far;
strength the same constant (taken from the first version's full-sample choice; every value of the grid in
Appendix Table~\ref{tab:adaptive_ridge}); constant mean label below five labels & one model fitted on the bought labels scores
every unread text before the next wave is drawn \\
Institutions (Section~\ref{sec:downstream}) & the T1 calibrator, held fixed in the simulation & as T1 \\
\bottomrule
\end{tabular}
\end{table}

\begin{assumption}[design ignorability]\label{as:ign}
Inclusion in $S$ depends on $(x_i)_{i \le N}$ only: $P(S \mid x, y) = P(S \mid x)$.
\end{assumption}
\noindent The assumption holds exactly for the committed design, whose only inputs are the vote pattern and the
sampling size. The vote pattern was computed under codebook version 1.1 while the human read under version 1.2;
this does not matter, because the design is a function of $x$ whatever version produced $x$. The assumption fails
if a person's willingness to label depends on the label, which the blind workbook design (the reader sees no
vote, stratum or pipeline label) rules out; every sampled text was labelled.

%% file: sections/testbed.tex
\section{Testbed}\label{sec:testbed}

The data are those of the companion report \parencite{koomullil2026companion}: every grant that US private foundations and
grantmaking public charities reported to US colleges and universities on Form 990-PF and Form 990 Schedule~I
\parencite{irs_pf,irs_schi} in the IRS e-file series, tax years 2020--2024, with recipients resolved to IPEDS
institutions \parencite{ipeds} and their affiliates, and funders typed so that internal transfers are set aside.
The purpose-text population is the set of unique normalised purpose texts on external grants, paid or committed,
in that database: $N = \csNTexts$ texts covering \csNRows{} grant rows and \$\csFrameB{}B of sampling size. The
headline domain (paid, 2020--2024, research-capital rows excluded) holds \$\csExcapB{}B on \csTextsExcap{} of those
texts; the other \csTextsZeroDomain{} texts have no domain dollars and contribute nothing to $T$. The companion's
rule-based pipeline assigns \$\csExplicitB{}B of the domain to an explicit research purpose.

\paragraph{Readers.} Four machine readings cover every text under one frozen codebook (version~1.1 for the population
runs; the readers and the sensitivity runs used version~1.2, which differs only in when a purpose type is required,
not in the status rules; both versions are released with this paper's record, Section~\ref{sec:release}): the pipeline's rule-based classifier (``Rules''), and three language models prompted with the
codebook verbatim and the grant's context (funder, recipient as filed, tax year, amount): Claude Sonnet~5.5, Gemini
2.5~Flash (thinking off) and GPT-5~mini (reasoning low), run in batch on 4~October 2026 with stored prompts and
raw answers. Each returns a status in $\{R, N, U\}$, a purpose type, the text's atomic
purposes with their own statuses, and a one-sentence rationale. The three LLMs disagree on the status of
\csLLMDisagreeTexts{} of texts carrying \csLLMDisagreeDollars{} of dollars.

\paragraph{Human readings.} There are three readers and five reading occasions, each blind to every machine
output and to the other readers, followed by a joint adjudication of the texts on which the readers split. Reader~1 labelled all $n = \csNSample$ sampled texts in October
2026: \csSampleR{} research, \csSampleN{} not research, \csSampleU{} unclear; some weeks later reader~1 re-read
\csNRecheck{} texts blind (status identical on \valSelfSame{}). Reader~2
labelled a within-stratum random subset of \csNReaderTwo{} texts in October 2026 and the remaining
\csNReaderTwoRest{} in November 2026 (over all \csNReaderTwoFull: \csSampleTwoR{} research, \csSampleTwoN{} not
research, \csSampleTwoU{} unclear). Reader~3 labelled all \csNReaderThree{}
texts in November 2026 (\csSampleThreeR, \csSampleThreeN, \csSampleThreeU). Readers~2 and~3 received the
codebook, the workbook in a fresh random order with the funder, recipient, year and amount, and a one-page cover
sheet; they worked independently of one another with no instruction beyond the codebook and the cover sheet, and
their labels are first answers and are kept as given; the readers are not identified. The texts on which the
three split were adjudicated afterwards, as described below.
Table~\ref{tab:readers} measures agreement. On research-or-not the three readers give the same answer on
\csUnanBinaryTexts{} of texts and \csUnanBinaryWd{} of dollars (pairwise \csAgBinaryTextsMin{}--\csAgBinaryTextsMax{}
of texts, Cohen's $\kappa$ \csAgKappaBinaryMin{}--\csAgKappaBinaryMax); on the three-way status on
\csUnanStatusTexts{} of texts (pairwise $\kappa$ \csAgKappaStatusMin{}--\csAgKappaStatusMax, Fleiss's
$\kappa = \csAgAllKappaStatus$). The design-weighted shares by text rest on a few small-dollar texts with weights in
the thousands: readers~2 and~3 give the same research-or-not answer on the heaviest of them and reader~1 differs
from both on some, which is why that pair's design-weighted agreement (\csAgTwoThreeBinaryW) exceeds its share of
texts (\csAgTwoThreeBinaryTexts) while the other pairs' do not. The readers differ most in how often they say ``unclear'': \csUShareOne{} of
texts for reader~1, \csUShareTwo{} for reader~2 and \csUShareThree{} for reader~3; where reader~1 says unclear,
reader~3 says not research on \csReaderOneUIsThreeN{} and research on \csReaderOneUIsThreeR{} (design-weighted).
Reader~3 calls the most texts research (\csRShareThree{} against \csRShareOne{} and \csRShareTwo), which is
\csRShareWdThree{} of dollars against \csRShareWdOne{} and \csRShareWdTwo. Agreement is near complete in the
stratum the models unanimously call not research and lowest in the stratum they unanimously call research, where
the readers split on bare field and programme names (Appendix Table~\ref{tab:readers_stratum}). The first version of this paper took reader~1's first pass as the reference; this version takes the
adjudicated reading defined below.

\begin{table}[tbp]\centering\small
\caption{Agreement among the three readers on the \csNSample{} sampled texts: the share of texts with the same
three-way status and the same answer on research or not, by text, design-weighted ($w_i = 1/\pi_i$, with a
\csReadersBoot-replicate stratified bootstrap interval) and dollar-weighted ($w_i d_i$); Cohen's $\kappa$ for
pairs and Fleiss's $\kappa$ for all three.}\label{tab:readers}
\input{tables/t_readers}
\end{table}

\paragraph{Adjudication of the split texts.} The \csNSplit{} texts on which the three readers do not give the same
answer on research or not were adjudicated in November 2026, after the first version of this paper. The
adjudication workbook listed them in a fresh random order with the context the readers saw (funder, recipient,
year, amount and the text), the codebook, and the three first labels in an order shuffled on every row, so that a
column does not identify a reader; it showed no machine output. One status was recorded for each text:
\csAdjStatusR{} research, \csAdjStatusN{} not research and \csAdjStatusU{} unclear
(Table~\ref{tab:adjudication}). \csAdjudicatorSentence{} The adjudication was not blind to the first labels, no reasons
were recorded, and it covered the split texts only: the \csNNotSplit{} texts on which
the readers agree on research or not were not re-opened, including the \csNStatusOnlySplit{} on which they differ
between not research and unclear. The \emph{adjudicated reading} used below is the adjudicated status on the
\csNSplit{} texts and the readers' shared answer on the rest, with their majority status where they agree on
research or not but not on the status (\csSampleAdjR{} research, \csSampleAdjN{} not research and
\csSampleAdjU{} unclear over the sample). It is not a fourth independent reading: it is built from the first
labels and was made by the readers with those labels in view. The first labels are unchanged, and every agreement
figure above is computed from them.

\emph{The adjudicated reading is the reference throughout this paper}: every estimate, rule and design below takes
it as $y_i$. It replaces reader~1's first pass, the reference of the first version: the adjudicated reading uses
all three readings and, where they differed, an answer the three readers agreed on. It remains one operational
measurement of the codebook. Section~\ref{sec:ppi} re-runs the reference-dependent estimates under each reader and
under their majority (the simulations on synthetic truths are held fixed), and Section~\ref{sec:discussion} says what the change of reference moved and what the
adjudication does not settle.

The adjudication did not simply follow the majority. Where two of the three readers had said research it kept
research on \csAdjTwoSayRAdjR{} of \csAdjTwoSayR{} texts, and where only one had it still chose research on
\csAdjOneSaysRAdjR{} of \csAdjOneSaysR. On research or not it agrees with the majority of the first labels on
\csAdjMajSame{} of the \csAdjTexts{} texts, moving \csAdjMajToR{} to research and \csAdjMajToNotR{} away from
it, and with the individual readers on \csAdjSameThreeN{} (reader~3), \csAdjSameOneN{} (reader~1) and
\csAdjSameTwoN{} (reader~2). It calls \csAdjSplitRWd{} of the split texts' design-weighted dollars research,
against \csAdjSplitMajRWd{} under the majority. The adjudicated reading is therefore closest to reader~3, who
called the most texts research; over the whole sample it has the same answer on research or not as reader~3 on
\csAdjSameThreeAllTexts{} of texts and as readers~1 and~2 on \csAdjSameOneAllTexts{} and
\csAdjSameTwoAllTexts.

\begin{table}[tbp]\centering\small
\caption{Adjudication of the \csNSplit{} texts on which the three readers split on research or not: the
adjudicated status by the pattern of the three first labels, and each pattern's share of the split texts'
design-weighted dollars.}\label{tab:adjudication}
\adjustbox{max width=\textwidth}{\input{tables/t_adjudication}}
\end{table}

\paragraph{What the companion reported.} On this domain the four readers' own totals span
\$\csReaderTotalMinB{}--\csReaderTotalMaxB{}B; the design estimate from reader~1 alone is \$\csRerunOneClassicB{}B
[\csRerunOneClassicLoB, \csRerunOneClassicHiB]; five-year growth under the readings spans \csGrowthReadersMin{}--\csGrowthReadersMax{}
against \csFedGrowth{} for federally financed university R\&D expenditure \parencite{nsf26305}. The companion's
conclusion was that the spread ``describes sensitivity to the reading, not a bound''. The rest of this paper
replaces the spread with decisions whose properties are stated, and checked where they cannot be proved.

%% file: sections/aggregation.tex
\section{Calibrated aggregation of disagreeing readers}\label{sec:aggregation}

\subsection{Models}
\paragraph{Dawid--Skene (DS).} Each text has a latent class $c_i \in \{R, N, U\}$ with prior $p_c$; reader $k$
reports $\ell_{ik}$ with probability $\Pi_k[c_i, \ell_{ik}]$. The $K = 4$ readers are conditionally independent
given $c_i$. EM on the $N \times 4$ label matrix converges in a few iterations; the three LLMs show all
\csNLLMPatterns{} possible compositions of three votes. The posterior $P(c_i = R \mid \ell_i)$ is the ``DS
posterior''. A variant with a latent item state $z_i \in \{\text{easy}, \text{hard}\}$ that switches every reader's
confusion matrix, the item-difficulty idea of GLAD \parencite{whitehill2009glad,paun2018annotation}, lets the
readers be dependent given $c_i$ (``DS-mix''); its held-out log-likelihood per text is \csDsHeldoutTwo{} against
\csDsHeldoutOne{} for DS, and its hard-state share is \csDsmixHardShare{} of texts. A two-state mixture with four
readers and three classes is not guaranteed identifiable, and the gain says only that the fitted distribution is
closer, not that the hard state is difficulty; we use DS-mix as a feature and claim nothing else for it.

\paragraph{Calibrator.} Neither posterior is anchored to the reference: an unsupervised model recovers at best the
readers' consensus, and \csDsPriorR{} of texts are research under the DS prior. We therefore fit a ridge
multinomial logistic regression of the reference label on \csNFeatures{} features (a stacked model in the sense of
\textcite{wolpert1992stacking}): the two posteriors, each reader's status, each LLM's purpose type and the share of
its atoms labelled research or unclear, and \csNTextFeatures{} text features (length, digits, capitalisation, rows
covered, dollars, and eleven keyword cues such as ``research'', ``scholarship'', ``fellow'', ``support'',
exempt-purpose boilerplate). The cues were written from the codebook's rule categories before any model was fitted
and were not selected by their fit to the sample. The ridge strength is chosen inside each outer training set by an
inner cross-validation on the dollar-weighted log-loss of $P(R)$ (10 outer frame folds, 5 inner), so no outer
label touches the choice; the modal choice is $C = \csCfinal$. The fit is unweighted, a working choice: under
Assumption~\ref{as:ign} the conditional law $P(y \mid x)$ is the same on the sample and the population, because
the features include the vote pattern and the dollars that drive the design, so an unweighted fit is consistent
for it when the model is correctly specified; under misspecification a design-weighted or dollar-weighted fit
would target a different population approximation, and we did not compare the three fitting objectives (the
inner validation is dollar-weighted, which steers the choice of strength but not the objective). All metrics are
design-weighted (``texts'') and design$\times$dollar weighted (``dollars''), and every probability on a sampled
text is out-of-fold. Log-loss and dollar AUC carry a 95\% interval in Table~\ref{tab:calibration} from \csBBoot{}
within-stratum bootstrap resamples of the sampled texts (the out-of-fold predictions are resampled as they
stand, without refitting the models, so these intervals and the paired ones of Tables~\ref{tab:paired}
and~\ref{tab:ablation} describe the scoring of fixed predictions on this sample and are not the uncertainty of
repeated training); intervals for ECE and text AUC are in Appendix
Table~\ref{tab:calibration_intervals}. Every population text is predicted by the model of its own frame fold, the
same rule as for the sampled texts (Section~\ref{sec:ppi}): a cross-fitted predictor in the sense of \textcite{zrnic2024cross}, with
the prediction of a text never depending on whether it was drawn. Hard labels are scored as probabilities clipped
at $\epsilon = \csEpsHard$. ECE uses bins cut at unweighted deciles of the predicted probability (about 65 texts
each); the weights enter within bins, and predictors with few distinct values have fewer bins (two for a hard label,
five for the vote share), so ECE is not strictly comparable between hard labels and continuous predictors.

\paragraph{What each baseline and each ablation may use.} A single-reader calibrator (``Claude Sonnet~5.5 +
text'' and its like) has that reader's own columns and the text cues and neither posterior, so nothing from the
other three readers reaches it: it is what an office calling one model would have. DS with isotonic recalibration
uses the posterior fitted on all four readers' votes, so it is a simpler calibration of the same panel, not a
smaller panel. The ablations of Table~\ref{tab:ablation} are of two kinds. Removing a reader's own columns
deletes its status and, for a language model, its purpose type and atom shares, and keeps the two posteriors,
which were fitted on all four readers and still carry its votes: this measures what the reader's own columns add
given the full panel's posteriors, not what its calls are worth. Removing a reader entirely also refits both
posteriors on the other three readers' votes over the whole population, so that no channel from the removed
reader remains; that is the calibrator an office would have without calling that reader.

\subsection{Results}
\begin{figure}[tbp]\centering
\includegraphics[width=0.98\textwidth]{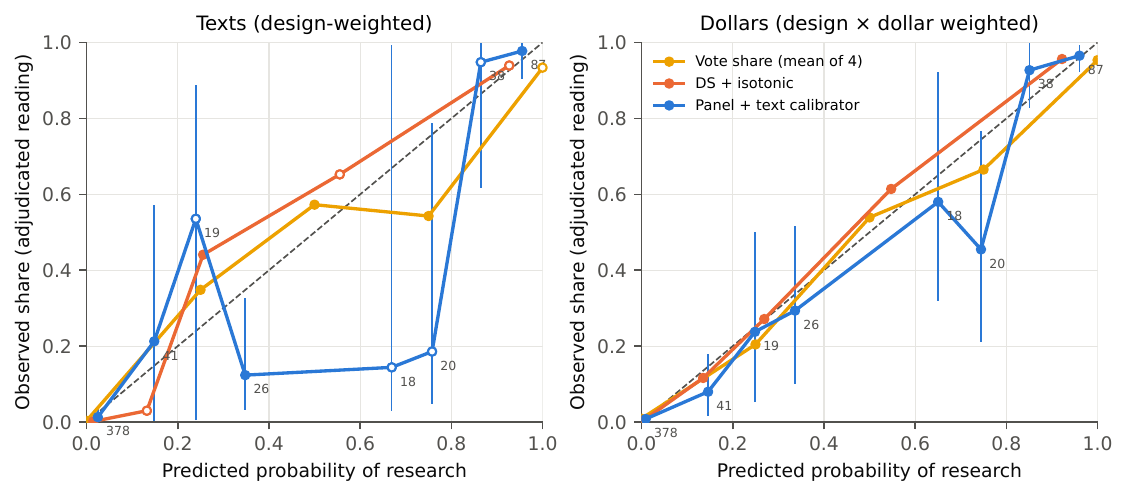}
\caption{Reliability diagrams on the human sample, by texts (left) and dollars (right): the mean of the four hard
readings (vote share), DS with isotonic recalibration, and the calibrated panel (out-of-fold). Bins are ten
equal-width intervals of the predicted probability, each method binned on its own predictions; a point is plotted at
the bin's weighted mean prediction and weighted share labelled research by the reference. Bins with fewer than 15 texts
(unweighted) are not drawn, and the numbers beside the panel calibrator are those unweighted counts. Vertical bars
are 95\% intervals from the \csBBoot{} within-stratum bootstrap resamples of Table~\ref{tab:calibration}, with bins
held fixed; hollow markers have a Kish effective size \parencite{kish1965} below 5. These bins differ from the decile bins used for ECE in
Table~\ref{tab:calibration}.}
\label{fig:reliability}
\end{figure}

Table~\ref{tab:calibration} compares hard readings, unsupervised posteriors, calibrated single readers and
calibrated panels against the reference. Three things stand out. First, hard labels scored as probabilities are
dominated by the clipping constant (a wrong confident label costs $-\log \epsilon$), so the fair comparison is
among calibrated models: DS with isotonic recalibration reaches a dollar log-loss of \csIsoLLdollar{} and the
vote share \csIsoVoteLLdollar; the full calibrator reaches \csPanelLLdollar{} [\csPanelLLdollarLo,
\csPanelLLdollarHi] with a dollar AUC of \csPanelAUCdollar. Second, the text-weighted metrics rest on a few small-dollar texts that carry design weights in the thousands
(the full calibrator's text log-loss is \csPanelLLtext{} with interval [\csPanelLLtextLo, \csPanelLLtextHi];
with reader~1 as the reference, whose label differs from the models' on two of the heaviest of those texts, it
is \csRerunOneLL); the estimand's measure is the dollar one, and we rank models by it. Third, and this is the finding
that matters for the method, on dollar log-loss the full panel shows no resolved improvement over two simpler
baselines: a single-model baseline, Claude Sonnet~5.5 with the text cues, and a simpler recalibration of the
same panel, the unsupervised posterior recalibrated on the sample, which still needs every reader's votes and
saves no model call. The paired bootstrap difference in dollar log-loss between the full calibrator and Claude Sonnet~5.5
with text is \csPairSonnetDollar{} [\csPairSonnetDollarLo, \csPairSonnetDollarHi] (Table~\ref{tab:paired}) and
against DS with isotonic \csPairIsoDollar{} [\csPairIsoDollarLo, \csPairIsoDollarHi], intervals that contain
zero (as does the one against GPT-5~mini with text, \csPairMiniDollar{} [\csPairMiniDollarLo,
\csPairMiniDollarHi]), against \csPairTextOnlyDollar{} [\csPairTextOnlyDollarLo, \csPairTextOnlyDollarHi] for
the text cues alone. The statement is specific to that metric and those baselines. The same table shows a
resolved difference in dollar log-loss against Gemini~2.5 Flash with text (\csPairFlashDollar{}
[\csPairFlashDollarLo, \csPairFlashDollarHi]) and a small advantage in dollar AUC, a ranking measure, over DS
with isotonic (\csPairIsoAuc{} [\csPairIsoAucLo, \csPairIsoAucHi]). These are exploratory paired comparisons
with no adjustment for their number, and an interval that contains zero is not a demonstration of equivalence,
for which a margin would have to be chosen in advance, nor a bound on what the extra readers contribute: it says
that \csNSample{} texts do not resolve the difference. Only the single-model baseline would save the other
readers' calls; we did not cost the calls, and Section~\ref{sec:acquisition} does not test whether a single-model
calibrator would serve the review design as well as the panel does.
The comparison with the unsupervised posterior depends on the reference: with reader~1's labels, the reference of
the first version, the same difference was \csPairIsoDollarOne{} [\csPairIsoDollarOneLo, \csPairIsoDollarOneHi],
an interval that excludes zero. A supervised calibrator adds to the machine readers' consensus to the extent
that the reference sits away from that consensus, and the adjudicated reading sits closer to it than reader~1
does. The ablations
(Table~\ref{tab:ablation}) are exploratory and agree: removing the unsupervised posteriors changes the text
log-loss from \csPanelLLtext{} to \csAblDsLL; removing the text features changes the dollar log-loss most
(\csAblTextDollarLL{} against \csPanelLLdollar, paired difference [\csAblTextDollarLo, \csAblTextDollarHi]);
removing the LLM purpose types and atoms (\csAblAtomsDollarLL, [\csAblAtomsDollarLo, \csAblAtomsDollarHi])
moves it by less than \csAblOtherBelow{} in the point estimate, and so does removing any one of the four
readers, in either of the two senses defined above: with the reader's own columns deleted and
the posteriors kept, or with the reader removed entirely and both posteriors refitted on the other three, where
the dollar log-loss rises by \csAblFullMinDollar{} to \csAblFullMaxDollar{} across the four readers (most for
\csAblFullWorst). Removing Claude Sonnet~5.5 entirely moves the text log-loss most, from \csPanelLLtext{} to
\csAblFullSonnetLL. Under each reader and their majority the full removals move the dollar log-loss by between
\csRerunAblFullMin{} and \csRerunAblFullMax. Every one of these paired intervals, under every reading, reaches zero, and we draw no significance from them: at
the precision of \csNSample{} texts no single reader is shown to be necessary for dollar log-loss, which is the
single-model comparison above seen from the other side, and no stronger than it. In the reliability
diagrams (Figure~\ref{fig:reliability}), where all four readings say research the reference labels
\csRelVoteTopObsTexts{} of the texts and \csRelVoteTopObsDollars{} of the dollars research; against reader~1
these shares were \csRelVoteTopObsTextsOne{} and \csRelVoteTopObsDollarsOne, and the vote share looked
over-confident at the top. \csRelPanelDollarSentence{} By texts the points are too imprecise to
assess: the design weights leave Kish effective sizes between \csRelPanelTextNeffMin{} and
\csRelPanelTextNeffMax{} texts per bin. The three-class task is harder: ``unclear'' is the difficult class (AUC for U of \csPanelAUCU{} for
the full calibrator, panel and text cues, and \csGbmAUCU{} for gradient boosting; Appendix Table~\ref{tab:threeclass}), which is why
the estimand and the accept rules below are stated for research-or-not.

\begin{table}[tbp]\centering\small
\caption{Readers, posteriors and calibrators against the reference (the adjudicated reading) on the \csNSample-text sample. Texts: design-weighted;
dollars: design $\times$ dollar weighted; intervals from \csBBoot{} within-stratum bootstrap resamples. Calibrators
are out-of-fold with the ridge strength chosen by inner cross-validation. Lower log-loss and ECE and higher AUC are
better.}\label{tab:calibration}
\adjustbox{max width=\textwidth}{\input{tables/t_calibration}}
\end{table}

\begin{table}[tbp]\centering\small
\caption{Paired bootstrap differences, full calibrator minus baseline (negative is better for log-loss, positive for
AUC), with 95\% intervals.}\label{tab:paired}
\adjustbox{max width=\textwidth}{\input{tables/t_paired}}
\end{table}

\begin{table}[tbp]\centering\small
\caption{Ablation of the full calibrator (out-of-fold; the ridge strength re-chosen inside each fit). The interval
is the paired bootstrap difference in dollar log-loss against the full model. A reader's own columns are its
status and, for a language model, its purpose type and atom shares. In the first block the two Dawid--Skene
posteriors are those of the full model, fitted on all four readers, so the removed reader's votes still reach the
calibrator through them; in the second block both posteriors are refitted on the other three readers' votes, so
no channel from the removed reader remains.}\label{tab:ablation}
\input{tables/t_ablation}
\end{table}

\paragraph{The machine readers differ from the reference together.} Table~\ref{tab:dependence} measures how
often the machine readers' answers differ from the reference on the same texts, with errors defined relative to
the reference and pooled over the \csNSample{} sampled texts. This is a pooled association: it motivates checking
the aggregation methods empirically, and it is not a test of the conditional independence that Dawid--Skene
assumes, which is stated given the latent class. All three
language models give an answer on research or not that differs from the reference on \csWtAdjN{} of those texts
(\csWtAdjTexts{} of the sampled texts, carrying \csWtAdjWd{} of design-weighted dollars), against
\csWtAdjIndepTexts{} of texts if their errors were independent, and \csWtAdjOfAnyTexts{} of the \csWtAdjAnyN{}
texts on which any of the three differs have all three differing. The errors of any two language models have a
correlation of \csDepAdjCorrTextsMin{}--\csDepAdjCorrTextsMax{} over the sampled texts and
\csDepAdjCorrWdMin{}--\csDepAdjCorrWdMax{} by dollars. The count depends on the reading the models are held
against (Appendix Table~\ref{tab:wrong_together}): it is \csWtNMin{} to \csWtNMax{} texts under the three
readers and their majority. Under reader~1 it is \csWtOneN, and on \csWtOneOthersSide{} of those texts at least
one of the other two readers gives the models' answer (both on \csWtOneOthersBoth). Part of what counts as
the models' joint error under one reader is that reader's distance from the others; under every reading the three
models still differ from it together far more often than independent errors would. We do not summarise the
dependence by the design-weighted share of texts, as the first version of this paper did. The design weights
reach the thousands on a few small-dollar texts: under reader~1 two texts with weights above
\csWtOneTopTwoWeightMin{} carried \csWtOneTopTwo{} of the design-weighted mass of the \csWtOneN{} texts on which
all three models differed from reader~1 (not of the whole sample's), and the share of \csWtOneW{}
reported there (pairwise correlations \csDepOneCorrWMin{}--\csDepOneCorrWMax) falls to \csWtOneWNoTopTwo{}
(\csDepOneCorrWNoTopMin{}--\csDepOneCorrWNoTopMax) without them; readers~2 and~3 gave the models' answer on
\csWtOneTopTwoOthers{} of the two. This is marginal dependence of pooled errors relative to one reading, with no
features. A gold-free multiple-measurement estimator \parencite{egamishin2026dmm} assumes independence given the
latent label and the unit's features, which this table does not test: readers that err independently given a
text's difficulty show the same pattern, and the reference is not the latent label. What the table does show is
what the simplest member of that family does here. The Dawid--Skene posterior with no human label gives a research
total of \$\csDsGoldFreeB{}B, which is \csDsGoldFreeVsRef{} the rectified interval of Section~\ref{sec:ppi}
under the adjudicated reference and was outside the interval under reader~1; the variant that lets the readers
be dependent, DS-mix, gives \$\csDsmixGoldFreeB{}B, \csDsmixGoldFreeVsRef{} it. Whether a gold-free total looks
biased therefore depends on the reading it is held against, and on the model, and the experiment does not isolate
the dependence as the cause. Agreement among language models is not a certificate of
correctness; it carries information, which the calibrator uses as a feature (the unsupervised panel's dollar AUC
is \csDsAUCdollar), and what anchors it is the human sample.

\begin{table}[tbp]\centering\small
\caption{Dependence of the machine readers' research-or-not errors relative to the reference, pooled over the
\csNSample{} sampled texts: by text, each sampled text counted once (a description of the sample, which
over-represents large-dollar texts), and by design-weighted dollars. For each pair: how often each differs from
the reference, how often both do, the share expected if they differed independently, and the correlation.}
\label{tab:dependence}
\input{tables/t_dependence}
\end{table}

\paragraph{Alternative references.} Scored without refitting against each reader's labels of the same
\csNSample{} texts, the model fitted to the adjudicated reading reaches log-loss \csSensReaderOneLL{} (AUC
\csSensReaderOneAUC) against reader~1, \csSensReaderTwoLL{} (AUC \csSensReaderTwoAUC) against reader~2,
\csSensReaderThreeLL{} (AUC \csSensReaderThreeAUC) against reader~3 and \csSensMajLL{} (AUC \csSensMajAUC)
against their majority, against \csSensRefLL{} (AUC \csSensRefAUC) on its own reference; refitted to reader~1's
labels with the blind second reading substituted where it exists, its log-loss is \csSensRecheckLL{} (AUC
\csSensRecheckAUC). Where one reader is outvoted by the other two on research or not (\csOutvotedOneN,
\csOutvotedTwoN{} and \csOutvotedThreeN{} texts for readers~1, 2 and~3), the model sides with the two on
\csOutvotedOneModel, \csOutvotedTwoModel{} and \csOutvotedThreeModel{} of those texts (\csOutvotedOneModelWd,
\csOutvotedTwoModelWd{} and \csOutvotedThreeModelWd{} dollar-weighted). Appendix
Table~\ref{tab:readers_machines} gives every machine reader's agreement with every reading: by text each
language model agrees with reader~3 about as often as with reader~1 (\csMachSonnetThreeT{} against
\csMachSonnetOneT{} for Sonnet), with reader~2 a little less and with the adjudicated reading most
(\csMachSonnetAdjT), and by dollars the readings are within a few points. Section~\ref{sec:ppi} refits the
calibrator to each reader and re-runs the estimates.

\paragraph{Where disagreement lives.} Table~\ref{tab:disagreement} fits two small logistic models on the text
features: whether the three LLMs disagree (over the population) and whether the human reference differs from the
LLM majority (over the sample), with the share of the disagreed dollars each cue carries. The reference and the
majority of the three language models give a different three-way status (research, not research, unclear) on
\csHumanVsMajTexts{} of design-weighted texts but only \csHumanVsMajDollars{} of dollars; this is the
three-way status, not the research-or-not answer of Table~\ref{tab:dependence}, and the disagreement is
concentrated in texts the models call ``unclear'' and the reference calls ``not research'': fellowships, chairs and
endowed positions, bare programme names, and general-support phrasing. The cues predict LLM disagreement only
weakly (cross-validated AUC \csDisagreeAUCLLM) and human-versus-majority disagreement more weakly still
(\csDisagreeAUCHuman): disagreement cannot be read off the surface form before reading. Texts that carry generic
phrasing (support, boilerplate, a centre or programme name, a fellowship or chair) and no purpose-naming word
(research, study, science, scholarship, building) hold \csGenericDisagreedDollars{} of the dollars the LLMs
disagree about and \csGenericHumanDisagreedDollars{} of the dollars on which the human and the majority disagree;
that is the share a controlled purpose vocabulary in the filing form could address, an estimate of its magnitude
and not a demonstration of its effect.

\begin{table}[tbp]\centering\small
\caption{Disagreement map: share of texts and of disagreed dollars carrying each cue, and standardised logistic
coefficients for LLM disagreement (population, $N=\csNTexts$) and for human-versus-majority disagreement
(sample).}\label{tab:disagreement}
\input{tables/t_disagreement}
\end{table}

Table~\ref{tab:stratum} (appendix) gives the calibrated probabilities' expected research and unclear dollars by
stratum. The population total of expected research dollars, $\sum_i d_i f_i = \$\csExpRTotalB$B, is a model-based
plug-in; the next section rectifies it on the human sample.

%% file: sections/ppi.tex
\section{Valid totals: design-weighted prediction-powered inference}\label{sec:ppi}

\subsection{Estimator}
Let $f_i \in [0,1]$ be a prediction for $\ind{y_i = R}$ available on every text, and let $D_s = \sum_{i \in s} d_i$
and $F_s = \sum_{i \in s} d_i f_i$ be the known frame dollars and predicted research dollars of stratum $s$. The
companion's estimator is the post-stratified ratio estimator
\begin{equation}
\hT_{\mathrm{cl}} = \sum_s D_s\, \frac{\sum_{i \in S_s} w_i d_i\, \ind{y_i = R}}{\sum_{i \in S_s} w_i d_i} .
\end{equation}
The design-weighted prediction-powered estimator rectifies the prediction within each stratum:
\begin{equation}
\hT_{\lambda} = \sum_s \Big[ \lambda F_s + D_s\, \frac{\sum_{i \in S_s} w_i d_i\, (\ind{y_i = R} - \lambda f_i)}{\sum_{i \in S_s} w_i d_i} \Big],
\qquad \lambda \in [0, \tfrac{3}{2}] ,
\label{eq:dwppi}
\end{equation}
so that $\lambda = 0$ recovers $\hT_{\mathrm{cl}}$, $\lambda = 1$ is the prediction-powered estimator of
\textcite{angelopoulos2023ppi} in Hájek form, and $\lambda^*$ chosen to minimise the bootstrap variance is the
PPI++ tuning of \textcite{angelopoulos2023ppipp}. In the survey vocabulary \eqref{eq:dwppi} is a post-stratified
difference estimator with $\lambda f_i$ as the auxiliary \parencite{sarndal1992,mozer2026difference}; the
stratification on the vote pattern is itself a generalised regression estimator with the pattern as auxiliary, which
is why a predictor can only reduce the within-stratum residual variance. The plug-in $\hT_{\mathrm{plug}} = \sum_i
d_i f_i$ is the predictor's own total with no residual correction on the sample (a supervised calibrator has
already used the human labels in its training; a raw machine reading or an unsupervised posterior has used
none) and equals $T$ only when the predictor's dollar-weighted errors cancel over the frame,
$\sum_i d_i (f_i - \ind{y_i = R}) = 0$, which no label checks. In the implementation every text of the frame is
assigned to one of ten folds by a seeded random permutation within each vote stratum, made without reference to
sample membership or labels and held fixed (the assignment was generated after the sample had been drawn, but
uses nothing about the draw, which is what the argument below needs); the model for fold $k$ is fitted on the
sampled texts outside fold $k$, with its ridge strength chosen by inner cross-validation on those texts and the
feature scaler fitted on the whole frame, and predicts every text in fold $k$, sampled or not. One vector $f$
enters the rectifier and $F_s$, and the prediction of a text does not depend on whether it was drawn. A
construction that gives a sampled text its out-of-fold model and an unsampled text the average of the fold
models makes the rule for $f_i$ depend on $I_i$, and Proposition~\ref{prop:ppi}(i) fails for it
(Appendix~\ref{app:proofs}).

\begin{proposition}[design properties]\label{prop:ppi}
Let $\hT^{\mathrm{HT}}_\lambda$ be \eqref{eq:dwppi} with $\sum_{i\in S_s} w_i d_i$ replaced by $D_s$, and let
Assumption~\ref{as:ign} hold.
\begin{enumerate}[label=(\roman*)]
\item If $f$ is fixed, or fitted on data independent of $S$, then $\hT^{\mathrm{HT}}_\lambda$ is exactly
  design-unbiased for $T$ for every fixed $\lambda$. If $f$ is cross-fitted on frame folds, so that $f_i$ is a
  function of $x$ and of $\{(I_j, y_j): j \notin \mathrm{fold}(i)\}$ with the folds fixed independently of sample
  membership and labels (before the draw, or afterwards by a rule that uses nothing about it), exact
  unbiasedness holds under Poisson sampling for every fixed $\lambda$; under a fixed-size design the bias is
  $-\lambda \sum_i d_i\, \E[(I_i/\pi_i - 1) f_i]$, which need not vanish, and is of the order of the dependence
  between $I_i$ and the inclusion of the other units. Leaving text $i$'s own label out is necessary but not
  sufficient: a construction whose rule for $f_i$ depends on $I_i$ need not be unbiased even under Poisson
  sampling.
\item The Hájek form \eqref{eq:dwppi} replaces $D_s$ by its estimate $\hat D_s = \sum_{i \in S_s} w_i d_i$ and is
  a ratio estimator. Its \emph{relative} bias in stratum $s$, the bias of the stratum's rectifier divided by $D_s$,
  is $O(n_s^{-1})$ under the usual conditions of the ratio expansion: a sequence of designs and populations in
  which every stratum's PPS count $n_s$ grows, $\hat D_s/D_s$ and the normalised residual total
  $\sum_{i \in S_s} w_i d_i e_i / D_s$ have variances of order $n_s^{-1}$ (bounded second moments of the weighted
  dollars, no weight that dominates its stratum), and $f$ and $\lambda$ are fixed. The absolute bias is that rate
  times $D_s$, so it scales with the stratum's dollars. These conditions are not verified for the committed design
  or for a predictor fitted on the sample, and the statement is not used for the reported intervals, which rest
  on the simulation below; a stratum whose realised denominator is zero contributes $\lambda F_s$ alone.
\item For fixed $f$ and fixed $\lambda$, $\Var(\hT^{\mathrm{HT}}_\lambda) = \Var_{\mathrm{HT}}\big(\sum_{i \in S} w_i d_i e_i\big)$
  with $e_i = \ind{y_i = R} - \lambda f_i$, which equals $\sum_i (d_i e_i)^2 (1-\pi_i)/\pi_i + \sum_{i \ne j} d_i d_j e_i e_j
  (\pi_{ij} - \pi_i \pi_j)/(\pi_i \pi_j)$; for a predictor fitted on the sample the term $\sum_i d_i f_i$ is random
  and its covariance with the rectifier enters, which the simulation below measures. For $\lambda = 1$, the conditional expectation $\E[e_i^2 \mid x_i]$ is
  minimised by $f_i = P(y_i = R \mid x_i)$; this minimises the diagonal term for every design and the whole
  variance under Poisson sampling.
\item Over $\lambda$ the variance of a fixed-$f$ estimator is a quadratic whose minimum is at most its value at
  $\lambda = 0$; an estimated $\lambda^*$ need not attain this in finite samples.
\end{enumerate}
\end{proposition}
\noindent The proof is in Appendix~\ref{app:proofs}, together with a two-unit example enumerated exactly
(Table~\ref{tab:two_unit}): under one-of-two sampling the frame-fold predictor gives a bias of one half of the
total and none under Poisson sampling, while the sampled/unsampled switch has expectation
\csTwoUnitSwitchPoissonExp{} against a total of $1$ under Poisson sampling, so the exactness in (i) depends on
the construction and not only on leaving the text's own label out. The by-year totals, the growth ratio and the
Hájek form are ratios and inherit (i) only approximately; a tuned $\lambda^*$ is not covered by (i) and (iii) and
is checked by simulation. On the committed design, which is fixed-size systematic PPS within strata, the operative
check of (i)--(iv) together is the simulation below, which refits the cross-fitted calibrator inside every re-draw.
All intervals are 95\% percentile intervals from 2,000 bootstrap replicates of the with-replacement type of
\textcite{raowu1988}: within each stratum the $n_s$ PPS draws are resampled with replacement $n_s - 1$ times and
each draw's design weight is multiplied by $n_s/(n_s - 1)$ times its resampling count; the certainty and take-all
texts keep weight one in every replicate and enter the stratum sums with their own dollars. Those texts sit
inside the stratum ratio of \eqref{eq:dwppi}, so their contribution is scaled by the same random normalisation
$D_s / \sum_{i \in S_s} w_i d_i$ as the draws'; this is a choice (the pooled ratio), not the estimator that adds
the read texts' dollars exactly and estimates only the rest of the stratum, which we did not compute. Further, the same replicate
weights are used for every tax year, so the growth ratio below carries the joint covariance of the yearly totals.
This is the construction the companion used, checked there by simulation against the systematic draw; the
systematic-PPS bootstrap of \textcite{nces1999bootstrap} is the closest documented alternative and was not used.

\subsection{Results}
\begin{figure}[tbp]\centering
\includegraphics[width=0.9\textwidth]{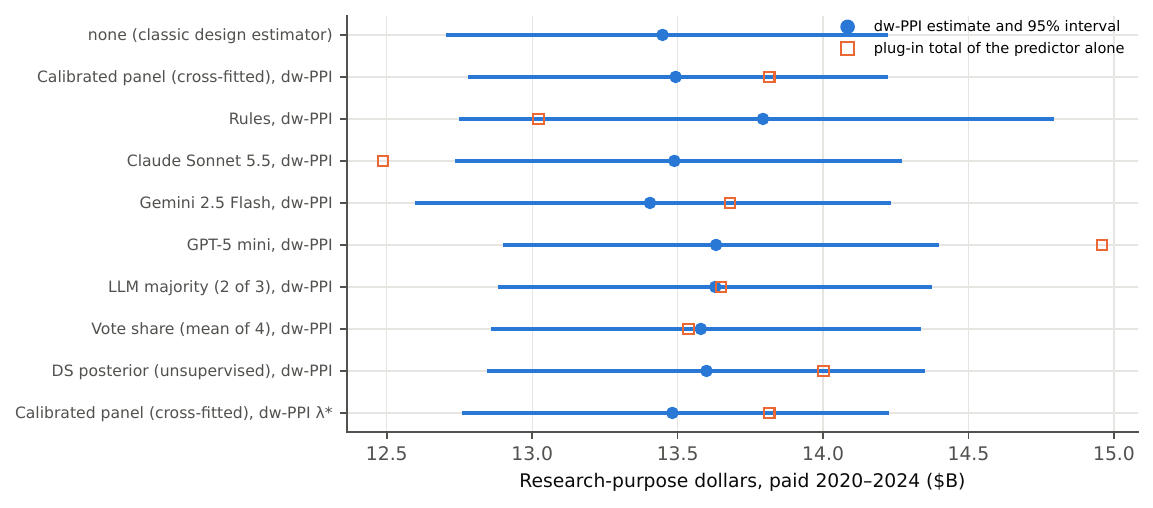}
\caption{Each predictor's plug-in total (open squares) and its dw-PPI estimate with 95\% interval (filled). The
readers disagree by \$\csReaderTotalMinB{}--\csReaderTotalMaxB{}B; rectified on the same human sample they agree.}
\label{fig:ppitot}
\end{figure}

Table~\ref{tab:ppi} gives the total of research-purpose dollars on the headline domain for every predictor. The
classic estimate is \$\csClassicB{}B [\csClassicLoB, \csClassicHiB]. With the calibrated panel, dw-PPI gives
\$\csPpiB{}B [\csPpiLoB, \csPpiHiB] at $\lambda = 1$ and \$\csPpiStarB{}B [\csPpiStarLoB, \csPpiStarHiB] at
$\lambda^* = \csLambdaStar$, an interval \csWidthRatioCal{} times the classic width (an effective-sample-size gain of
\csEssGainCal). The result is not the number, which the human sample alone already gave, but the agreement: the
\csNPredictors{} predictors' plug-in totals span \$\csPluginAllMinB{}--\csPluginAllMaxB{}B (the four readers alone
\$\csReaderTotalMinB{}--\csReaderTotalMaxB{}B), and their dw-PPI estimates span
\$\csPpiAllMinB{}--\csPpiAllMaxB{}B, every one inside the classic interval (Figure~\ref{fig:ppitot}). The rectifier
removes each reading's bias, and a biased reading costs width rather than validity: the rules classifier, whose
plug-in total is \$\csReaderTotalRulesB{}B, gives an interval \csWidthRatioRules{} times wider at $\lambda = 1$ and
is tuned back to $\lambda^* = \csLambdaStarRules$. The unclear total is \$\csUnclearPpiB{}B [\csUnclearPpiLoB,
\csUnclearPpiHiB] with the calibrated $P(U)$; the unclear status of the reference was adjudicated only on the
\csNSplit{} split texts and is the readers' majority status on the \csNStatusOnlySplit{} texts where they differ
between not research and unclear (Section~\ref{sec:testbed}).

\begin{table}[tbp]\centering\small
\caption{Research-purpose dollars, paid 2020--2024 (\$B), by predictor. ``Plug-in'' is the predictor's own total
with no residual correction on the sample (the calibrated predictors were trained on the human labels); the
dw-PPI estimate rectifies it on the human sample; $\lambda = 0$ reproduces the classic
estimator for every predictor and is shown once. ESS gain is the squared ratio of the classic interval width to
the estimator's.}\label{tab:ppi}
\adjustbox{max width=\textwidth}{\input{tables/t_ppi}}
\end{table}

\paragraph{Why the gain is small: a regime analysis.} The committed design draws by dollars within strata that are
the readers' own vote patterns, so most of what a predictor knows is already in the post-stratification. To
measure this, Table~\ref{tab:regime} re-runs three designs on the real frame, a simple random sample of texts
(no certainty units), PPS by dollars without strata (texts whose size would give $\pi_i \ge 1$ taken with
certainty) and the committed design (the per-stratum PPS counts scaled by $n/651$, at least five per stratum,
the take-all texts kept), at $n \in \{150, 300, 651, 1300\}$, with three fixed predictors (rules, GPT-5 mini,
the calibrated panel) and two synthetic truths (A: labels drawn once from the calibrated probabilities, a
model-aligned scenario favourable to the panel; B: one machine reader's labels, Claude Sonnet~5.5's, taken as the
truth, so the panel is biased), \csRegDraws{}
draws per cell, no bootstrap. The quantity reported is the \emph{variance ratio}: the Monte Carlo variance of the
classic estimator over that of the Hájek dw-PPI estimator at $\lambda = 1$ on the same draws, with a Monte Carlo
standard error from a bootstrap over the draws (between \csRegSeMin{} and \csRegSeMax{} across cells; with
concentrated masses, \csRegDraws{} draws leave the largest ratios imprecise). Under a simple random sample of
texts the calibrated panel's ratio is \csRegASrsSixFiftyOne{} at $n = 651$ (truth A; \csRegBSrsSixFiftyOne{}
under truth B), under unstratified PPS \csRegAPpsSixFiftyOne, and under the committed design
\csRegAStratSixFiftyOne{} (\csRegBStratSixFiftyOne{} under truth B), with a single language model giving
\csRegAStratMiniSixFiftyOne{} there. Under the committed design the ratio at $\lambda = 1$ falls below one for
the rules classifier and, under truth~B, for every predictor: the post-stratification already absorbs what the
predictor knows, and at $\lambda = 1$ the rectifier adds the predictor's noise. Tuning can reduce the loss from an
unhelpful predictor (the tuned estimator returns to the classic one when the predictor adds nothing; under
truth~B the coverage simulation below tunes the fixed predictor to $\lambda^* \approx \csLambdaBFix$), but the
gain of the estimated tuning rule is a separate question, assessed in that simulation, and is not bounded below
by the untuned ratio, because an estimated $\lambda^*$ can do worse than $\lambda = 0$ in finite samples
(Proposition~\ref{prop:ppi}(iv)). These are results for one frame and two truths, not a general policy. What
they support is: an office that can stratify its validation sample on the machines' votes gains little from a
calibrated predictor in the estimate of the total; an office that draws its sample uniformly gains a lot; and an
office that has already drawn should calibrate anyway, because the same predictor is what the accept rules and
the review policies of the next two sections run on.

\begin{table}[tbp]\centering\small
\caption{Regime analysis on the real frame: the variance ratio of the classic estimator to dw-PPI ($\lambda = 1$)
under the same design, from \csRegDraws{} Monte Carlo draws per cell (fixed predictors, no bootstrap), with the
Monte Carlo standard error of each ratio from a bootstrap over the draws. Truth A is a model-aligned scenario in
which the calibrated panel is exactly calibrated; it is favourable to the panel, not a bound.}\label{tab:regime}
\adjustbox{max width=\textwidth}{\input{tables/t_regime}}
\end{table}

\begin{figure}[tbp]\centering
\includegraphics[width=0.8\textwidth]{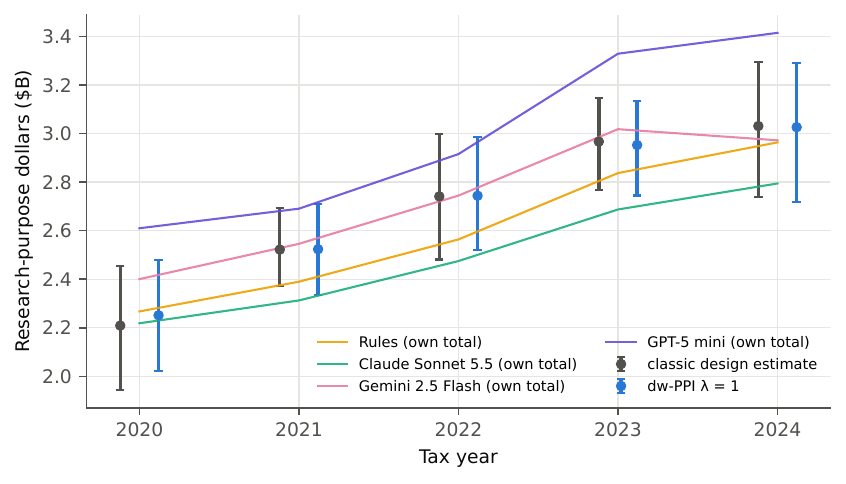}
\caption{Totals by tax year: the readers' own totals (lines) against the classic and dw-PPI estimates with 95\%
intervals.}\label{fig:byyear}
\end{figure}

\paragraph{By year and growth.} The same estimator applied with the year's dollars $d_{i,y}$ gives
Table~\ref{tab:byyear} and Figure~\ref{fig:byyear}. Each year's rectifier has its own Hájek denominator and its own
$\lambda$, so the yearly estimates need not add to the pooled one (classic: \$\csSumYearsClassicB{}B summed against
\$\csClassicB{}B pooled). The per-year $\lambda^*$ ranges from \csLambdaYearMin{} to \csLambdaYearMax{} with
\csNYearMin{}--\csNYearMax{} sampled texts carrying dollars in a year, which is the small-sample noise of a bootstrap
minimiser rather than evidence about the predictor; we therefore report growth at $\lambda = 1$, at the per-year
$\lambda^*$ clipped to $[0, 1]$ as PPI++ recommends in small samples, at the pooled $\lambda^* = \csLambdaStar$, and
at the unclipped per-year value. Growth from 2020 to 2024 is \csGrowthClassic{} [\csGrowthClassicLo,
\csGrowthClassicHi] under the classic estimator and \csGrowthPpiOne{} [\csGrowthPpiOneLo, \csGrowthPpiOneHi] under
dw-PPI at $\lambda = 1$ (\csGrowthPpiClip{} clipped, \csGrowthPpiPooled{} pooled; Table~\ref{tab:growth}); the
readers' own totals grow by \csGrowthReadersMin{}--\csGrowthReadersMax{} and the whole frame by \csGrowthFrame.

The benchmark is the growth of federally financed R\&D expenditure at US universities from FY2020 to FY2024 in
current dollars, \csFedGrowth{} \parencite{nsf26305}. We test $H_0$: growth $\ge$ \csFedGrowth{} with a one-sided
design-based bootstrap test (a bootstrap-$t$ p-value from the replicate standard error, and a percentile-shift
p-value; both in Table~\ref{tab:growth}). The p-value is \csPFedPpiOne{} (bootstrap-$t$; \csPFedShiftPpiOne{} shift)
under dw-PPI at $\lambda = 1$ and \csPFedClassic{} under the classic estimator; the p-value for $H_0$: no growth is
\csPZeroPpiOne. The estimated increase in research-labelled reported amounts is therefore \csGrowthPpiOne, with an
interval from \csGrowthPpiOneLo{} to \csGrowthPpiOneHi{} under the stated sampling and reference assumptions. This
interval excludes no growth and does not resolve parity with the \csFedGrowth{} expenditure-series benchmark;
differences in accounting and coverage between reported grant payments and R\&D expenditure further limit the
comparison, and under reader~1's labels the interval reaches down to zero (Table~\ref{tab:reference_rerun}). What the rectifier changes is the point estimate and the test: the classic estimator's \csGrowthClassic{}
and the readers' \csGrowthReadersMin{}--\csGrowthReadersMax{} become \csGrowthPpiOne{} once each year's labelling
error is estimated on that year's dollars; Section~\ref{sec:invariance} shows the size of that error by year.

\begin{table}[tbp]\centering\small
\caption{Research-purpose dollars by tax year (\$B): sampled texts with dollars in the year, frame, the four
readers' own totals, the classic estimate, and dw-PPI at $\lambda = 1$, at the per-year $\lambda^*$ clipped to
$[0,1]$ (value in parentheses) and at the pooled $\lambda^*$.}\label{tab:byyear}
\adjustbox{max width=\textwidth}{\input{tables/t_by_year}}
\end{table}

\begin{table}[tbp]\centering\small
\caption{Growth of research-purpose dollars 2020$\to$2024 with one-sided design-based bootstrap tests against the
federal benchmark (\csFedGrowth) and against no growth. ``$t$'' is the bootstrap-$t$ p-value, ``shift'' the
percentile-shift p-value.}\label{tab:growth}
\adjustbox{max width=\textwidth}{\input{tables/t_growth}}
\end{table}

\paragraph{Alternative references and a noisy reference.} Table~\ref{tab:altref} re-runs the classic and
dw-PPI estimators under every labelling of the full sample, with the calibrator fitted to the adjudicated reading
held fixed. Reader~1's labels give \$\csReaderOneAltPpiB{}B [\csReaderOneAltPpiLoB, \csReaderOneAltPpiHiB]
(\$\csRecheckAltPpiB{}B [\csRecheckAltPpiLoB, \csRecheckAltPpiHiB] with reader~1's blind re-read substituted
where it exists), reader~2's \$\csReaderTwoPpiB{}B [\csReaderTwoPpiLoB, \csReaderTwoPpiHiB], reader~3's
\$\csReaderThreePpiB{}B [\csReaderThreePpiLoB, \csReaderThreePpiHiB] and the majority of the three
\$\csMajRPpiB{}B [\csMajRPpiLoB, \csMajRPpiHiB]; counting a text as research only when all three readers say so
gives \$\csAllRPpiB{}B and when any reader does \$\csAnyRPpiB{}B. The three readers' totals span
\$\csAltReadersSpreadPpiB{}B against a \$\csPpiWidthB{}B sampling interval under the reference: these are
different operational measurements of the same codebook, and their spread is the reference uncertainty that the
sampling intervals do not contain. The adjudicated total lies between the majority's and reader~3's. Because the
readings label the same sampled texts, the change of reference is a paired contrast, and
Table~\ref{tab:reference_contrast} computes it on the same bootstrap replicates: the adjudicated total is
\$\csConOneB{}B [\csConOneLoB, \csConOneHiB] above the total under reader~1, the reference of the first version,
\$\csConTwoB{}B [\csConTwoLoB, \csConTwoHiB] above reader~2's and \$\csConMajB{}B [\csConMajLoB, \csConMajHiB]
above the majority's, while its difference from reader~3's, \csConThreeSign\$\csConThreeAbsB{}B [\csConThreeLoB, \csConThreeHiB],
has an interval that contains zero. With the predictor and $\lambda$ held fixed the prediction terms cancel, so
the contrast is the same for the classic estimator and for dw-PPI; it isolates the labels. The
\$\csAdjMinusOnePpiB{}B quoted in the abstract and in Section~\ref{sec:discussion} is a different comparison,
against the pipeline refitted to reader~1 (Table~\ref{tab:reference_rerun}), where the calibrator changes too. The
table also shows both directions of the change, which the totals hide: relative to reader~1 the adjudicated
reading counts \csConOneGain{} more sampled texts as research (\csConOneGainU{} that reader~1 called unclear and
\csConOneGainN{} not research, worth \$\csConOneGainB{}B in the estimate) and \csConOneLoss{} fewer
(\$\csConOneLossB{}B); relative to reader~3 the two directions nearly cancel (\csConThreeGain{} against
\csConThreeLoss{} texts). A second view treats the reference's research-or-not label as a screening test with
sensitivity $Se$ and specificity $Sp$ relative to the codebook's intended label and corrects the dollar-weighted
prevalence as $(p - (1 - Sp))/(Se - (1 - Sp))$ \parencite{rogangladen1978}; Appendix Table~\ref{tab:rogan} gives
the corrected total over a grid, and with reader~1 as the comparator ($Se = \csRGOneSe$, $Sp = \csRGOneSp$,
dollar-weighted on all \csNSample{} texts) the total becomes \$\csRGOneTotalB{}B, with reader~2
($Se = \csRGSe$, $Sp = \csRGSp$) \$\csRGTotalB{}B and with reader~3 ($Se = \csRGThreeSe$,
$Sp = \csRGThreeSp$) \$\csRGThreeTotalB{}B. These sensitivities and specificities are rates of agreement between
the reference and a comparator reading, both human, not error rates validated against the intended label, which
no one observed. The inputs are normalised differently: $p$ is the post-stratified classic share of the domain
dollars (\csRGPostShare), while $Se$ and $Sp$ come from the pooled design-weighted two-by-two table of the
sample, in which the adjudicated share is \csRGPooledShare. A correction whose three inputs come from one
two-by-two table returns the comparator's own prevalence exactly, which we checked on the pooled table; with the
post-stratified $p$ it returns a total near the comparator's and not equal to it (\$\csRGOneTotalB{}B against
reader~1's classic \$\csReaderOneAltClassicB{}B). The corrected totals are therefore the change of reference in
another form, up to that normalisation, and are not a second validation of the estimate. The efficient treatment of a noisy reference with a gold subset is
that of \textcite{chen2026judge}; we have no gold subset, so this remains a sensitivity analysis.

\begin{table}[tbp]\centering\small
\caption{Totals (\$B) under alternative labellings of the full \csNSample-text sample, calibrator held fixed:
each reader, reader~1 with the blind re-read substituted, the majority of the three, the strict and loose
consensus, and the adjudicated reading; the research share of texts and of design-weighted dollars under each
labelling.}\label{tab:altref}
\adjustbox{max width=\textwidth}{\input{tables/t_altref}}
\end{table}

\begin{table}[!tb]\centering\small
\caption{From each reading to the adjudicated reading on the same \csNSample{} sampled texts. Panel~A: the
contrast is the adjudicated total minus the total under the other reading, with a 95\% percentile interval from
the same \csPpiBoot{} replicate weights (predictor and $\lambda$ fixed, so the classic and dw-PPI contrasts
coincide). Panel~B: the texts that one reading counts as research and the other does not, with their status (U
unclear, N not research) under the reading that does not, and the dollars each group carries in the estimate, both
printed as positive amounts; the contrast of panel~A is the first dollar column minus the second. Every entry is
computed from unrounded values, so a difference of displayed cells can miss the displayed contrast by one in the
last digit (reader~3: \csConRoundingNote{} against \csConThreeSign\csConThreeAbsB).}\label{tab:reference_contrast}
\input{tables/t_reference_contrast}
\end{table}

\paragraph{The pipeline re-run under each reader.} Holding the calibrator fixed answers only part of the
question, because a reader who draws the research line elsewhere would also have produced a different
calibrator. Table~\ref{tab:reference_rerun} re-runs the pipeline from the frame onward with each reader's labels,
and their majority, as the reference, beside the main run under the adjudicated reading: the calibrator is refitted
(cross-fitted, ridge strength chosen again), and the totals, the growth test, the Learn-then-Test rule of
Section~\ref{sec:selective} and the protected replay of the review policies of Section~\ref{sec:acquisition} are
recomputed; Appendix Table~\ref{tab:reference_rerun_sections} does the same for the sensitivity study
(Section~\ref{sec:invariance}), the institution totals (Section~\ref{sec:downstream}) and the designs with every
label charged (Section~\ref{sec:adaptive}). The ranges quoted below are over the three readers and their
majority; the adjudicated reading is built from their labels and is quoted beside them. The panel and the text
cues reproduce the adjudicated reading best on the design-weighted text metrics (held-out log-loss
\csRerunAdjLL, AUC \csRerunAdjAUC; by dollars the majority's log-loss, \csRerunMajDLL, is a little below the
adjudicated reading's \csRerunAdjDLL), then
reader~2, reader~3 and the majority (\csRerunTwoLL, \csRerunThreeLL{} and \csRerunMajLL; AUC \csRerunTwoAUC,
\csRerunThreeAUC{} and \csRerunMajAUC), and reader~1 worst (\csRerunOneLL; AUC \csRerunOneAUC): reader~1's
research line is the hardest of the three for the panel and the text cues to reproduce. The totals move with the
reading, not with the calibrator: the dw-PPI total is \$\csRerunOnePpiB{}B [\csRerunOnePpiLoB,
\csRerunOnePpiHiB] under reader~1, \$\csRerunTwoPpiB{}B [\csRerunTwoPpiLoB, \csRerunTwoPpiHiB] under
reader~2, \$\csRerunThreePpiB{}B [\csRerunThreePpiLoB, \csRerunThreePpiHiB] under reader~3 and
\$\csRerunMajPpiB{}B [\csRerunMajPpiLoB, \csRerunMajPpiHiB] under the majority, each within
\$\csRerunVsFixedMaxB{}B of the fixed-calibrator total of Table~\ref{tab:altref}, and the intervals under
readers~2 and~3 do not overlap. The regime does not change: $\lambda^*$ is \csRerunLamMin{}--\csRerunLamMax{}
and the effective-sample-size gain of the rectified estimator over the classic one
\csRerunEssMin{}--\csRerunEssMax{} under every reader, against \csRerunAdjLam{} and \csRerunAdjEss{} under the
adjudicated reading. The growth test does change: the dw-PPI growth estimate at the per-year $\lambda^*$ runs from
\csRerunGrowthMin{} to \csRerunGrowthMax{} across the readers and the one-sided $p$-value against the federal
benchmark from \csRerunGrowthPMin{} to \csRerunGrowthPMax{} (\csRerunAdjGrowth{} and \csRerunAdjGrowthP{}
under the adjudicated reading), so what the growth interval admits is a statement about the reading as much as
about the sample. The acceptance rule of Section~\ref{sec:selective} keeps its behaviour and its yield follows
the reading: the Learn-then-Test rule at $\alpha = 0.05$ accepts \csRerunAdjLtt{} of the domain's dollars under
the adjudicated reading, \csRerunOneLtt{} under reader~1 and \csRerunMajLtt{} under the majority but nothing
under \csRerunLttEmptyRefs, whose labels put the estimated dollar error of the candidate sets above the bound,
and at $\alpha = 0.10$ it accepts \csRerunLttTenMin{}--\csRerunLttTenMax{} under every reader, with a held-out
dollar error of at most \csRerunLttTenErrMax, and \csRerunAdjLttTen{} under the adjudicated reading
(\csRerunAdjLttTenErr). The protected replay of the review policies gives the same ranking under every reading,
with a margin that depends on it: the mixed active design beats dollar-PPS at 100 texts under each reader, by
\csRerunAcqRatioMin{}--\csRerunAcqRatioMax$\times$ with coverage \csRerunAcqCovMin{}--\csRerunAcqCovMax, and
under the adjudicated reading by \csRerunAdjAcqRatio$\times$ with coverage \csRerunAdjAcqCov.

\begin{table}[tbp]\centering\small
\caption{Headline results under the adjudicated reading (the reference of this paper, first column) and with the
pipeline re-run under each reader's labels and under their majority (calibrator refitted to that labelling): the
calibrator's held-out quality, the totals and intervals, the regime, the growth test, the
Learn-then-Test rule at $\alpha = 0.05$ and $0.10$, and the protected replay of the active review design at 100
texts (the mixed design's RMSE ratio against dollar-PPS, and its coverage).}
\label{tab:reference_rerun}
\input{tables/t_reference_rerun}
\end{table}

\paragraph{Coverage.} Because the design is systematic PPS with take-all units, no textbook variance applies
exactly; the companion checked its estimator by simulation and we extend the check to the whole fitted procedure.
For three synthetic truths on the whole frame, (A) labels drawn from the calibrated probabilities, (B) one machine
reader's labels (Claude Sonnet~5.5's, not a human reader's) taken as the truth so that the predictor is biased, and (C) truth~A with 10\% of texts flipped uniformly at
random, we re-ran the committed draw procedure \csSimDraws{} times. In each re-draw the calibrator is refitted on
the drawn sample's synthetic labels (the frame-fold cross-fit of Section~\ref{sec:ppi}, ridge strength fixed at
the value chosen on the real sample, so this check is conditional on that choice), $\lambda^*$ is re-tuned, and
every estimator is computed with 300 bootstrap replicates (``refit''); the real-sample calibrator is also used as a
fixed predictor (``fixed'') so that the two sources of error can be told apart. A further \csRetuneDraws{} draws
under truth~A re-choose the ridge strength inside every draw by the nested inner cross-validation of
Section~\ref{sec:aggregation}, the full model-selection pipeline (Appendix Table~\ref{tab:coverage_retune}):
coverage \csRetuneCov{} (the classic estimator's on the same draws is \csRetuneClassicCov) and bias
\csRetuneBias{} (Monte Carlo s.e.\ \csRetuneBiasSE), with the real-sample strength chosen in
\csRetuneShareSameC{} of the draws. The classic estimator, which uses no predictor, shifts by
\csRetuneClassicBias{} on the same \csRetuneDraws{} draws, so the shift is the draws' common sampling fluctuation
and not an effect of re-tuning (paired difference \csRetuneBiasMinusClassic); with \csRetuneDraws{} draws the
check is imprecise, and it says that re-tuning did not visibly change the picture, not that tuning is immaterial.
Within a draw, $f$ and $\lambda^*$ are held fixed across the bootstrap replicates; refitting happens across
re-draws, not inside the resampling. Truth~C's total of \$\csTruthCB{}B exceeds its expectation of
\$\csTruthCExpectedB{}B because the flips happened to include \csTruthCFlippedTop{} of the twenty largest texts
(\$\csTruthCFlippedTopB{}B); it is deliberately far from the predictor. Table~\ref{tab:coverage} gives the results
with Monte Carlo standard errors (about \csCovMCSE{} for a coverage near 0.95). With the refitted calibrator the
dw-PPI intervals cover at \csCovARefit{} (A), \csCovBRefit{} (B) and \csCovCRefit{} (C), the classic estimator at
\csCovAClassic{} under~A; the bias of the refitted, cross-fitted estimator is \csBiasARefit{} (A), \csBiasBRefit{}
(B) and \csBiasCRefit{} (C), the largest of them \csBiasRefitMaxRatio{} Monte Carlo standard errors from zero,
so the fixed-size cross-fitting term of Proposition~\ref{prop:ppi}(i) is at or below the resolution of
\csSimDraws{} draws on this design. Under truth~B the refitted estimator covers at \csCovBRefit, within one Monte Carlo standard error of nominal,
and its bias is negligible. That scenario is the one to watch: the refitted calibrator reproduces the truth almost
exactly, the residual variance is small (Table~\ref{tab:coverage}), and the bootstrap replicates can understate
it. With the ridge setting chosen when reader~1 was the reference ($C = \csSimOneCfinal$ against
$C = \csCfinal$ here) the refitted calibrator came closer still, with a mean interval width of
\$\csSimOneWidthBRefit{}B against \$\csWidthBRefit{}B here, and the same scenario covered at
\csSimOneCovBRefit, \csSimOneCovBShortSE{} Monte Carlo standard errors below nominal (the replication unit is
the re-draw of the sample). That is modest evidence from one configuration, and we have not separated the
with-replacement approximation from the tuned $\lambda^*$ as its cause, so the coverage
we advertise is for predictors that leave a residual of the size seen on the real sample, not for a predictor
that nearly equals the truth. Under truth~B the fixed predictor's plug-in misses by \csPlugBFix{} while dw-PPI with the fixed
predictor is tuned to $\lambda^* \approx \csLambdaBFix$ and stays unbiased; the refitted calibrator learns the new
truth and keeps $\lambda^* \approx \csLambdaBRefit$.

\begin{table}[tbp]\centering\small
\caption{Coverage simulation: \csSimDraws{} re-draws of the committed design on three synthetic truths, with the
calibrator held fixed or refitted (cross-fitted) in every draw. Monte Carlo standard errors in parentheses.}\label{tab:coverage}
\adjustbox{max width=\textwidth}{\input{tables/t_coverage}}
\end{table}

%% file: sections/selective.tex
\section{Selective machine labelling at a stated dollar-weighted error}\label{sec:selective}

\subsection{The decision and the two measures}
For an unlabelled text $i$ the machine decision is $\hat y_i$ (research if $\hat p_i(R) \ge \tfrac12$ in the
binary task; the arg-max class in the three-class task) with confidence $c_i$ and score $s_i = 1 - c_i$. An
office wants to accept the machine label on as many texts as possible while the \emph{dollars} that carry a wrong
accepted label stay below a share $\alpha$ of the accepted dollars. Following \textcite{jincandes2023} and
\textcite{huang2025selective}, the null hypothesis for each text is $H_i$: ``the machine label is wrong''; a
conformal p-value compares $s_i$ with the scores of the calibration texts whose labels were wrong (the nulls), and a
Benjamini--Hochberg step selects a set with false-discovery control. The calibration texts are the human sample with
out-of-fold scores; \csNNullBinary{} of its PPS draws are nulls in the binary task.

The question is what ``exchangeable'' means when the calibration set is a stratified PPS sample and the guarantee
is wanted in dollars. Weighted conformal inference \parencite{tibshirani2019} gives validity under covariate shift
when every unit, calibration or test, enters with the likelihood ratio $w(x) = \mathrm{d}P_{\mathrm{test}}/\mathrm{d}P_{\mathrm{cal}}$
of the two laws at its covariate, and the design determines that ratio. Model the $n_s$ PPS draws of stratum $s$
as independent draws from the size measure, $P_{\mathrm{cal}}(i) \propto (n_s/n)\, a_i/A^{\circ}_s$ (the with-replacement
approximation to systematic sampling); certainty and take-all texts are in the sample with probability one and are
not draws from this law.
\begin{description}[leftmargin=1.2em]
\item[Dollar measure.] A test unit is a random dollar of the unsampled population: $P_{\mathrm{test}}(i) \propto d_i$.
  The likelihood ratio at any text in the PPS-eligible part of stratum $s$ is $\propto d_i A^{\circ}_s/(n_s a_i) =
  d_i/\pi_i = m_i$, the dollars the text stands for, and it is zero at a certainty or take-all text. So
  \begin{equation}
  p_i = \frac{\sum_{j \in \mathcal{N}} m_j\, \ind{s_j \le s_i} + m_i}{\sum_{j \in \mathcal{N}} m_j + m_i},
  \qquad m_i = d_i/\pi_i ,
  \label{eq:pdollar}
  \end{equation}
  with $\mathcal{N}$ the mislabelled PPS draws. The test text must enter with $d_i/\pi_i$ and not with $d_i$: in
  an equal-dollar example with 100 texts, 10 drawn by simple random sampling and every label wrong, the formula
  with $d_i$ gives $P(p \le 0.01) = \csExUncorrPle$ and a false-discovery rate of \csExUncorrFDR{} for BH at
  $\alpha = 0.05$ over \csExReps{} repetitions, while \eqref{eq:pdollar} gives \csExCorrPle{} and \csExCorrFDR{}
  (Appendix~\ref{app:stress}).
\item[Text measure.] A test unit is a random unsampled text: $P_{\mathrm{test}}(i) \propto 1$, the likelihood ratio
  is $\propto 1/\pi_i$, and every unit enters with $1/\pi_i$. This is the count guarantee one would usually want.
\end{description}

\begin{proposition}[validity of the p-values in an idealised experiment; power under the text measure]\label{prop:conformal}
Consider the following experiment. A score function $s(\cdot)$ and a machine decision rule are fixed using
training information separate from the units below; the null calibration units are independent draws from
$P_{\mathrm{cal}}$ restricted to the event ``the machine label is wrong''; the test unit is an independent draw
from $P_{\mathrm{test}}$ restricted to the same event; $P_{\mathrm{test}}$ is absolutely continuous with respect to
$P_{\mathrm{cal}}$ with likelihood ratio $w(x) \propto \mathrm{d}P_{\mathrm{test}}/\mathrm{d}P_{\mathrm{cal}}$, finite on
the support. Then the weighted conformal p-value with weights $w$ on the calibration units and on the test unit is
super-uniform, $P(p_i \le t) \le t$ for all $t$: under the dollar measure, with $w = d/\pi$ on the PPS-eligible
texts and $0$ on certainty texts, this is \eqref{eq:pdollar}; under the text measure it is the $1/\pi$
analogue. Under the text measure every p-value satisfies $p_i \ge (1/\pi_i) / (\sum_{j \in \mathcal{N}} 1/\pi_j +
1/\pi_i)$, so a text whose design weight exceeds $\alpha/(1-\alpha)$ times the summed weight of the nulls can
never be accepted at level $\alpha$ by any BH-type step.
\end{proposition}
\noindent The first statement is Theorem~2 of \textcite{tibshirani2019} applied to the null calibration draws and
the test unit with a common, separately fixed score; the second is read off the formula
(Appendix~\ref{app:proofs}). The deployed procedure departs from this experiment in four ways, each stated here
so that the proposition is not read as a theorem for the realised unsampled population: the calibration units are
a fixed number of systematic PPS draws per stratum, not independent draws from a mixture; the target population
is the realised set of unsampled texts, which depends on the calibration draw; the sampled texts are excluded
from the target; and the calibration scores are cross-fitted rather than produced by one separately trained
score. The checks below measure what these departures cost. Under the dollar measure the dollar weights $d_i$
then make the weighted BH step of \textcite{bh1997weighted} the natural selection rule (reject the $k$ smallest
p-values where $k$ is the largest index with $p_{(k)} \le \alpha \sum_{j \le k} d_{(j)}/\sum_j d_j$), whose target is
the \emph{dollar-weighted} FDR, the expected share of accepted dollars carrying a wrong label. BH needs positive
regression dependence, which conformal p-values with a shared calibration set have in the unweighted case
\parencite{bates2023outliers} but need not have under covariate shift \parencite{jincandes2023weighted}. We do not
have a proof for the weighted step with dollar weights, so its error control is assessed empirically in the
scenarios below, not proved. Zero-dollar texts and texts with $\pi_i = 0$ have no mass under the dollar measure
and are left to a person.

\paragraph{A rule with no dependence assumption, evaluated rather than certified.} Learn-then-Test
\parencite{angelopoulos2021ltt} controls a risk over a family of thresholds by testing each threshold and
correcting for multiplicity. For a confidence threshold $t$ let $R(t)$ be the share of dollars with a wrong machine
label among the unsampled texts with $c_i \ge t$. The design (Hájek) estimate from the PPS draws is
$\hat R(t) = \sum_{j \in S^{\circ}: c_j \ge t} m_j e_j / \sum_{j \in S^{\circ}: c_j \ge t} m_j$ with $e_j$ the error
indicator and $S^{\circ}$ the PPS draws; its Kish effective size is $n_{\mathrm{eff}}(t) = (\sum m_j)^2/\sum m_j^2$
\parencite{kish1965}. For each $t$ on a grid of \csNGrid{} quantiles of the calibration texts' confidence (a grid
chosen from the data, not in advance) we test $H_t$: $R(t) > \alpha$ with the Hoeffding--Bentkus p-value of
\textcite{angelopoulos2021ltt} at $n = \lfloor n_{\mathrm{eff}}(t) \rfloor$, and with a Rao--Wu bootstrap p-value as
a check; Bonferroni over the grid at level $\delta = \csDelta$ (which does not need $R$ to be monotone in $t$, and
it need not be) would give $P(R(t^*) \le \alpha) \ge 1 - \delta$ for the largest accepted set $t^*$ \emph{if the
p-values were valid}. They are not in general. The Hoeffding--Bentkus bound is exact for independent equal-weight
draws, and substituting the Kish size for unequal weights has no finite-sample justification: a weighted error
share can be zero while its target exceeds $\alpha$. With 1,000 independent units, one of them carrying a tenth
of the weight and wrong with probability 0.85 and the rest always right, the weighted risk is 0.085 and the
effective size 92.5; with probability 0.15 the observed error is zero, the substituted p-value is 0.0004, below
the Bonferroni cutoff $\delta/\csNGrid$, and a rule whose true error exceeds $\alpha = 0.08$ is accepted in 15\%
of samples, above $\delta$. The committed design adds sampling
without replacement, fixed stratum sizes, inclusion probabilities near one and the estimation of a risk among the
unsampled texts, none of which the bound addresses. We therefore report Learn-then-Test as an empirically
evaluated acceptance rule, not a certificate: the quantity its statement is about, the probability that the
realised error of the accepted set exceeds $\alpha$, is measured directly on re-draws of the design in
Appendix~\ref{app:stress}, including a truth whose expected error sits just above $\alpha$ where a false
acceptance is hardest to detect. A valid version needs an audit sample with a finite-sample bound derived for its design
(Section~\ref{sec:discussion}). The sample's effective size under the dollar measure, about \csNEffAll{} at the
loosest threshold, is what limits the rule's yield.

\subsection{Evaluation and results}
\begin{figure}[tbp]\centering
\includegraphics[width=0.98\textwidth]{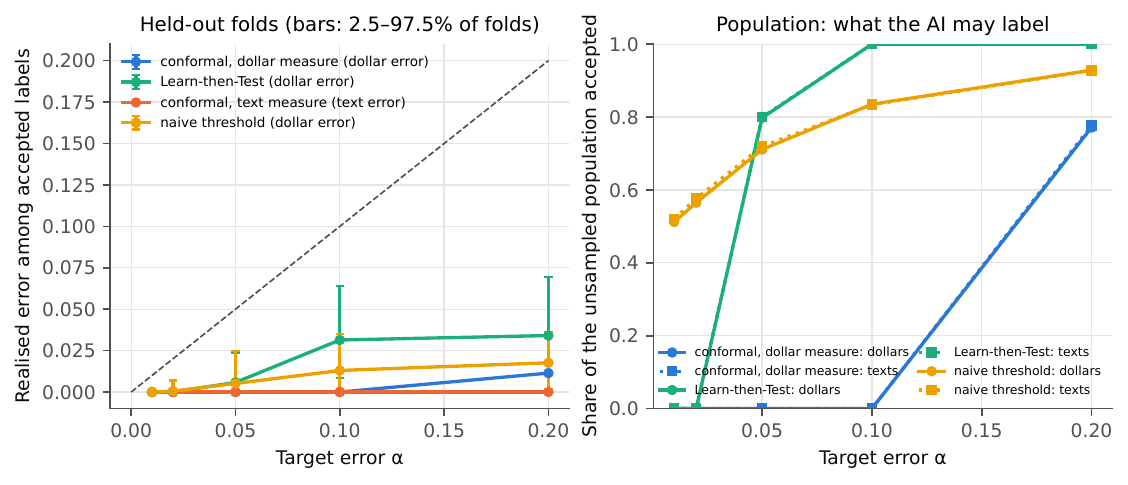}
\caption{Left: realised error among accepted labels against the target $\alpha$ on held-out folds, with the
2.5--97.5\% range over the 100 repeated folds (descriptive; the folds share the same sample). Right: the share of
the unsampled population each rule accepts for machine labelling.}\label{fig:selective}
\end{figure}

The population rules cannot be checked on the population, so they are checked twice: on held-out folds of the
sample, and on re-draws of the design with synthetic truths. In each of 20 repetitions of a 5-fold split the
calibrator is refitted on the four calibration folds alone, cross-fitted within them for their own scores and
with one model fitted on all of them scoring the held-out fold and the unsampled population, so that no held-out
label enters any score, null set or threshold; the calibration folds supply the nulls and the risk estimates, the
population p-values, BH thresholds and Learn-then-Test thresholds are computed exactly as deployed, and the
thresholds are applied to the held-out fold, whose texts enter with their own likelihood-ratio weight and whose
realised error rates are design-weighted so the fold stands for the population. The ridge strength is re-chosen
inside every split on the calibration folds alone (modal value \csNestedModalC, equal to the full-sample choice in
\csNestedShareSameC{} of the splits), so no held-out label touches any model choice. (The calibration scores are
cross-fitted and the held-out scores come from one model, the usual cross-conformal approximation; an unnested
variant that uses the full sample's out-of-fold scores throughout, letting held-out labels shape other texts'
scores, is reported as a diagnostic and gives a held-out error of \csUnnestedLttFiveFdr{} for Learn-then-Test at
$\alpha = 0.05$ against \csLttFiveFdr{} here.) Table~\ref{tab:selective} and Figure~\ref{fig:selective} give the
results for the research-or-not task with the 2.5--97.5\% range of the 100 folds; the folds reuse the same
\csNSample{} labels and their training sets overlap, so these ranges describe the spread over folds and are not
confidence intervals. Three findings.

\emph{The evaluated weighted conformal rule has no power at $\alpha \le 0.10$ and useful power only at $0.20$.} With
\csNNullBinary{} null draws the smallest p-value a text can reach is about one over their number plus one, and the
weighted BH threshold lies below it for the large-mass strata; the rule accepts nothing at
$0.01$, $0.02$, $0.05$ or $0.10$. At $\alpha = 0.20$ the deployed rule, fitted on the whole sample, accepts
\csWbhTwentyDollarShare{} of the unsampled dollars (\csWbhTwentyTextShare{} of texts); on the held-out folds, where
each split's rule is fitted on four fifths of the sample and so has fewer nulls and less power, it accepts
\csWbhTwentyPowerHeld{} of the held-out dollars with a realised dollar error of \csWbhTwentyFdr{} [\csWbhTwentyFdrLo,
\csWbhTwentyFdrHi] and accepts nothing at all in \csWbhTwentyEmpty{} of the folds; the two yields are for different
rules on different populations and are not interchangeable. The slack between target and realised error is large
because the errors concentrate on small, low-confidence texts: the p-values of the accepted texts are far below
the threshold. The rule's power falls as the machine labels improve against the reference, because its nulls are
the sampled texts on which the label is wrong: with reader~1's labels as the reference there were
\csNNullBinaryOne{} nulls and the rule accepted \csWbhTenDollarShareOne{} of the unsampled dollars at
$\alpha = 0.10$.

\emph{Learn-then-Test is the more useful rule here, as an evaluated rule.} At $\alpha = 0.05$ the
Hoeffding--Bentkus version accepts \csLttFiveDollarShare{} of the unsampled dollars (\csLttFiveTextShare{} of
texts, confidence at least \csLttFiveThr) with a realised held-out dollar error of \csLttFiveFdr{} [\csLttFiveFdrLo,
\csLttFiveFdrHi], above $\alpha$ in \csLttFiveExceed{} of the 100 folds (maximum \csLttFiveFdrMax); at
$\alpha = 0.10$ it accepts \csLttTenDollarShare{} with error \csLttTenFdr{} [\csLttTenFdrLo, \csLttTenFdrHi], above
$\alpha$ in \csLttTenExceed{} of folds; at $0.01$ and $0.02$ it accepts nothing, because no threshold's effective
size can rule out a 1\% error. The
bootstrap-p version accepts more (\csLttBootFiveDollarShare{} at $0.05$) with error \csLttBootFiveFdr{}
[\csLttBootFiveFdrLo, \csLttBootFiveFdrHi]; its p-value is asymptotic and we report it as a check. A naive rule
that accepts any label with confidence at least $1-\alpha$ accepts \csNaiveFiveDollarShare{} at $0.05$ with error
\csNaiveFiveFdr; it carries no guarantee and its error is what the calibrator's error is. Unanimity of the three
LLMs, the companion's auto-labelling gate, covers \csUnanShareText{} of texts and \csUnanShareDollar{} of dollars
but its labels are wrong on \csUnanErrTextBin{} of those texts and \csUnanErrDollarBin{} of those dollars, with no
dial to turn.

\emph{The held-out null p-values are consistent with super-uniformity at the reported thresholds; the re-draws
measure the violation probability.} Among the held-out null texts (\csSUn{} fold observations, which reuse the
same sample texts across repetitions), the likelihood-ratio-weighted share with $p \le t$ is \csSUFive{} at
$t = 0.05$, \csSUTen{} at $0.10$, \csSUTwenty{} at $0.20$ and \csSUFifty{} at $0.50$ (Appendix
Table~\ref{tab:superuniformity}, Figure~\ref{fig:superuniformity}). The stress tests of Appendix~\ref{app:stress}
are fixed-score tests: the calibrator and its scores are the real ones and are not refitted, the synthetic labels
of each truth are drawn once and held fixed, and the committed design is re-drawn \csStressDraws{} times, recording
for each rule and $\alpha$ the mean realised error, the share of draws in which a non-empty accepted set has
realised error above $\alpha$ (the violation probability of the Learn-then-Test statement), and the worst draw.
Under the truth drawn from the real scores the mean dollar error of the weighted BH step is \csStressAWbhTen{} at
$\alpha = 0.10$ and the Learn-then-Test rule's \csStressALttFive{} at $\alpha = 0.05$ (violation probability
\csStressALttFiveViol); all-null, both accept nothing at any $\alpha$. Under the truth whose expected error is
about 6\% on every candidate set, just above $\alpha = 0.05$, the Learn-then-Test rule accepts nothing at
$\alpha \le 0.05$ in any draw, and at $\alpha = 0.10$ accepts a set in \csStressDLttTenPAny{} of draws with a
realised error of \csStressDLttTenGivenNonempty, below that target; under the truth that puts the wrong labels on
the largest texts the design can leave unread, it again accepts nothing at $\alpha \le 0.05$ and at $0.10$
accepts a set in \csStressELttTenPAny{} of draws with error \csStressELttTenGivenNonempty{}, because the large
wrong texts are drawn into the PPS calibration sample often enough to raise the observed error at every
threshold. No cell of the table shows a violation in \csStressDraws{} draws, so each cell's violation probability
has a one-sided 95\% upper bound of \csStressLttViolUpMax{} under that cell's simulation law, within the
$\delta = \csDelta$ a valid rule would allow; the bound is per cell and says nothing about other data-generating
mechanisms. These are the numbers the rule should be judged by, and they are what ``evaluated'' means above: a
measured violation probability on this design under these truths, not a bound, and the substitution itself
remains unjustified.

\begin{table}[tbp]\centering\small
\caption{Selective labelling, research-or-not task. Top: realised error of the rule on held-out folds (20 $\times$
5-fold, calibrator and strength refitted on the calibration folds of each split), design-weighted, with the
2.5--97.5\% range over folds (descriptive), the worst fold, and the share of held-out dollars accepted; a dash
means that no fold accepted anything, so the error is undefined, and where only some folds accept nothing such a
fold counts as zero error in the mean (the last column gives their share). Bottom: the share of dollars / texts of the
\csNEligible{} unsampled texts with domain dollars and positive inclusion probability that the rule fitted on the
whole sample accepts at each $\alpha$.}\label{tab:selective}
\adjustbox{max width=\textwidth}{\input{tables/t_selective}}\\[6pt]
\adjustbox{max width=\textwidth}{\input{tables/t_selective_pop}}
\end{table}

For the three-class task, where ``unclear'' is a legitimate answer that must itself be right, the conformal rule
accepts nothing at $\alpha \le 0.10$ and Learn-then-Test accepts nothing at $\alpha \le 0.05$,
\csThreeLttTenDollarShare{} of dollars at $\alpha = 0.10$ and \csThreeLttTwentyDollarShare{} at $0.20$
(Appendix Table~\ref{tab:selective_three}; these are shares accepted by the rule fitted on the whole sample).
On held-out folds most fold-level rules accept nothing at $\alpha = 0.10$ (\csThreeWbhTenEmpty{} of folds for
the conformal rule, \csThreeLttTenEmpty{} for Learn-then-Test), and a fold that accepts nothing scores zero in
the mean; over the folds that accept something the error is \csThreeWbhTenFdrNonEmpty{} and
\csThreeLttTenFdrNonEmpty, against all-fold means of \csThreeWbhTenFdr{} and \csThreeLttTenFdr, and the table gives both. The unclear class has an AUC of \csPanelAUCU{} and the
error among confident three-class labels is too high for either rule at the levels an office would ask for. The estimand needs only research-or-not,
and that is the task an office should ask the rule to control.

\paragraph{No power under the text measure from a dollar-drawn sample.} Under the text measure the rule
accepts \csTextMeasureMaxCert{} population texts at any $\alpha \le 0.2$ (Table~\ref{tab:selective}). The reason is
Proposition~\ref{prop:conformal}: a small-dollar text has a tiny inclusion probability, so its weight $1/\pi_i$
dwarfs the summed weight of the \csNNullBinary{} nulls and its p-value is near one; the smallest p-value any
unsampled text can reach is \csTextMinP, and only \csTextShareBoundBelowTwenty{} of unsampled texts can reach
$0.2$ at all. The sample that makes a dollar guarantee cheap leaves this family of count-controlled rules without
power on this sample; the bound is about these rules and this sample, not about every count-based procedure, and
unique texts are not grant rows. An office that wants error control on \emph{records} rather than dollars needs a
calibration sample drawn with that aim, and this is a concrete design recommendation for the next validation
round.

%% file: sections/acquisition.tex
\section{Budgeted human review as active inference}\label{sec:acquisition}

\subsection{Which text should a person read next?}
Suppose the calibrated probabilities exist and a person can read $b$ more texts. The expansion of
Section~\ref{sec:ppi} says what the label buys: text $i$ enters the estimator's residual with mass
$m_i = w_i d_i$, the dollars it stands for in the population, times its residual $e_i = \ind{y_i = R} - f_i$.
Under Poisson sampling of the review set with inclusion probabilities $q_i$ and expected size $b$ the
Horvitz--Thompson rectifier has variance $\sum_i m_i^2 e_i^2 (1 - q_i)/q_i$, and minimising it subject to
$\sum_i q_i = b$ gives the Neyman allocation \parencite{neyman1934}
\begin{equation}
q_i \propto m_i\, |e_i| \;\approx\; m_i \sqrt{f_i (1 - f_i)} ,
\label{eq:neyman}
\end{equation}
capped at one, where the unobserved $|e_i|$ is replaced by its predictive root-mean-square under a calibrated
$f_i$; if the true conditional probability is $p_i$ the second moment is $p_i(1-p_i) + (p_i - f_i)^2$, so the
replacement under-states the residual exactly on confidently wrong texts. This is the ``active statistical
inference'' design of \textcite{zrnic2024active} with two changes forced by the setting: the mass $m_i$ is the
dollars of every record that shares the text, so a text that covers thousands of rows is worth reading even when
it is only mildly uncertain; and the weights $w_i$ of a first-phase design, if any, enter the mass, which makes
the replay below a second-phase problem on a pseudo-population \parencite{klugerbates2026multiwave}. The estimator
built on the revealed labels with weights $m_i/q_i$ is design-unbiased for the pool total for any $f$
(Proposition~\ref{prop:ppi}(i), the predictor being fixed before the draw). A deterministic policy (read the
largest texts first, the most uncertain first) has inclusion probabilities of zero or one, so without further
assumptions no design-unbiased estimator of the unread total exists for it, and the estimator it is given here is
the plug-in, revealed labels plus predictions elsewhere, which inherits the predictor's bias on what was not read;
the comparison below is therefore with these plug-in policies, not with a practical scheme that reads a
deterministic high-value set and audits the remainder at random, which we did not run. The derivation is for
Poisson sampling; the replay draws a fixed-size PPS sample on the same measure
(the committed routine, whose empirical inclusion frequencies match their targets to within \csPpsMaxDev{} on an
enumerated frame; Appendix~\ref{app:stress}) and uses the Poisson-type variance estimator $\sum_{i \in \mathrm{read}}
(m_i e_i)^2 (1 - q_i)/q_i^2$ with a normal interval; both are approximations whose consequences the replay measures.

\subsection{Replay against hidden human labels}
The human sample is the pool: the \csNSample{} texts with masses $m_i$ stand for the population as a
pseudo-population, and the target is its Horvitz--Thompson total $T^* = \$\csTstarB$B (the post-stratified classic
estimate of Section~\ref{sec:ppi} is \$\csClassicB{}B; the replay is internally consistent with the HT form).
Design-based policies (uniform; dollar-PPS, a single unstratified draw proportional to the mass $m_i$ each text
stands for, not the companion's stratified design; the active design \eqref{eq:neyman}; a 50/50 mixture of the
active and dollar-PPS measures; and a 50/50 mixture of the active and uniform measures, which bounds $1/q_i$
whatever the masses) draw $b$ texts by fixed-size PPS on their measure, capped at one and renormalised, and
estimate with the HT rectifier; greedy policies reveal the top-$b$ texts by their score and estimate by plug-in
(appendix Table~\ref{tab:acqgreedy}). The same comparators appear under the same names in
Tables~\ref{tab:acqprot}--\ref{tab:adaptive}.

Two replays are needed. In the \emph{in-pool} replay the predictions are the calibrator's out-of-fold
probabilities on the same pool, so each fold model saw 90\% of the pool's labels: this is a conditional diagnostic
of the designs given a good predictor, not an experiment an office could run, and its plug-in is close to the target
(\$\csPluginCalB{}B against $T^*$) for that reason. In the \emph{protected} replay the pool is split by stratum,
the calibrator is fitted on two thirds and scored on the other third, whose labels enter none of its coefficients (its ridge strength alone is the
inherited constant of Section~\ref{sec:adaptive}), and the
policies are replayed on that third against its own total, averaged over \csProtSplits{} splits of
\csProtReps{} replays each (about \csProtSize{} texts per replay pool); the two thirds are the pilot an office would
have to read first, and count in its budget. The in-pool replay is also run with the unsupervised DS posterior as
the predictor every policy must live with.

\begin{figure}[tbp]\centering
\includegraphics[width=0.98\textwidth]{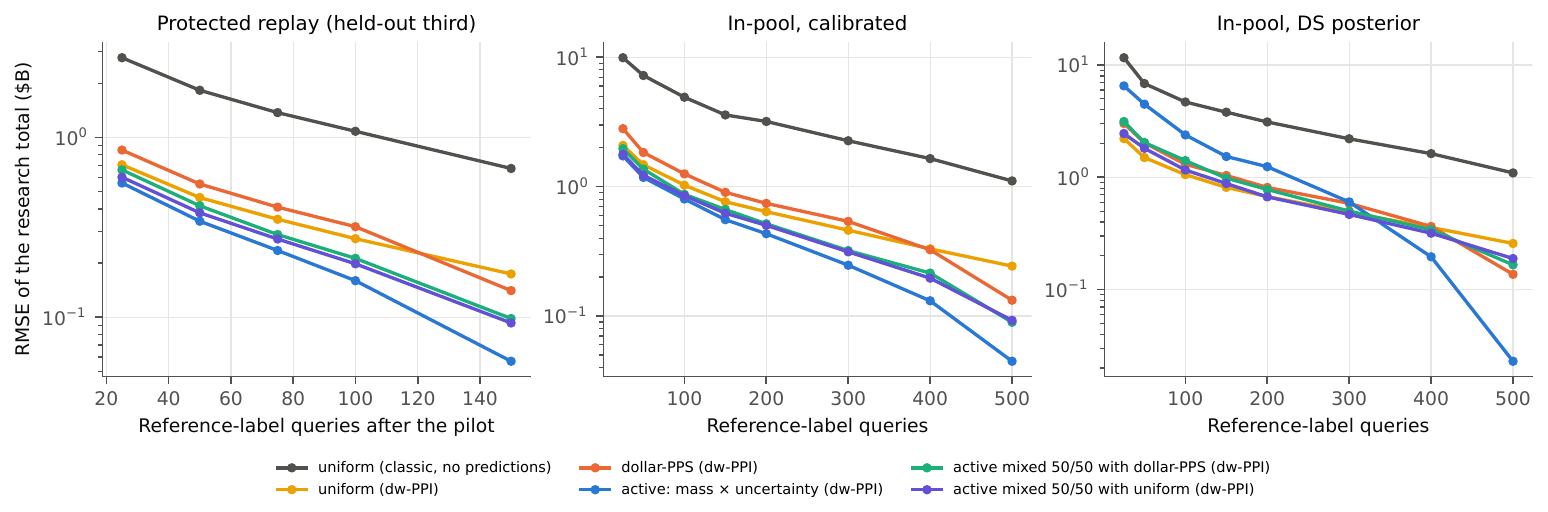}
\caption{RMSE of the research total against the number of reference-label queries. Left: protected replay
(calibrator fitted on two thirds of the pool, policies replayed on the other third, \csProtSplits{} splits; the
budget counts the queries after that separate pilot). Middle: in-pool
replay with the cross-fitted calibrator. Right: in-pool replay with the unsupervised DS posterior.}\label{fig:acq}
\end{figure}

\begin{table}[tbp]\centering\small
\caption{Protected replay: RMSE of the research total (\$B) and coverage of the 95\% interval, by policy and
budget, over \csProtSplits{} splits $\times$ \csProtReps{} replays. The budget counts reference-label queries on
the replay third; the pilot on the other two thirds is separate and is not in it.}\label{tab:acqprot}
\adjustbox{max width=\textwidth}{\input{tables/t_acquisition_protected}}
\end{table}

Table~\ref{tab:acqprot} and Figure~\ref{fig:acq} give the protected results. With a predictor whose coefficients never saw the
pool's labels, the active design has RMSE \$\csProtActiveHundred{}B at 100 texts against \$\csProtPpsHundred{}B for
dollar-PPS, \$\csProtUnifHundred{}B for uniform dw-PPI and
\$\csProtClassicHundred{}B for uniform sampling with no predictions: a factor of \csProtRatioHundred{} over the
dollar design at 100 texts and \csProtRatioFifty{} at 50. Paired across splits, the ratio of the dollar-PPS RMSE
to the active RMSE at 100 texts is \csPairRatioHundred{} (s.d.\ \csPairRatioSdHundred{} over splits, the active
design better in \csPairShareHundred{} of them). The in-pool figures (appendix Table~\ref{tab:acqcal}) are
similar, \csRmseCalActiveHundred{} against \csRmseCalPpsHundred{} at 100 texts: the circularity of the in-pool
replay changes the predictor's quality less than one might fear, because the calibrator's inputs are mostly the
readers' votes. The protected result at 100 texts is conditional on a separate pilot of two thirds of the sample
and is on a smaller pool than the designs of Section~\ref{sec:adaptive}; the two RMSEs are not compared
directly.

\paragraph{Coverage, and why the pure active design loses it.} The active design's interval covers the pool total
at \csProtCovActiveFifty{} (50 texts), \csProtCovActiveHundred{} (100) and \csProtCovActiveHundredFifty{} (150)
in the protected replay, and in the in-pool replay at \csCovCalActiveHundred{} (100) and \csCovCalActiveFiveHundred{}
(500). The diagnosis is in appendix Table~\ref{tab:acqdiag}: the Neyman measure gives near-zero $q_i$ to every
text whose $f_i$ is near 0 or 1, and such a text, when its label is wrong, carries a residual $m_i e_i$ that the
variance estimator never sees because the text is almost never drawn, while at large budgets a growing share of
texts sit at the cap $q_i = 1$ and the Poisson-type variance formula no longer describes the draw. The interval is
then too narrow for a heavy-tailed estimator, the standard failure of inverse-probability weights without a floor
\parencite{zrnic2024active}. The mixtures improve coverage in these replays: the 50/50 mixture with dollar-PPS
covers at \csProtCovMixedHundred{} (100 texts) with RMSE \$\csProtMixedHundred{}B, and the mixture with uniform,
which bounds $1/q_i$ by $2n/b$, covers at \csProtCovMixedUnifHundred{} with RMSE \$\csProtMixedUnifHundred{}B;
at 150 texts the two cover at \csProtCovMixedHundredFifty{} and \csProtCovMixedUnifHundredFifty, still short of
nominal, because the Poisson-type variance estimator does not describe a fixed-size draw with many texts at the
cap.
These coverages are over \csProtSplits{} splits of \csProtReps{} replays each, and the replays within a split
share its calibrator and its pool, so their uncertainty is that of \csProtSplits{} dependent units, not of
independent runs. The practical design is a mixture, and the price of its floor is the ratio of those RMSEs to the
pure design's. The equal shares of the two mixtures are a convention and were not tuned: no other share was run, so
these replays say that a floor helps on this design, not how small it can be made before coverage is lost.

\begin{table}[tbp]\centering\small
\caption{In-pool replay with the unsupervised DS posterior as the predictor: RMSE (\$B) by policy and budget
(\csAcqReps{} replays).}\label{tab:acqds}
\adjustbox{max width=\textwidth}{\input{tables/t_acquisition_ds}}
\end{table}

With the uncalibrated predictor the picture reverses and is the more instructive (Table~\ref{tab:acqds},
Figure~\ref{fig:acq} right). The DS posterior is nearly binary, so $\sqrt{f(1-f)}$ is near zero on most texts and
the active design concentrates its budget on a few texts; the over-confident wrong labels it skips then enter the
rectifier with huge weights, and its RMSE at 100 texts (\$\csRmseDsActiveHundred{}B) is worse than dollar-PPS
(\$\csRmseDsPpsHundred{}B), with coverage \csCovDsActiveHundred{} there and between \csCovDsActiveMin{} and
\csCovDsActiveMax{} across budgets. The mixture with dollar-PPS removes most of the
loss (\$\csRmseDsMixedHundred{}B) without restoring the active design's advantage, at a cost of
\csRmseCalMixedHundred{} against \csRmseCalActiveHundred{} when the predictor is calibrated. The practical rule is the one the sections together support: calibrate first
(Section~\ref{sec:aggregation}), then read by mass times predicted error, and keep a share of the budget on a
mass-proportional or uniform draw so that no text is unreachable.

\subsection{Why selective acceptance can damage total estimation}
One might hope to chain the tasks: texts accepted by the selective rule of Section~\ref{sec:selective} take their
machine label; the remainder is read by the active design with the budget $b$; the total is the accepted dollars
with their machine labels plus the dw-PPI estimate of the remainder. Its error has two parts: the error dollars
among accepted texts, a bias that the acceptance rule aims to hold below a share $\alpha$ of the accepted
dollars but, as Section~\ref{sec:selective} says, does not certify, and the sampling error of the remainder,
which the design controls. The interval of the remainder ignores the first part altogether, and
Table~\ref{tab:hybrid} shows what that costs. The table aggregates over the \csHybSplits{} protected splits: in
each split the accept rule is Learn-then-Test fitted on the calibration part's PPS draws, the accepted set is
fixed, and the active design is replayed \csProtReps{} times on the remainder; across splits the accepted set,
the remainder and the accepted part's signed error differ, so the queries used, the remainder and the bias are
means over splits (the remainder ranges from \csHybFiveRemainingMin{} to \csHybFiveRemainingMax{} texts at
$\alpha = 0.05$ and from \csHybTenRemainingMin{} to \csHybTenRemainingMax{} at $0.10$) and the RMSE is the root
mean square over splits of the split-level RMSE. Every split is kept: where the rule accepts the whole replay
pool (\csHybTenAllAccepted{} of the \csHybSplits{} splits at $\alpha = 0.10$, \csHybFiveAllAccepted{} at $0.05$)
nothing is left to read, the estimate is the accepted dollars with their machine labels, its error is the
accepted part's signed error and its interval has no width, so it covers only if that error is exactly zero. Acceptance at $\alpha = 0.05$ takes on average
\csHybFiveAccepted{} of the replay pool's texts (\csHybFiveAcceptedDollars{} of its dollars) with a realised error
of \csHybFiveAcceptedErr{} among them and a signed error of \$\csHybFiveBiasB{}B on average (root mean square
\$\csHybFiveBiasRmsB{}B over splits), leaving \csHybFiveRemaining{} texts; reading 50 of them by the active
design gives RMSE \$\csHybFiveRmseFifty{}B against \$\csProtActiveFifty{}B for reading 50 of the whole pool,
little gain in precision, because the accepted texts are the confident ones whose residuals are small and whose
draws the active design already avoids. The coverage of the nominal 95\% interval is \csHybFiveCovFifty{} at 50
texts and falls as the remainder's own error shrinks; at $\alpha = 0.10$, where the accepted share is
\csHybTenAcceptedDollars{} with error \csHybTenAcceptedErr, it is at most \csHybTenCovMax{} at any budget and
zero from \csHybTenCovZeroFrom{} texts on. When the budget
covers the whole remainder, every text is read, the remainder's error is zero within a split and the total's error
is the accepted part's signed error, so the RMSE settles at that error's root mean square over splits,
\$\csHybTenBiasRmsB{}B at $\alpha = 0.10$, not at the absolute value of its mean. A combined interval has to carry
both parts. If the accepted set's error dollars were bounded by $\alpha D_A$
with probability $1 - \delta$, $D_A$ its known dollar mass, and the remainder's interval $[L_R, U_R]$ held with
probability $1 - \beta$, then $[\hat T_A + L_R - \alpha D_A,\ \hat T_A + U_R + \alpha D_A]$ would cover with
probability at least $1 - \delta - \beta$ by the union bound, which is 85\% at $\delta = 0.10$ and $\beta = 0.05$,
and only if the first bound were valid, which the rule of Section~\ref{sec:selective} does not establish; the
alternative is to audit a probability sample of the accepted texts as well and estimate the accepted part's
signed error rather than treat it as exact. We therefore keep the hybrid as a negative result, not as a pipeline:
accepting labels can invalidate an otherwise useful interval, and selective labelling does not by itself improve
the total, because the active design already spends little on the texts a rule would accept. In panel~B of the
table the queries used fall short of $b$ in the splits whose remainder is smaller than $b$, so their mean is below $b$
from $b = \csHybFiveShortFrom$ on at $\alpha = 0.05$ and from $b = \csHybTenShortFrom$ on at $0.10$; the signed error of the accepted part is the realised
error of each split's fixed accepted set, which does not depend on the budget, so panel~A gives it once for each threshold.

\begin{table}[!tb]\centering\small
\caption{Why accepting labels can invalidate an otherwise useful interval: the hybrid estimator in the protected
replay, accepting at $\alpha$ by Learn-then-Test ($\delta = 0.1$), reading $b$ of the remaining texts by the
active design and estimating the total as accepted dollars plus dw-PPI on the remainder, whose interval ignores
the accepted part's error. Every entry aggregates the same \csHybSplits{} protected splits (\csProtReps{} replays
each), including the splits in which the rule accepts the whole pool, where nothing is read, the error is the
accepted part's signed error and the interval has no width. Panel~A: the accepted set is fixed within a split and
differs across splits, so its entries are means over splits; the error share is the realised dollar error among
the accepted texts, and the RMS is the root mean square over splits of the accepted part's signed error. Panel~B:
queries used (reference labels revealed on the remainder, after the pilot) and coverage are means over splits,
and the RMSE is the root mean square over splits, which equals the RMS of panel~A once the whole remainder is
read.}\label{tab:hybrid}
\input{tables/t_hybrid}
\end{table}

\subsection{With every label charged: two-phase and batch-sequential designs}\label{sec:adaptive}
The protected replay charges the calibrator's pilot to the office but still gives it two thirds of the sample.
The harder question is nearer to the one an office faces: with no labels at all, how should it split a budget
between the labels that teach the calibrator and the labels the design then chooses? One unit of budget in this
replay is one previously hidden reference label revealed to the policy, and the pilot's labels are counted like
any other. What the count leaves out is the cost of producing that label: here each reference label rests on
three independent readings, and on \csNSplit{} of the \csNSample{} texts on a joint adjudication as well. Three
readings per text would multiply every policy's cost by the same factor and leave the comparison unchanged; the
adjudication would not, because a policy that seeks out uncertain texts may draw more of the texts readers split
on, and we did not cost it. A revealed label is also the recorded reference, not a fresh single reading that
could differ from it. The replay therefore compares policies with equal access to the same reference at equal
query counts; it does not estimate a saving in readings, hours or adjudication meetings. We replay three families on
the full 651-text pool, every label hidden until a policy reads it, so that the only predictor a policy has is the
calibrator fitted on the labels it has already bought. The information a policy may use is stated exactly. The
\csAdFeatures{} features (the votes, the unsupervised posteriors fitted on all $N$ machine readings, and the text
cues) carry no human label and are available for every text; the feature scaler is fitted on the texts the policy
has read; the ridge strength is a constant, $C = \csRidgeCMain$, the same for every policy and never tuned
inside a replay. The constant is not free of the pool's labels, however: it is the value the nested
cross-validation of Section~\ref{sec:aggregation} chose on the full sample when reader~1 was the reference
(under the adjudicated reading the same procedure chooses \csCfinal). The replays, the protected one included,
took the constant from that choice and have kept it. The results below are conditional on that specification, and a
sensitivity analysis at the end of this subsection shows what they owe to it. With fewer than five labels, or labels of one class, the calibrator is the constant mean
label; the first phase has at least ten texts; phase sizes are rounded to integers and the batches of the
sequential design differ by at most one text; texts already read are excluded from later draws; each wave's
inclusion probabilities are the fixed-size PPS probabilities on the wave's measure, capped at one and
renormalised, and a text with no domain dollars has a negligible measure and contributes nothing to the total. A
\emph{two-phase} design spends a share $\phi$ of the budget on a dollar-PPS first phase, fits the calibrator on
it, and draws the rest from the unread texts by the active measure on that calibrator; a \emph{batch-sequential}
design does the same in \csAdK{} equal batches, refitting after each; the \emph{classic} design is $\phi = 1$.
The estimator is written for one wave at a time. Before wave $t$ let $O_t$ be the texts already read and $U_t$
the rest; the calibrator $f_t$ is fitted on the labels of $O_t$ only, the wave $B_t \subset U_t$ is drawn with
inclusion probabilities $q_{it}$ that are a function of $f_t$ and the masses, fixed before any label in $B_t$ is
revealed, and the estimate and its variance estimator are
\begin{align}
\hat T_t &= \sum_{i \in O_t} m_i\, \ind{y_i = R} + \sum_{i \in U_t} m_i f_t(i) + \sum_{i \in B_t}
\frac{m_i\,(\ind{y_i = R} - f_t(i))}{q_{it}}, \label{eq:wave}\\
\hat V_t &= \sum_{i \in B_t} \frac{m_i^2\,(\ind{y_i = R} - f_t(i))^2\,(1 - q_{it})}{q_{it}^2}. \notag
\end{align}
The estimate is design-unbiased for the total conditionally on $O_t$ and its labels, because $f_t$ and
$q_{\cdot t}$ are fixed given them and every unread text with domain dollars has $q_{it} > 0$
(Proposition~\ref{prop:ppi}(i); the multiwave setting of \textcite{klugerbates2026multiwave}). The labels of
$B_t$ enter the calibrator only after $\hat T_t$ is computed, and the two-phase design is the single-wave case
with $O_2$ the first phase. For the batch design every batch yields such an estimate; their errors are martingale
differences under this filtration, so the equal-weight average of the $K - 1$ wave estimates has variance equal
to the sum of the expected conditional variances divided by $(K-1)^2$, with no cross terms, and we estimate it by
$\sum_t \hat V_t/(K-1)^2$, a Poisson-type approximation to the fixed-size draw. We report both the last-batch and
the averaged estimator with normal intervals from these estimators, whose coverage the replay reports. A
reference line uses the in-pool cross-fitted calibrator, whose labels are not charged, so it is the gain a free
pilot would give.

\begin{figure}[tbp]\centering
\includegraphics[width=\textwidth]{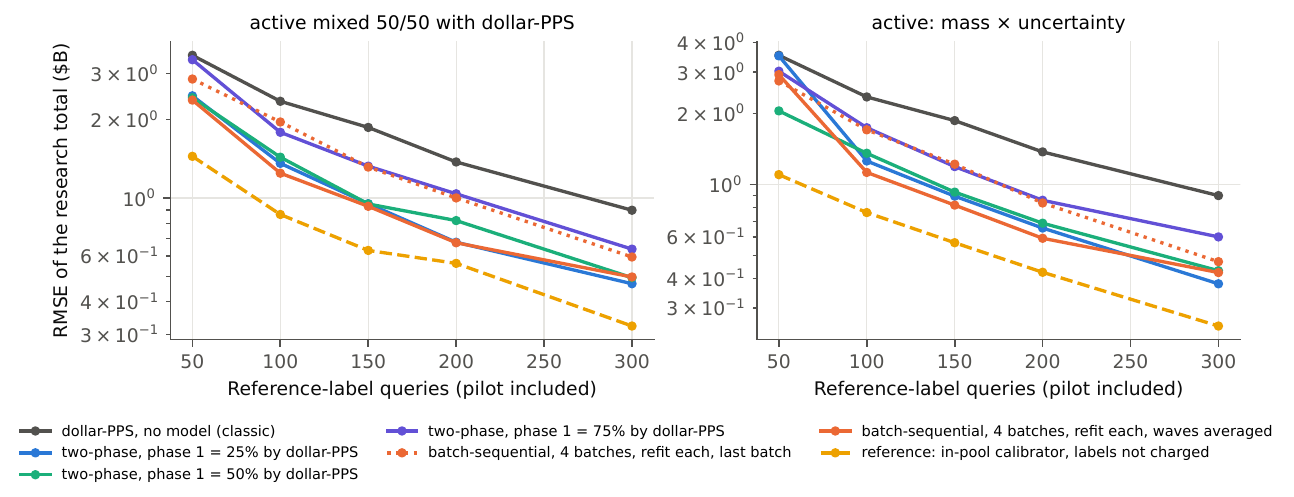}
\caption{RMSE of the research total against the budget of reference-label queries (pilot included) for the
designs with every label charged, with the classic design and the free-pilot reference; left with the mixed measure in the later phases, right with the pure
active measure.}\label{fig:adaptive}
\end{figure}

Table~\ref{tab:adaptive} and Figure~\ref{fig:adaptive} give the results. At 100 texts the classic design has RMSE
\$\csAdClassicHundredRmse{}B and the reference \$\csAdRefMixedHundredRmse{}B; the two-phase design with a quarter of
the budget in the first phase reaches \$\csAdTwentyFiveMixedHundredRmse{}B with coverage
\csAdTwentyFiveMixedHundredCov, half with \$\csAdFiftyMixedHundredRmse{}B and three quarters with
\$\csAdSeventyFiveMixedHundredRmse{}B; at 200 texts the quarter split gives \$\csAdTwentyFiveMixedTwoHundredRmse{}B
against \$\csAdClassicTwoHundredRmse{}B classic and \$\csAdRefMixedTwoHundredRmse{}B for the free pilot. At a
common total budget, with every label charged, the classic design's RMSE is therefore
\csAdTwoPhaseVsClassicHundred{} times the two-phase design's at 100 texts (a reduction of
\csAdTwoPhaseVsClassicHundredPct; Monte Carlo s.e.\ of the ratio \csAdTwoPhaseVsClassicHundredSe{} from
\csAdReps{} independent replays per policy) and \csAdTwoPhaseVsClassicTwoHundred{} times at 200
(\csAdTwoPhaseVsClassicTwoHundredPct), and \csAdBatchAvgVsClassicHundred{} and
\csAdBatchAvgVsClassicTwoHundred{} times the averaged batch design's; this, not the free-pilot ratio of
Table~\ref{tab:acqprot}, is the comparison at a common budget of reference-label queries. Appendix Table~\ref{tab:adaptive_detail} gives the bias,
the mean interval width and the coverage beside each RMSE. The best first-phase share is \csAdBestPhiFifty{} at
50 texts and \csAdBestPhiHundred{} at 100 and \csAdBestPhiTwoHundred{} at 200, so it depends on the budget, and a
quarter is what this pool, these budgets and this calibrator favour rather than a rule; a calibrator fitted on
25 labels is already good enough for the design to use, because its inputs are mostly the readers' votes. The
batch-sequential design with the last-batch estimator is worse than the two-phase design
(\$\csAdBatchLastMixedHundredRmse{}B at 100 texts) because only the last batch rectifies; the averaged estimator
(\$\csAdBatchAvgMixedHundredRmse{}B) uses every batch, has the lowest RMSE of the mixed designs with every label
charged at small budgets and is level with the quarter split from 200 texts on; its interval covers at only
\csAdBatchAvgMixedFiftyCov{} at 50 texts, the lowest coverage in Table~\ref{tab:adaptive}, so the lowest RMSE is
not the design to prefer where an interval is to be reported at a small budget. With \csAdReps{} replays a
coverage near 0.94 has a Monte Carlo standard error of about \csAdCovMcSe, so a coverage of 0.93 or 0.94 in the
table is not evidence of under-coverage, and \csAdBatchAvgMixedFiftyCov{} is. We read this as the calibrator having learned
most of what the features carry by about fifty labels on this benchmark; the ratios alone do not identify a
learning threshold. The price of having no pilot is the ratio of the two-phase RMSE to the reference, about
\csAdNoPilotRatioHundred{} at 100 texts and \csAdNoPilotRatioTwoHundred{} at 200; the classic design pays
\csAdClassicRatioHundred{} and \csAdClassicRatioTwoHundred. Re-run under each reader and under their majority (Appendix
Table~\ref{tab:reference_rerun_sections}), the price of having no pilot is
\csRerunAdNoPilotHundredMin{}--\csRerunAdNoPilotHundredMax{} at 100 texts and
\csRerunAdNoPilotTwoHundredMin{}--\csRerunAdNoPilotTwoHundredMax{} at 200, the averaged batch-sequential design
pays \csRerunAdBatchAvgHundredMin{}--\csRerunAdBatchAvgHundredMax{} at 100, and the classic design pays
\csRerunAdClassicHundredMin{}--\csRerunAdClassicHundredMax{} and
\csRerunAdClassicTwoHundredMin{}--\csRerunAdClassicTwoHundredMax; the two-phase design beats the classic one at
both budgets under \csRerunAdRankingHolds{} one of them.

\paragraph{What the result owes to the ridge strength.} Because the constant $C = \csRidgeCMain$ was taken
from a choice made on this sample, Appendix Table~\ref{tab:adaptive_ridge} repeats the quarter-split design at
every strength of the grid the main calibrator is tuned over ($C \in \{\csRidgeGrid\}$, a small $C$ being a
strong penalty), and once more with the strength chosen by cross-validation on the policy's own first-phase
labels, which uses nothing outside what the policy has bought; \csRidgeReps{} replays per cell on random streams
independent of Table~\ref{tab:adaptive}, with the classic design re-run on the same footing, and every setting
reported. At 100 queries the reduction in RMSE against the classic design is \csRidgeMainHundredPct{} at
$C = \csRidgeCMain$ (\$\csRidgeMainHundredRmse{}B against \$\csRidgeClassicHundredRmse{}B; the
\csAdReps-replay figure of Table~\ref{tab:adaptive} is \csAdTwoPhaseVsClassicHundredPct), between
\csRidgeModerateHundredMinPct{} and \csRidgeModerateHundredMaxPct{} for $C$ from 0.03 to 1, and
\csRidgeTunedHundredPct{} with the strength chosen on the first-phase labels; at 200 queries the three figures
are \csRidgeMainTwoHundredPct, \csRidgeModerateTwoHundredMinPct{}--\csRidgeModerateTwoHundredMaxPct{} and
\csRidgeTunedTwoHundredPct. Under a strong penalty the gain shrinks, to \csRidgeStrongHundredPct{} at
$C = 0.01$ and \csRidgeVeryStrongHundredPct{} at $C = 0.003$ at 100 queries: a calibrator shrunk that far on 25
labels is close to a constant, and the design it drives is close to the classic one. Under the adjudicated
reading the constant is therefore not what produces the gain, which a policy choosing its own penalty also
obtains, but it sits in the favourable part of the grid, and the result is conditional on a penalty that lets
the calibrator move. Coverage stays between \csRidgeCovMin{} and \csRidgeCovMax{} across the two-phase cells of
the table. The same table under each reader and under their majority is less kind to the constant. At 100
queries the reduction at $C = \csRidgeCMain$ is \csRerunRidgeMainHundredMin{}--\csRerunRidgeMainHundredMax{}
across the four, and with the strength chosen on the first-phase labels it is
\csRerunRidgeTunedHundredMin{}--\csRerunRidgeTunedHundredMax, lower than the constant's under every one of
them and lowest under reader~2 (\csRerunRidgeTwoTunedHundred{} against \csRerunRidgeTwoMainHundred); at 200
queries the two ranges are \csRerunRidgeMainTwoHundredMin{}--\csRerunRidgeMainTwoHundredMax{} and
\csRerunRidgeTunedTwoHundredMin{}--\csRerunRidgeTunedTwoHundredMax. Under readers~1 and~2 the constant is the
best fixed setting of the grid at 100 queries. The
two-phase design still beats the classic one under every reading with either rule, but under the individual
readers part of the margin reported for the constant is owed to the constant. We did not vary it in the
protected replay.

\begin{table}[tbp]\centering\small
\caption{Designs with every label charged on the 651-text pool, every label hidden until read: RMSE of the
research total (\$B) and coverage of the 95\% interval in parentheses, by total budget of reference-label queries
(the first phase or first batch included), \csAdReps{} independent
replays per policy and budget (Monte Carlo standard errors, bias and interval width in Appendix
Table~\ref{tab:adaptive_detail}).}\label{tab:adaptive}
\input{tables/t_adaptive}
\end{table}

\paragraph{Review time as the budget.} An office pays in reading time, not in texts. With the word count of a text
as its cost (mean \csAdWordsMean{} words, median \csAdWordsMedian), the cost-aware allocation under Poisson
sampling is $q_i \propto m_i\sqrt{v_i / c_i}$ (Appendix~\ref{app:proofs}). Appendix Table~\ref{tab:adaptive_cost}
compares the policies at equal expected words read, with the in-pool calibrated predictor. At the word budget of
100 dollar-PPS texts (\csAdCostBudgetHundred{} words) the cost-adjusted active design reads
\csAdCostMixedCostHundredTexts{} texts for an RMSE of \$\csAdCostMixedCostHundredRmse{}B, against
\$\csAdCostMixedHundredRmse{}B for the count-based active design (\csAdCostMixedHundredTexts{} texts) and
\$\csAdCostPpsHundredRmse{}B for dollar-PPS; at twice the budget the figures are \$\csAdCostMixedCostTwoHundredRmse{}B
against \$\csAdCostMixedTwoHundredRmse{}B. The adjustment buys about a tenth of the RMSE by reading half again as
many, shorter, texts for the same words; it would matter more where long texts carried the dollars, and the cost
here is a proxy, since no reading times were recorded.

\paragraph{Several questions at once.} The yearly totals of Section~\ref{sec:ppi} are five more questions on the
same sample. The total-only Neyman allocation is optimal for one of them; the allocation that minimises the
largest variance ratio across the six questions (each question's anticipated variance relative to its own optimal
allocation, the ratio being the price the question pays for sharing the sample; multiplicative weights on the
question weights, Appendix Table~\ref{tab:adaptive_multi}) trades a little precision on the total for the worst
year. At 100 texts the total-only allocation gives the total a relative RMSE of \csAdMultiHundredTotTotalRel{} and
the worst year (\csAdMultiHundredTotWorstYear) \csAdMultiHundredTotMaxYearRel; the minimax allocation gives
\csAdMultiHundredMmTotalRel{} and \csAdMultiHundredMmMaxYearRel{} (at 200 texts,
\csAdMultiTwoHundredTotTotalRel{} and \csAdMultiTwoHundredTotMaxYearRel{} against \csAdMultiTwoHundredMmTotalRel{}
and \csAdMultiTwoHundredMmMaxYearRel\csAdMultiTwoHundredNote). The minimax allocation is found by sixty multiplicative-weights iterations on
the anticipated variance ratios, keeping the best iterate; the objective is the largest anticipated ratio, so the
anticipated ratios need not be equal at the chosen allocation and the realised relative RMSEs in the table, which
are Monte Carlo quantities, need not be either. The anticipated variance ratios say why the gain is modest: under
the total-only allocation the worst year pays \csAdMultiHundredTotMaxRatio{} times its own optimum at 100 texts,
but the yearly dollars sit largely on the same large texts as the total, so the realised errors move less than
the anticipated ones. Where the questions conflict more (an institution against the nation, say) the same machinery
applies, and the small-area results of Section~\ref{sec:downstream} show how far one sample can be stretched
before an interval is no longer honest.

%% file: sections/invariance.tex
\section{Sensitivity of the machine readings to instruction, context and version}\label{sec:invariance}

A machine reading is an instrument. How much of what it reports is the text, and how much is the instruction, the
context shown beside the text, the run, or the composition of the year? The companion listed these as untested.
We test them on the human-labelled sample, so that every comparison can also be weighted to the population and
scored against the reference, using the three panel models with the exact settings of the population run
(Claude Sonnet~5.5, reasoning low; Gemini 2.5~Flash, thinking off; GPT-5~mini, reasoning low): \csInvCalls{}
calls, \csInvTokensIn{}M input tokens, run on 30--31~October 2026 (Appendix~\ref{app:prompts} gives the
wrappers). What follows is a sensitivity study of named perturbations, each run once, together with one repeat of
the population run's own wrapper. The repeat is not of the identical prompt: every sensitivity run, the repeat
included, inserts codebook version~1.2, and the population run inserted version~1.1 (Section~\ref{sec:testbed}).
With one run per variant and one repeat, the study cannot identify how much of
the variation is attributable to wording, to the codebook version, to generation randomness or to changes in the
deployed model; it
reports the observed repeat and wrapper sensitivities separately, and it is not a general invariance claim. Every comparison with the human reference below is against the adjudicated reading; Appendix
Table~\ref{tab:reference_rerun_sections} repeats the reference-dependent figures under each reader and their
majority. Across the three readers and their majority the population reading agrees with the reader on
\csRerunInvSonnetHumanWdMin{}--\csRerunInvSonnetHumanWdMax{} of dollars for Sonnet,
\csRerunInvFlashHumanWdMin{}--\csRerunInvFlashHumanWdMax{} for Flash and \csRerunInvMiniHumanWdMin{}--\csRerunInvMiniHumanWdMax{}
for mini, and the calibrated panel's bias by year stays within \csRerunDriftCalAbsMax{} points of the year's
dollars under each of them.

\paragraph{Instruction variants.} Four wrappers around the unchanged codebook differ in emphasis as well as
wording (a plain instruction; an ``annotator in a measurement study'' framing that stresses the codebook's
authority; a terse one; a three-step decomposition), so they are instruction variants rather than paraphrases.
Against the original reading, Gemini Flash gives the same status on
\csParaFlashSameTextsMin{}--\csParaFlashSameTextsMax{} of texts and
\csParaFlashSameDollarsMin{}--\csParaFlashSameDollarsMax{} of dollars per variant, and all \csParaFlashNPrompts{}
readings agree on \csParaFlashAllSameTexts{} of texts (\csParaFlashAllSameDollars{} of dollars); GPT-5 mini (two
variants) gives \csParaMiniSameTextsMin{}--\csParaMiniSameTextsMax{} and all \csParaMiniNPrompts{} agree on
\csParaMiniAllSameTexts; Claude Sonnet (the same two variants) gives
\csParaSonnetSameTextsMin{}--\csParaSonnetSameTextsMax{} of texts and
\csParaSonnetSameDollarsMin{}--\csParaSonnetSameDollarsMax{} of dollars, with all \csParaSonnetNPrompts{} agreeing
on \csParaSonnetAllSameTexts. By texts, agreement with the human reference varies across wrappers
(\csParaFlashHumanMin{}--\csParaFlashHumanMax{} for Flash against \csOrigFlashHuman{} for the original), because
the variants move small texts between ``unclear'' and ``not research''; by dollars it is stable
(\csOrigFlashHumanDollars{} for the original, \csMajFlashHumanDollars{} for the majority over wrappers). The
majority over wrappers reaches \csMajFlashHuman{} (Flash), \csMajMiniHuman{} (mini) and \csMajSonnetHuman{}
(Sonnet) by texts, no better than the best single reading of the same model, so aggregating over wrappers buys nothing here and costs a multiple of the calls; this differs
from \textcite{liu2026interprompt}, where majority over prompts helps on texts, and the likely reason is the one
this paper keeps meeting, that the dollars sit on texts where the wrappers agree. The decomposition of
Table~\ref{tab:decomposition} makes the size explicit: for the research indicator over (text, wrapper) pairs, the
within-text share of the variance, the variation across the tested readings of the same text, is
\csWithinFlashTexts{} [\csWithinFlashTextsLo,
\csWithinFlashTextsHi] by texts and \csWithinFlashDollars{} [\csWithinFlashDollarsLo, \csWithinFlashDollarsHi] by
dollars for Flash (\csWithinMiniTexts{} and \csWithinMiniDollars{} for mini, \csWithinSonnetTexts{} and
\csWithinSonnetDollars{} for Sonnet).

\paragraph{The same comparisons inside one codebook version.} The variants and the no-context reading inserted
codebook version~1.2 and the population reading version~1.1, so the comparisons above carry the change of
version as well as the change of wrapper. The repeat of the population wrapper shares version~1.2 and the model
settings with the variants, and Appendix Table~\ref{tab:invariance_within} takes it as the baseline instead, on
the texts every reading parsed and with the same weights. Against it Gemini Flash gives the same status on
\csParaVFlashSameTextsRange{} of texts and
\csParaVFlashSameDollarsRange{} of dollars per variant, GPT-5 mini on
\csParaVMiniSameTextsRange{} of texts and
\csParaVMiniSameDollarsRange{} of dollars, and Claude Sonnet on
\csParaVSonnetSameTextsRange{} and
\csParaVSonnetSameDollarsRange; the research decision is unchanged on at least
\csParaVMinSameRDollars{} of dollars under every variant of every model, and the table counts the texts each
variant moves into and out of research. The within-text share of variance over the readings under version~1.2
is \csWithinVFlashDollars{} by dollars for Flash, \csWithinVMiniDollars{} for mini and \csWithinVSonnetDollars{}
for Sonnet (Table~\ref{tab:decomposition}). These figures are of the same size as those against the population
reading, so the wrapper sensitivity reported above is not produced by the mismatch of codebook versions. Each
variant was still run once, so generation variability is inside every one of these comparisons, and the
comparison of the population reading under version~1.1 with its repeat under 1.2 remains the separate, combined
run-and-version diagnostic of the paragraph below.

\begin{table}[tbp]\centering\small
\caption{Variance decomposition of the research indicator over (text, reading) pairs. The readings compared are
the wrappers with the population reading as the original (P0 under codebook 1.1, the variants under 1.2), the
wrappers with the repeat of the population wrapper as the original (all under codebook 1.2), or the population
reading and its repeat (codebook 1.2 against 1.1). The within-text share is the weighted mean over texts of the
across-reading variance divided by the weighted total variance, with intervals from 500 within-stratum bootstrap
resamples; ``mean R'' is the mean research share over the readings compared, under the same weighting.}\label{tab:decomposition}
\input{tables/t_decomposition}
\end{table}

\begin{table}[tbp]\centering\small
\caption{Sensitivity of the readings on the human sample (design-weighted). ``Same R-or-not'' compares only the
research decision. ``Original'' is the population reading, made under codebook 1.1; the variants, the no-context
reading and the repeat were made under 1.2, and Appendix Table~\ref{tab:invariance_within} repeats the
comparisons with the repeat as the baseline.}\label{tab:invariance}
\input{tables/t_invariance}
\end{table}

\paragraph{Context.} Removing the funder, recipient, year and amount and showing the purpose text alone leaves
Flash's status unchanged on \csNoCtxFlashSameTexts{} of texts and \csNoCtxFlashSameDollars{} of dollars
(research-or-not unchanged on \csNoCtxFlashSameRDollars{} of dollars), and mini's on \csNoCtxMiniSameTexts{} of
texts; agreement with the human reference, who saw the context, is \csNoCtxFlashHuman{} without it against
\csOrigFlashHuman{} with it. The codebook asks the reader to judge the text alone, and the readings mostly do.

\paragraph{Contrasting contexts for the most-shared texts.} Of the 100 most-shared texts in the population, the
\csContrastFlashN{} that also appear under a different funder, covering \csContrastFlashRows{} grant rows and
\$\csContrastFlashB{}B, were read twice under the original wrapper: with the context of the largest grant carrying
the text (what the population run saw) and with the context of a randomly chosen grant of a different funder
carrying the same text. Flash gives the same status in \csContrastFlashSame{} of pairs (same research decision in
\csContrastFlashSameR{}), mini in \csContrastMiniSame{}. This tests robustness to a selected substitution of
context that the codebook deems irrelevant, on the texts whose label propagates furthest; it does not test whether
a single text label recovers each grant's purpose, which the codebook does not ask of it.

\paragraph{Repeat runs and codebook versions.} A second run of the population wrapper (P0) on the sample, with
the model settings of the population run and codebook version~1.2 in place of 1.1, returns the same status as the
first on \csRepeatSonnetSameTexts{} of texts and
\csRepeatSonnetSameDollars{} of dollars for Sonnet, \csRepeatFlashSameTexts{} and \csRepeatFlashSameDollars{} for
Flash, and \csRepeatMiniSameTexts{} and \csRepeatMiniSameDollars{} for mini (the research decision is unchanged on
\csRepeatSonnetSameRDollars, \csRepeatFlashSameRDollars{} and \csRepeatMiniSameRDollars{} of dollars), and agrees
with the human reference as often as the first (\csRepeatSonnetHumanDollars, \csRepeatFlashHumanDollars{} and
\csRepeatMiniHumanDollars{} of dollars against \csOrigSonnetHumanDollars, \csOrigFlashHumanDollars{} and
\csOrigMiniHumanDollars). The exception is GPT-5 mini, whose repeat agrees with the reference on
\csRepeatMiniHuman{} of texts against \csOrigMiniHuman{} for the population run of 4~October, with the same
status on only \csRepeatMiniSameTexts{} of texts; by dollars the two runs agree. The comparison therefore
measures a combined perturbation of run and codebook version. Version~1.2 leaves the status rules as they were
and changes only when a purpose type is required (on research texts only, where version~1.1 also required one on
unclear texts), but an instruction about one output field can move another, and its effect on the status a
model returns is not separately identified here. Nor can a single repeat distinguish a change in the deployed
model between the two dates from ordinary generation variability, and we do not attribute the difference. The within-text share of variance between the two runs
(Table~\ref{tab:decomposition}, ``repeat run'') is \csWithinRepFlashTexts{} by texts and
\csWithinRepFlashDollars{} by dollars for Flash (\csWithinRepMiniTexts{} and \csWithinRepMiniDollars{} for mini,
\csWithinRepSonnetTexts{} and \csWithinRepSonnetDollars{} for Sonnet), against \csWithinFlashTexts{} and
\csWithinFlashDollars{} across wrappers (\csWithinMiniDollars{} and \csWithinSonnetDollars{} by dollars for mini
and Sonnet). The two decompositions rest on different numbers of readings and are not additive components of one
model, so their difference is not a bound on a wording effect; what they show is that the repeat of a single
wrapper under the later codebook version and the instruction variants both move the readings, by amounts of the same order, which is the
sensitivity an office should expect from this instrument. From the committed dry runs on 300 texts, made before
the panel models were chosen and with earlier models (claude-fable-5-1, gpt-5, gemini-2.5-pro), the change from
codebook version~1.0 to 1.1, which added three rules, changed
\csDryVersionChangedMin{}--\csDryVersionChangedMax{} of statuses (appendix Table~\ref{tab:dryruns}, with those
models' repeat runs); these are historical diagnostics of other models.

\begin{figure}[tbp]\centering
\includegraphics[width=0.7\textwidth]{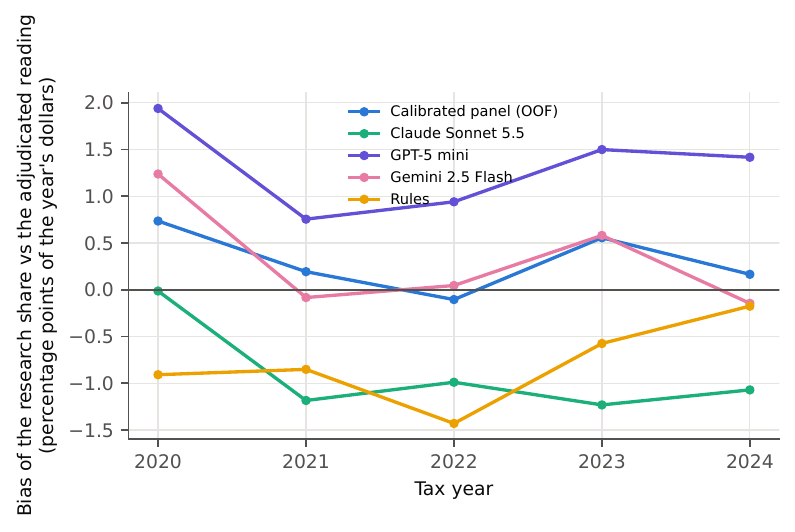}
\caption{Bias by tax year of each reading against the reference, in percentage points of the year's dollars.}\label{fig:drift}
\end{figure}

\paragraph{Composition across tax years.} An uncorrected comparison of machine-labelled totals across years is
distorted when the dollar-weighted labelling error moves with the year; the year-specific rectifiers of
Section~\ref{sec:ppi} estimate that error separately for each year, and the growth figures of that section use
them. This paragraph measures how large the error they correct is. On the sample, weighting each text by its dollars in each year, the rule-based reader's research share
differs from the reference by \csDriftRulesTwenty{} points of 2020 dollars and \csDriftRulesTwentyFour{} points of
2024 dollars; the calibrated panel's bias ranges from \csDriftCalMin{} to \csDriftCalMax{} points across the five
years (appendix Table~\ref{tab:drift}, Figure~\ref{fig:drift}; \csRerunDriftCalMin{} to \csRerunDriftCalMax{}
points with the panel refitted to each reader and to their majority, Appendix
Table~\ref{tab:reference_rerun_sections}). This re-weights one set of labels by each year's dollars, so it
measures how the composition of the yearly estimand moves the error of a fixed instrument, not drift of a
deployed model over time. What
this table adds to the year-level rectifiers is the size of what they correct: at most \csDriftCalAbsMax{} points in any year for the calibrated panel and up to
\csDriftSingleAbsMax{} points for a single reader, with a year-to-year movement of up to
\csDriftSingleStepMax{} points that an uncorrected growth figure would inherit.

%% file: sections/downstream.tex
\section{Totals for one institution: a small-area estimator, and a forecast comparison}\label{sec:downstream}

What a research office or a statistical agency asks for is rarely the national total: it is one institution's
research-purpose dollars, with an interval. That is a small-area problem. The domain total of institution $g$ is
$T_g = \sum_i D_{ig}\,\ind{y_i = R}$, where $D_{ig}$ is the part of text $i$'s domain dollars carried by rows at
institution $g$ (a text's dollars are split across its institutions in proportion to the rows' amounts, so
$\sum_g D_{ig} = d_i$). The sample was drawn for the national total: \csSaShareWithSample{} of the \csSaNInst{}
institutions have at least one sampled text carrying their dollars (a median of \csSaMedianSampled{} such texts),
but those are mostly the shared generic texts, and the dollars of a small institution sit on texts no one read.
Four estimators are natural. The rules plug-in $P_g = \sum_i D_{ig}\ind{\text{rules}_i = R}$ and the calibrated
plug-in $F_g = \sum_i D_{ig} f_i$ are synthetic estimators, exact only if the predictor is unbiased within the
domain. The direct dw-PPI estimator adds the Horvitz--Thompson rectifier of the domain,
$\hat T_g = F_g + \hat R_g$ with $\hat R_g = \sum_{i \in S} w_i D_{ig}(\ind{y_i = R} - f_i)$, which is
design-unbiased for a fixed $f$ (Proposition~\ref{prop:ppi}) with design variance $V_g$ estimated within strata
by the with-replacement approximation; it is useless where the domain has few sampled texts. The composite
$\tilde T_g = F_g + \gamma_g \hat R_g$ with $\gamma_g = \hat\kappa A_g / (\hat\kappa A_g + \hat V_g)$ is a
Fay--Herriot-inspired working composite whose synthetic part is the calibrated plug-in and which has no
regression parameters:
$A_g = \sum_i D_{ig}^2 f_i(1 - f_i)$ is the variance the calibrated probabilities themselves imply for the
domain's residual sum, and $\hat\kappa = \max\{0, \sum_g (\hat R_g^2 - \hat V_g) / \sum_g A_g\}$ is estimated
across institutions by moments; $\kappa = 1$ when $f$ is calibrated and the residuals are independent, and larger
when the predictor's bias differs by institution. This is a scalar working composite, not a full area-level
model: institutions are not independent areas, because one text's residual enters every institution that carries
it, so the residuals of institutions $g$ and $h$ have covariance $\sum_i D_{ig} D_{ih} v_i$ with $v_i$ the text's
residual variance, plus the design's own covariance, and the scalar $\hat\kappa$ absorbs calibration error,
this dependence and error in $\hat V_g$ as well as institution-specific disagreement. The standard error we
report for the composite is $\sqrt{\gamma_g \hat V_g}$, the square root of the working model's leading
mean-squared-error term: under a correctly specified scalar model with known $\kappa$ that term already carries
the modelled uncertainty of the synthetic part, but it ignores the estimation of $\kappa$, the cross-institution covariance and any misspecification of
the scalar model or of the fitted predictor, and we do not add the corrections for estimated parameters of
\textcite{prasadrao1990}; it is a working-model quantity, anti-conservative in the simulation below. When
$\hat\kappa = 0$ the composite is the plug-in and this term is zero, a degenerate working-model variance rather
than a statement that the plug-in has no error; the tables print a dash in that case. A contrast between two
institutions would also need their shared-text covariance, which we do not report.

\paragraph{The real sample.} Table~\ref{tab:smallarea_inst} gives the largest institutions. For
\csSaInstOneName{} (\$\csSaInstOneDollarsB{}B of domain dollars, \csSaInstOneN{} sampled texts) the rules plug-in
is \$\csSaInstOneRulesB{}B, the calibrated plug-in \$\csSaInstOneCalB{}B, the direct estimate
\$\csSaInstOneDirectB{}B (s.e.\ \csSaInstOneDirectSeB) and the composite \$\csSaInstOneCompB{}B
(s.e.\ \csSaInstOneCompSeB, $\gamma = \csSaInstOneGamma$); for \csSaInstTwoName{} the direct estimate moves the
plug-in the other way, from \$\csSaInstTwoCalB{}B to \$\csSaInstTwoDirectB{}B. Across institutions
$\hat\kappa = \csSaKappaReal$, below the value of one that calibrated probabilities with independent residuals
would give: the dispersion of the domain rectifiers beyond their design variance is smaller than the calibrated
probabilities themselves imply, so the composite puts little weight on the rectifier ($\gamma_g$ near
\csSaGammaTopTwenty, median over the top twenty) and stays close to the calibrated plug-in. The scalar estimate
does not separate institution-specific disagreement between the reference and the calibrated panel from the
shared-text covariance, calibration error and error in $\hat V_g$. Among the twenty largest institutions the
direct estimate differs from the calibrated plug-in by more than two standard errors for
\csSaTopTwentyDirectMoves{} of them.
With the calibrator refitted to each reader or to their majority (Appendix
Table~\ref{tab:reference_rerun_sections}) the direct estimate for \csSaInstOneName{} runs from
\$\csRerunSaTopDirectMinB{}B to \$\csRerunSaTopDirectMaxB{}B against a calibrated plug-in of
\$\csRerunSaTopCalMinB{}--\csRerunSaTopCalMaxB{}B, and the direct estimate moves
\csRerunSaMovesMin{}--\csRerunSaMovesMax{} of the twenty largest institutions by more than two standard errors;
$\hat\kappa$ is \csRerunOneSaKappa{} under reader~1, \csRerunTwoSaKappa{} under reader~2 and
\csRerunMajSaKappa{} under the majority, and is truncated at zero under reader~3, a boundary estimate: the moment
estimator found no positive dispersion, which is not evidence that institution-level disagreement is absent, and
the composite returns the calibrated plug-in there. The dispersion $\hat\kappa$ measures is largest under
reader~1, the reader whose research line the calibrator predicts worst, so the size of the composite's
correction, like the total, follows the reading.

\begin{table}[tbp]\centering\small
\caption{Research-purpose dollars (\$B) of the fifteen largest institutions by domain dollars: the two plug-ins,
the direct dw-PPI estimate with its design standard error, and the composite with its working-model standard
error $\sqrt{\gamma_g \hat V_g}$ (the root of the leading mean-squared-error term, which omits the estimation of
$\kappa$ and the shared-text covariance); sampled texts are those in the human sample that carry any of the institution's
dollars.}\label{tab:smallarea_inst}
\input{tables/t_smallarea_inst}
\end{table}

\paragraph{Which estimator to trust, by size.} Table~\ref{tab:smallarea_sim} re-draws the committed design
\csSaDraws{} times under the three synthetic truths of Section~\ref{sec:ppi} with the real-sample calibrator held
fixed, and scores every institution's four estimates against its truth, by the institution's research dollars
under that truth (no institution reaches \$1B). Three findings. First, for institutions with \$100M or more
(about \csSaAHundredPps{} PPS-drawn texts carry their dollars in a draw) the direct interval covers at
\csSaAHundredDirectCov{} under truth~A and the composite's at \csSaAHundredCompCov; under truths B and C the
direct interval covers at \csSaBHundredDirectCov{} and \csSaCHundredDirectCov{} and the composite's at
\csSaBHundredCompCov{} and \csSaCHundredCompCov, so even the largest class has no interval that is reliable
across the truths. The direct estimate's median relative RMSE under truth~A is
\csSaAHundredDirectRel{} against \csSaAHundredCompRel{} for the composite and \csSaAHundredCalRel{} for the
calibrated plug-in: the composite has a lower RMSE than the direct estimate for \csSaAHundredCompBeatsDirect{}
of these institutions and than the calibrated plug-in for \csSaAHundredCompBeatsCal. Second, when the predictor
is biased (truth~B, where the truth is one machine reader's labels, Claude Sonnet~5.5's, and the calibrated plug-in is off by
\csSaBHundredCalRel{} in the median), the composite corrects part of the bias (\csSaBHundredCompRel) and the
direct estimate all of it at the price of variance (\csSaBHundredDirectRel); the rules plug-in looks best under
truth~B (\csSaBHundredRulesRel) only because the rules reader and the reader whose labels define truth~B agree on
most dollars, which is a statement about that truth, not about the rules. Third, below \$10M no interval is
reliable: in the \$1M--\$10M class under truth~A the direct interval covers at \csSaAOneDirectCov{}
(\csSaAOneDirectCovFive{} among the draws with at least five PPS texts in the domain) and the composite's
working-model interval at \csSaAOneCompCov, and under \$1M at \csSaASmallDirectCov{} and \csSaASmallCompCov,
because the correction rests on one or two texts and its error is far from normal. The supported statement about
the point estimates is narrower than ``best everywhere'': the composite improves on the direct estimate in every
class and truth, and on the calibrated plug-in in some classes but not uniformly; under truth~B in the
\$1M--\$10M class its median relative RMSE is \csSaBOneCompRel{} against the plug-in's \csSaBOneCalRel, under
truth~A, where the plug-in is exactly calibrated, it is no lower than the plug-in's in any class
(\csSaAOneCompRel{} against \csSaAOneCalRel{} in the \$1M--\$10M class), and in the smallest class
under truth~A both are large (\csSaASmallCompRel{} against \csSaASmallCalRel), so the weakest-measured
institutions stay poorly measured. Its intervals under-cover in every class, and the analysis is exploratory.
Relative RMSE is undefined for institutions with no research dollars under a truth, which the medians omit:
\csSaAZeroN{} institutions under truth~A, \csSaBZeroN{} under~B and \csSaCZeroN{} under~C, all in the smallest
class, whose median absolute RMSE is below \$0.1M for both the calibrated plug-in and the composite under every
truth (the released table gives the values). The design lesson is the one of Sections~\ref{sec:selective} and~\ref{sec:acquisition}: an
office that must report institution totals with intervals needs a validation round stratified on institutions, or
a model-based mean-squared-error estimate (a bootstrap over the design, with the shared texts' covariance) that
this paper does not supply.

\begin{table}[tbp]\centering\small
\caption{Small-area simulation: \csSaDraws{} re-draws of the committed design, calibrator fixed, by the
institution's research dollars under each truth. Relative RMSE is the median over institutions of
$\mathrm{RMSE}_g / T_g$ for the rules plug-in, the calibrated plug-in, the direct dw-PPI estimate and the
composite; coverage of the nominal 95\% intervals over all draws and, after the slash, over the draws in which at
least five PPS-drawn texts carry the institution's dollars; $\gamma_g$ is the mean shrinkage weight.}\label{tab:smallarea_sim}
\input{tables/t_smallarea_sim}
\end{table}

\paragraph{A forecast comparison that is descriptive only.} A calibrated probability on every grant row also changes
the graph that decision tools are built on: an edge weight becomes expected research dollars rather than a rule's
verdict. We asked whether that matters in a task research offices ask for, a ranked list of funders likely to
start supporting an institution's research, and it does not, at this resolution. The evaluation is a tax-year
split: features from tax years 2020--2021 and outcomes in 2022 to train, features from 2020--2022 and outcomes in
2023--2024 to test, over every pair of a funder and an institution active in research giving in the feature
window and with no research grant between them in it (\csLpRulesNTest{} candidate pairs, \csLpRulesPosTest{} of
which form a new research tie; prevalence \csLpRulesPrev). A tax-year split is not an information-availability
split (a 2022 filing may be published during the outcome window), so this is a retrospective temporal association
benchmark, not a strict forecast. The outcome is defined twice, by the rules layer and by the calibrated label,
and the feature set four times: structural and context features only, plus rules-based research dollars, plus
calibrated expected research dollars, plus both, each with an Adamic--Adar score \parencite{adamicadar2003}; the
learner is gradient-boosted trees; baselines are popularity, locality, a prior any-purpose
tie, and Adamic--Adar alone, the heuristic \textcite{shekhtman2024} used on a static split. Appendix
Table~\ref{tab:linkpred} gives the results. Structure alone predicts a new tie (AUC \csLpRulesCommonAUC, average
precision \csLpRulesCommonAP, precision at 10 of \csLpRulesCommonPat{} against a prevalence of \csLpRulesPrev);
adding the label features raises average precision to \csLpRulesBothAP{} on the rules-labelled outcome and
\csLpCalBothAP{} on the calibrated one, and replacing rules-based with calibrated label features moves it by less
than 0.01 in either direction depending on which outcome is scored. We report no paired uncertainty for these
differences, so the comparison is descriptive. The AUC at this prevalence is dominated by easy negatives and the
outcome is itself a machine label; the descriptive comparison did not establish a benefit of the calibrated
labels for this forecast, and it does not establish their equivalence with the rules-based labels. Where label
quality matters is the institution total above, where plug-in bias does not
average out.

%% file: sections/discussion.tex
\section{Discussion and limitations}\label{sec:discussion}

\paragraph{What the methods assume, and what each claim is.} Every validity claim rests on
Assumption~\ref{as:ign} and on the reference labels. Proved: dw-PPI with a fixed $\lambda$ is exactly unbiased
under the design for a predictor fixed or trained independently of the sample, and for the frame-fold
cross-fitted predictor under Poisson sampling; the variance formula holds for fixed $f$ and $\lambda$; the
conformal p-values are super-uniform under an idealised draw model (independent calibration draws from the size
measure, an independent test unit from the target measure, a common score fixed with training information
separate from both). Approximate, and
checked by simulation: cross-fitting under the committed fixed-size design, the Hájek form, the by-year and growth
ratios, the tuned $\lambda^*$, and the Rao--Wu interval, which the simulation shows to be slightly conservative
under truths A and C and at nominal under truth B (below it, at \csSimOneCovBRefit, in the configuration reported
in Section~\ref{sec:ppi}); the deployed conformal rule, which pools strata, samples
without replacement, excludes the sampled texts from the target and uses cross-fitted scores. Evaluated only: the
weighted BH step's error control, which is checked and not proved, and the Learn-then-Test rule, whose p-values
are exact for equal-weight independent draws and rest on an effective-sample-size substitution that has no
finite-sample justification under unequal weights, so that its violation probability is measured on re-draws
(Appendix~\ref{app:stress}) rather than bounded. The active design's optimality is for the variance of a
Horvitz--Thompson rectifier under Poisson sampling with a calibrated predictor; the replays keep that
Horvitz--Thompson form (no replay uses a ratio estimator) and measure what fixed-size draws, miscalibration and
inclusion probabilities near zero do to it, and the sequential
estimator of the designs with every label charged is conditionally unbiased wave by wave with intervals whose coverage the
replay reports. None of these claims is about a different codebook: a reader who labels under other rules
measures a different quantity.

\paragraph{The reference is a reading, and readings differ.} Three readers labelled the whole sample under the
frozen codebook, blind to every machine label and to one another, and then resolved together the \csNSplit{}
texts on which they had split on research or not; the adjudicated reading is the reference of this paper
(Section~\ref{sec:testbed}). The first labels are kept and released as given, and Tables~\ref{tab:readers},
\ref{tab:altref}, \ref{tab:reference_rerun} and~\ref{tab:reference_rerun_sections} say what the choice of
reading moves. The readers give the same answer on research-or-not for \csUnanBinaryTexts{} of texts and
\csUnanBinaryWd{} of dollars, and the \csNSplit{} texts they split on move the total by
\$\csAltReadersSpreadPpiB{}B between readers~2 and~3, \csAltSpreadOverWidth{} times the sampling interval under
the reference; reader~1 sits between them, the majority of the three within \$\csMajVsOneB{}B of reader~1, and
the adjudicated total between the majority's and reader~3's. The methods' conclusions hold under every reading:
the rectified total moves with the labels and not with the calibrator, the regime (little gain from the predictor
under this design) is the same, and the active review design beats dollar sampling under each. What changes with
the reading is the substantive number, the growth test, the yield of the rules that accept machine labels at a
stated dollar error, the size of the active design's gain, and the institution-level dispersion the composite
estimator shrinks toward ($\hat\kappa$ is \csSaKappaReal{} under the adjudicated reading, \csRerunOneSaKappa,
\csRerunTwoSaKappa{} and \csRerunMajSaKappa{} under reader~1, reader~2 and the majority, and zero under
reader~3).

\paragraph{What the change of reference moved.} The first version of this paper took reader~1's first pass as the
reference, the only reading with a blind re-read, and named an adjudication of the split texts as the next step.
The adjudication chose research on \csAdjStatusR{} of the \csAdjTexts{} texts, more often than the majority of
the first labels (\csAdjSplitMajRN); of those \csAdjStatusR, reader~1 had said unclear on \csAdjOneUToR{} and
not research on \csAdjOneNToR, and the adjudication also took \csConOneLoss{} texts that reader~1 had called
research out of it (Table~\ref{tab:reference_contrast} gives both directions for every reading). The total is
\$\csPpiB{}B [\csPpiLoB, \csPpiHiB] where it was \$\csRerunOnePpiB{}B [\csRerunOnePpiLoB, \csRerunOnePpiHiB]
under reader~1 with the pipeline refitted to that reader: \$\csAdjMinusOnePpiB{}B higher, and inside the range
the three readers span. On the labels alone, with the predictor held fixed, the paired contrast is
\$\csConOneB{}B [\csConOneLoB, \csConOneHiB], an interval that excludes zero; against reader~3 the same contrast
does not (\csConThreeSign\$\csConThreeAbsB{}B [\csConThreeLoB, \csConThreeHiB]). Reader~1 sat low among the
readings, and several findings of the first version were findings relative to that reader. All three language
models differ together from the adjudicated reading on \csWtAdjN{} texts where they differed from reader~1 on
\csWtOneN. The calibrator's held-out log-loss is \csRerunAdjLL{} where it was \csRerunOneLL. The gold-free
Dawid--Skene total is \csDsGoldFreeVsRef{} the rectified interval where it was outside. The unsupervised
posterior recalibrated on the sample shows no resolved difference from the supervised calibrator in dollar
log-loss in the descriptive paired comparison of Table~\ref{tab:paired}, where under reader~1 it did (the small
difference in dollar AUC remains), and where all four readings say research the reference labels \csRelVoteTopObsDollars{} of the dollars
research, against \csRelVoteTopObsDollarsOne{} under reader~1. The growth interval excludes zero where it reached
it. None of this is a validation of the machines: the adjudication saw no
machine output, and agreement with a reading is evidence about where that reading draws its line, not about
correctness. What did not move is the method: the rectifier, the accept rules and the review designs behave under
the adjudicated reading as they did under reader~1, with the yields and margins of
Table~\ref{tab:reference_rerun}.

\paragraph{What the adjudication does not settle.} An adjudication replaces a spread by one number without making
that number the truth. It was made by the same three readers with their first labels in view, on the split texts
only and with no reasons recorded: it is not independent of the readers, it cannot be audited text by text, the
\csNStatusOnlySplit{} texts on which the readers differ between not research and unclear were not re-opened, and
a second adjudication by other readers could draw the line elsewhere. No reading is the reference in any absolute
sense: the three readers apply the same codebook with different thresholds for ``unclear'' (\csUShareOne,
\csUShareTwo{} and \csUShareThree{} of texts), and their spread is a property of the codebook as much as of the
readers. The paper therefore reports every reference-dependent result under each reader as well
(Tables~\ref{tab:reference_rerun} and~\ref{tab:reference_rerun_sections}), and offers no one total without the
range beside it. The methods are indifferent to who the reader is: every estimator, rule and policy takes the
reference as given, and the released sample carries its inclusion probabilities. The right next steps are a
second, independent adjudication with reasons recorded and the unclear-against-not-research texts included, a
year-stratified top-up, and the design lesson of Section~\ref{sec:selective}: draw it with the guarantee one
wants in mind.

\paragraph{Dependence among readers.} All three language models differ from the reference on \csWtAdjTexts{} of
the sampled texts (\csWtAdjWd{} of dollars), against \csWtAdjIndepTexts{} if their pooled errors were
independent, and from the three readers and their majority on \csWtTextsMin{}--\csWtTextsMax. This is marginal
dependence of pooled errors relative to one reading, with no features and no conditioning on the label, and part
of what it counts as joint error under one reader is that reader's distance from the others. It does not by itself
refute the assumption that a gold-free estimator \parencite{egamishin2026dmm} makes, independence of the readers'
errors given the latent label and the unit's features: readers that err independently given a text's difficulty
are wrong together far more often than the product of their marginal error rates predicts, and the reference is
not the latent label. What the experiment shows about the simplest member of that family, homogeneous
Dawid--Skene with no human label, is that its total is \csDsGoldFreeVsRef{} the rectified interval under the
adjudicated reference and outside it under reader~1: a gold-free total cannot be judged without a human sample
to hold it against, and agreement among the models is not a certificate of correctness. No validity claim in this
paper uses inter-model agreement as evidence; agreement carries information, which the calibrator uses as a
feature, and the human sample anchors it. The same association may be why no resolved gain of the calibrated
panel over the best single language models with text cues was detected on dollar log-loss; whether a feature-conditional gold-free
estimator would do better on these \csNSample{} texts is untested and is future work.

\paragraph{What is not established.} The totals are counts of reported amounts by stated purpose, not research
expenditure; a grant for ``research and scholarships'' is counted whole, as the codebook prescribes, and a grant
with a generic purpose is counted as unclear whatever the money did. Completeness of the filing record, recipient
resolution and the funder typology are outside this paper and documented in the companion. The growth test
compares reported private grants with federally financed university R\&D expenditure, two series with different
accounting; the interval excludes no growth and admits parity. The replay pools are pseudo-populations built from one
sample, and the protected replay's reported uncertainty is over splits of that sample, not over fresh frames. The
forecast is retrospective and predicts a machine-labelled outcome. The sensitivity study runs each wrapper once
with one repeat of the population wrapper, and that repeat carried codebook version~1.2 where the population
run carried version~1.1, so it cannot attribute the differences it finds to wording, the codebook version,
generation randomness or a change in the deployed model; comparing the variants with that repeat holds the
version fixed (Appendix Table~\ref{tab:invariance_within}) and still rests on one run each. The institution totals of Section~\ref{sec:downstream} are
exploratory: the composite improves on the direct estimate in the simulated regimes and on the calibrated plug-in
in some size classes, its intervals under-cover in every size class (the direct interval comes nearest to
nominal in the largest class, at \csSaHundredDirectCovMin{}--\csSaHundredDirectCovMax{} across the truths), the
shared texts make institutions dependent, and $\hat\kappa$ is a boundary estimate under reader~3; their
simulation holds the calibrator fixed. The Learn-then-Test acceptance rule is
evaluated, not certified, and its stress tests hold the calibrator's scores fixed. The coverage simulation holds
the ridge strength at its real-sample value except in the \csRetuneDraws{}-draw re-tuning check. The held-out
fold ranges of Section~\ref{sec:selective} and the split-level uncertainty of the protected replay are
conditional Monte Carlo summaries on one sample, not inference to a new population. The designs with every label
charged are replayed on the same pseudo-population as the other policies and inherit its limits, and their budget
counts reference labels revealed, not the three readings and the adjudication behind each label: the
\csAdTwoPhaseVsClassicHundredPct{} reduction in RMSE is a comparison of policies at equal query counts against a
fixed reference, not a measured saving in reading time. It is also conditional on the penalty of the policy's
calibrator, a constant taken from a choice made on this sample: the reduction is
\csRidgeTunedHundredPct{} when the policy chooses the penalty on its own first-phase labels
(\csRerunRidgeTunedHundredMin{}--\csRerunRidgeTunedHundredMax{} under the individual readers and their
majority, where the constant does better than the policy's own choice) and falls to
\csRidgeVeryStrongHundredPct{} under the strongest penalty of the grid (Appendix
Table~\ref{tab:adaptive_ridge}). The reader ablations of Section~\ref{sec:aggregation} concern dollar log-loss;
what the panel is worth to the review design, against a calibrator built on one model, was not tested. The unclear status of the reference is the readers'
majority on the \csNStatusOnlySplit{} texts that were not adjudicated, so the three-class results are scored
against a reference that is adjudicated in part.

\paragraph{Implications for practice.} For an office classifying its own records with a language model, the
sequence is: read a designed sample by hand, stratified on the machines' votes if the total is what matters and
uniformly if counts must be controlled; calibrate the model's labels on it, where the panel showed no resolved
advantage in dollar log-loss over the best single models with text cues at the precision of this sample;
estimate totals with the rectifier rather than the plug-in; and spend further reading time by mass times
predicted error with a floor on every text's probability of being read. Accepting machine labels is a separate
task with its own objective, not a step toward the total: the labels the model may keep are accepted at the
error one can defend, in the unit one reports, by the Learn-then-Test rule, treating its bound as an evaluated
target rather than a certificate until an audit sample with a bound derived for its design has been drawn
(freeze the candidate rules, sample the candidate population with known probabilities, and bound the error
dollars of every candidate set simultaneously). If accepted labels are nevertheless counted in a total, their
error has to be audited on a probability sample of the accepted texts and carried in the total's uncertainty; an
interval for the unaccepted part alone is not an interval for the total (Section~\ref{sec:acquisition}). An office with no validation sample at
all should spend a first phase on a dollar-proportional draw, fit the calibrator on it with its penalty chosen on
those labels, and let the design choose
the rest, with the first-phase share chosen for the budget at hand (a quarter was best at 100 and 200 texts here;
Section~\ref{sec:acquisition}); an office that must report institution totals may use the composite as a point
estimate, with the caveats of Section~\ref{sec:downstream}, and should stratify its next validation round on
institutions before it reports intervals. For a statistical agency, the same sequence is a protocol for machine-assisted
coding of free-text fields with stated and checked properties. For the filing form, the disagreement map puts a magnitude on what a controlled purpose vocabulary
could address, about a fifth of the dollars the readers disagree about, and no more than that.

%% file: sections/release.tex
\section{Reproducibility and release}\label{sec:release}

Every number and table in this paper is generated by the analysis code from the text-level frame, through the
calibration, the estimators and their coverage simulation, the accept rules and stress tests, the review replays,
the sensitivity analysis, the forecast, the regime analysis, the institution totals, the designs with every label
charged, the reader and adjudication workbooks and the re-runs under each reading, to this document; no number in the text is typed
by hand. The replication record for this paper \parencite{koomullil2026record} releases what the analyses consume
at the level of the sample and of aggregates: the \csNSample{}-text calibration sample with its strata, inclusion
probabilities, sampling sizes, domain dollars by year, the four machine statuses, the out-of-fold calibrated
probabilities, the three readers' labels, the blind re-read and the adjudicated status of the split texts, the
frame-fold assignment and the chosen ridge
strength of every fold; the population totals by stratum and year that the estimators use (texts, sampling size,
domain dollars, predicted dollars under each machine reading and under the calibrator), with no text-level rows;
the codebook in both versions; the prompt wrappers and the raw answers of the sensitivity runs, which were made on
the sampled texts and, for the contrasting-context runs, on the \csContrastFlashN{} most-shared texts filed under more
than one funder, with the provider model identifiers and timestamps; every table of this paper as a CSV, including the
per-fold results of the nested held-out evaluation, the per-draw results of the stress tests, the accepted mass and
realised error at each $\alpha$ under both rules, and each review policy's estimate, interval, texts read and
accepted mass per replay; and the analysis code. The grant database, the \csNTexts{}-text population with its
machine readings and calibrated probabilities, the pipeline that builds them and the text-level accepted sets are
not released: they are the working data of the author's company, and are available to researchers under
agreement. Two consequences follow. The population accept rules of Section~\ref{sec:selective} score every text
of the population, and the replays of Section~\ref{sec:acquisition} run on the \csNSample{} sampled texts but
need the \csNFeatures{}-feature matrix of those texts (the machine readings, the unsupervised posteriors fitted
on the whole population, and the text cues), which the record does not carry; both can be checked against the
released per-$\alpha$ and per-replay summaries but not re-run from the record, the review-budget result of
Section~\ref{sec:adaptive} included; and the institution totals of Section~\ref{sec:downstream}
can be checked from the released aggregates by institution, not from rows. The record is therefore a partial
replication record, and we do not describe the paper as fully reproducible from public files. What a reader can
recompute from it are the sample-level results (the calibration metrics from the released out-of-fold
probabilities, the rectified totals from the sample and the stratum
aggregates, the agreement among readers, and the sensitivity study from the raw answers); what a reader can only
check against a released summary are the results that need a calibrator to be refitted: the review replays, the
ridge-strength table of Appendix~\ref{app:tables} and the coverage simulation. The companion report
\parencite{koomullil2026companion} documents the database and the codebook and publishes the domain tables.

\paragraph{Data card.} Unit: a sampled purpose text in the sample file; a stratum by year cell in the aggregates
file. Sample fields: text, sampling size $a_i$, domain dollars $d_i$ and $d_{i,y}$, rows covered, four machine
statuses with purpose types and atom shares, stratum, $\pi_i$, selection type (PPS, certainty, take-all), the
status and purpose type of readers~1, 2 and~3, reader~1's re-read status (subset), the adjudicated status of the
\csNSplit{} split texts and the adjudicated reading, the cross-fitting fold and the
out-of-fold calibrated probabilities. Aggregate fields: texts, sampling size, domain dollars, and predicted
dollars under each reading and the calibrator, by stratum and year, and the institution totals under each truth.
Result files: the per-fold results of the nested held-out evaluation with the ridge strength chosen in each
split, the per-draw results of the stress tests, the split-level results of the hybrid replay (accepted mass,
remainder, signed accepted error, texts read) and the per-$\alpha$ accepted mass and realised error under both
rules. Licence: CC BY 4.0 for the released tables, labels and aggregates, MIT for the code; the source texts are
public IRS records. Cost: the sensitivity runs used \csInvCalls{} model calls
(\csInvTokensIn{}M input tokens, most of them served from cache) at about \csInvCostUSD{} US dollars at list
prices; the population runs are the companion's.

\paragraph{Disclosures.} The author is the founder of a company that builds research-funding tools; the
companion's disclosures apply. The readers' first labels were frozen before any model comparison; the three
readers labelled blind to every machine label and to one another, and none of the three readers is identified
in this paper. The adjudication of the split texts, which completes the reference of this version, was made by
the three readers after the first version of this paper was public, from a workbook that showed their first
labels and no machine output; they were shown none of this paper's estimates. The reference of the first
version, reader~1's first pass, is retained as one of the re-runs, so every number of that version can still be
checked. Language models were used to label texts as described and to assist with the analysis and the
prose under the author's direction; the author is responsible for the results. The
counterexamples of Sections~\ref{sec:ppi} and~\ref{sec:selective} were derived independently and are presented as
diagnostics of the constructions they test.

%% file: sections/appendix.tex
\section{Proofs}\label{app:proofs}

\paragraph{Proposition~\ref{prop:ppi}~(i).} Write $\hT^{\mathrm{HT}}_\lambda = \lambda \sum_i d_i f_i + \sum_{i}
(I_i/\pi_i)\, d_i e_i$ with $e_i = \ind{y_i = R} - \lambda f_i$, so $\hT^{\mathrm{HT}}_\lambda - T = \sum_i (I_i/\pi_i - 1)
d_i (\ind{y_i=R} - \lambda f_i)$. Under Assumption~\ref{as:ign}, $\E[I_i \mid x, y] = \pi_i$, so the $y$ term has
zero expectation and the bias is $-\lambda \sum_i d_i\, \E[(I_i/\pi_i - 1) f_i]$. If $f$ is fixed or independent of
$S$ this is zero. Under the frame-fold construction the folds are fixed independently of sample membership and
labels (in the implementation they were generated after the draw by a seeded rule that uses nothing about it)
and $f_i$ is a function of
$x$ and of $\{(I_j, y_j): j \notin \mathrm{fold}(i)\}$, the same function whether or not $i$ is drawn; under
Poisson sampling $I_i$ is independent of $\{I_j: j \ne i\}$ and hence of $f_i$, so $\E[(I_i/\pi_i - 1) f_i] =
\E[f_i]\,\E[I_i/\pi_i - 1] = 0$ and the bias is zero. The argument needs the \emph{rule} that produces $f_i$ to be
the same on the events $I_i = 1$ and $I_i = 0$: the sampled/unsampled switch (a sampled text gets its
out-of-fold model, an unsampled text the fold average) produces $f_i$ by a rule that depends on $I_i$,
and the factorisation fails. Under a fixed-size design $I_i$ and $I_j$ are dependent and the term need not vanish
for any construction. Table~\ref{tab:two_unit} enumerates the two-unit example exactly: $d = (1,1)$, $y = (1, 0)$,
$\pi = (\tfrac12, \tfrac12)$, one unit per fold, the fold model predicting the drawn label of its training fold
and $0$ if that fold is empty. With the frame-fold predictor under one-of-two sampling, drawing unit 1 gives
$f = (0, 1)$ and estimate $1 + 2(1 - 0) = 3$, drawing unit 2 gives $f = (0, 0)$ and estimate $0$, expectation
$1.5$ against $T = 1$; under Poisson sampling the expectation is $1$. With the switch under Poisson sampling,
drawing unit 1 alone gives $f = (0, \tfrac12)$ and estimate $2.5$, both units $f = (0, 1)$ and estimate $1$,
unit 2 alone or neither $0$, expectation $0.875$: every observed unit's prediction excludes its own label, and
the estimator is biased all the same. The fixed predictor is unbiased under both designs. The table also gives
$\lambda = \tfrac12$; the statement is for fixed $\lambda$, and a tuned $\lambda^*$ is checked only by the
simulation of Section~\ref{sec:ppi}.

\paragraph{(ii).} The Hájek form replaces $D_s$ by its unbiased estimate $\hat D_s = \sum_{i \in S_s} w_i d_i$ in
the denominator. Write the stratum's rectifier as $D_s \hat R_s$ with $\hat R_s = \hat N_s / \hat D_s$ and
$\hat N_s = \sum_{i \in S_s} w_i d_i e_i$, both unbiased for their frame totals $N_s$ and $D_s$ when $f$ and
$\lambda$ are fixed. The Taylor expansion of a ratio of unbiased estimators \parencite{sarndal1992} gives
$\E[\hat R_s] - N_s/D_s \approx \big(R_s \Var(\hat D_s) - \mathrm{Cov}(\hat N_s, \hat D_s)\big)/D_s^2$ with $R_s = N_s/D_s$, a
relative quantity: it is $O(n_s^{-1})$ when $\Var(\hat D_s/D_s)$ and $\Var(\hat N_s/D_s)$ are, which is the usual
regime of a sequence of designs and populations in which $n_s \to \infty$ and no single weighted unit dominates
its stratum, so that $\hat D_s/D_s$ concentrates around one. A denominator bounded away from zero is not enough
by itself. The absolute bias of the stratum's
contribution is $D_s$ times that rate. This is an asymptotic statement for a fixed predictor, not a finite-sample
guarantee, and its conditions are not verified for the committed design;
with many zero-domain texts the realised denominator of a small stratum can be far from $D_s$, which is one
reason the simulation of Section~\ref{sec:ppi} is the operative check. The implementation sets a stratum's
rectifier to zero when its realised denominator is zero.

\paragraph{(iii).} For fixed $f$ and fixed $\lambda$ the first term of $\hT^{\mathrm{HT}}_\lambda$ is constant
given $x$, so the variance of $\hT^{\mathrm{HT}}_\lambda$ is the Horvitz--Thompson variance of
$\sum_{i \in S} w_i d_i e_i$, whose expansion in the inclusion and
joint-inclusion probabilities is the displayed one \parencite{horvitzthompson1952}. For $\lambda = 1$, $\E[e_i^2
\mid x_i] = \Var(\ind{y_i = R} \mid x_i) + (P(y_i = R \mid x_i) - f_i)^2$ is minimised by $f_i = P(y_i = R \mid x_i)$,
which minimises every diagonal term; under Poisson sampling the off-diagonal terms vanish and the whole variance is
minimised. Under fixed-size designs the off-diagonal terms are of indeterminate sign and the statement is about the
diagonal term only. The Rao--Wu bootstrap estimates the with-replacement variance, known to err on the wide side for
systematic sampling on a random order \parencite{raowu1988,nces1999bootstrap}, which is why the simulation of
Section~\ref{sec:ppi} is the operative check.

\paragraph{(iv).} For fixed $f$, $\Var(\hT^{\mathrm{HT}}_\lambda) = \Var(A) - 2\lambda\,\mathrm{Cov}(A, B) +
\lambda^2 \Var(B)$ with $A = \sum_S w_i d_i \ind{y_i=R}$ and $B = \sum_S w_i d_i f_i$, a quadratic in $\lambda$ whose
minimum over any interval containing $0$ is at most $\Var(A)$. The bootstrap minimiser $\lambda^*$ estimates the
population minimiser with error, so the realised variance at $\lambda^*$ can exceed $\Var(A)$ in finite samples;
Table~\ref{tab:coverage} reports both.

\paragraph{Proposition~\ref{prop:conformal}.} Let the null calibration units $\{(X_j, s_j)\}_{j \in \mathcal{N}}$ be
independent draws from $P_{\mathrm{cal}}$ restricted to nulls and the test unit $(X_i, s_i)$ a draw from
$P_{\mathrm{test}}$ restricted to nulls, with $\mathrm{d}P_{\mathrm{test}}/\mathrm{d}P_{\mathrm{cal}} \propto w(x)$.
Theorem~2 of \textcite{tibshirani2019} gives, for the weighted conformal p-value with weights $w(X_j)$ on the
calibration units and $w(X_i)$ on the test unit, $P(p_i \le t) \le t$. Under the dollar measure $w(x) = d_x/\pi_x$
up to a constant on the PPS-eligible part of each stratum and $0$ on certainty units, which is \eqref{eq:pdollar};
under the text measure $w(x) = 1/\pi_x$. For the power bound, the numerator of the text-measure p-value is at least
the test unit's own weight, so $p_i \ge (1/\pi_i)/(\sum_{\mathcal{N}} 1/\pi_j + 1/\pi_i)$; a BH-type step at level
$\alpha$ never rejects a p-value above $\alpha$, and $p_i > \alpha$ whenever $1/\pi_i > \alpha/(1-\alpha) \sum_{\mathcal{N}}
1/\pi_j$. The with-replacement model of the PPS draws, the exclusion of the sampled texts from the test population,
and the use of out-of-fold scores on the calibration units are the approximations; the checks of
Section~\ref{sec:selective} and Appendix~\ref{app:stress} measure them.

\paragraph{Optimal review design \eqref{eq:neyman}.} Under Poisson sampling with inclusion probabilities $q_i$ and
expected size $\sum_i q_i = b$, the HT rectifier $\sum_i (J_i/q_i) m_i e_i$ has variance $\sum_i m_i^2 e_i^2 (1 -
q_i)/q_i = \sum_i m_i^2 e_i^2 / q_i - \text{const}$; a Lagrange multiplier gives $q_i \propto m_i |e_i|$, capped at
one and renormalised. With $|e_i|$ unobserved, the design that minimises the expected variance under the
predictive distribution replaces $e_i^2$ by $\E[e_i^2 \mid x_i]$, which equals $f_i(1 - f_i)$ when $f_i = P(y_i = R
\mid x_i)$ and $p_i(1-p_i) + (p_i - f_i)^2$ otherwise. When $f_i$ is not calibrated the replacement under-states the
residual of over-confident texts, and a mixture bounds $1/q_i$ \parencite{zrnic2024active}: the 50/50 mixture
with the uniform measure gives every text $q_i \ge b/(2N)$ under the capped proportional allocation used here,
while the mixture with the dollar-PPS measure gives a floor proportional to the text's mass, which can be
arbitrarily small for a small-dollar text. The fixed-size PPS draw used in the replay has the same first-order
inclusion probabilities as the Poisson design, but not a smaller variance in general: with residuals
$e = (\tfrac12, -\tfrac12)$, unit masses and $\pi = \tfrac12$ the Poisson Horvitz--Thompson variance is $\tfrac12$
and the one-of-two variance is $1$. Its joint inclusion probabilities differ from the Poisson ones, the
Poisson-type variance estimator omits the covariance terms, and the replay reports the coverage that results.
Agreement of the routine's empirical inclusion frequencies with $\pi$ (Appendix~\ref{app:stress}) validates the
first-order probabilities only. With a cost $c_i$ per text
and the constraint $\sum_i q_i c_i = B$ in place of $\sum_i q_i = b$, the same Lagrangian gives
$q_i \propto m_i |e_i| / \sqrt{c_i}$, capped at one, which is the cost-adjusted allocation of
Section~\ref{sec:adaptive}. For several questions with masses $m_{iq}$ and non-negative weights $\lambda_q$, the
allocation that minimises $\sum_q \lambda_q V_q(q)$ is $q_i \propto \sqrt{v_i \sum_q \lambda_q m_{iq}^2}$; the
minimax allocation of Section~\ref{sec:adaptive} chooses $\lambda$ by multiplicative weights on the variance
ratios.

\section{Stress tests}\label{app:stress}
Table~\ref{tab:stress} reports the re-draw checks of the accept rules on the committed design under four truths.
These are fixed-score stress tests: the calibrator fitted to the real sample and its scores are held fixed, the
synthetic labels of each truth are drawn once and held fixed across the \csStressDraws{} draws, and only the
sample is re-drawn, so the tests check the acceptance step on this design and complement the nested held-out
evaluation of Section~\ref{sec:selective}, which refits the calibrator; they do not reproduce the whole learned
pipeline. Under scenario~A the truth is drawn from the calibrated probabilities and the machine label is the
arg-max, so the null set is random and the scores are the real ones; under the all-null scenario every machine
label is wrong; under scenario~D every label is wrong with probability 0.06 independently of its confidence, so
the expected dollar error of every candidate set is about 0.06, just above $\alpha = 0.05$ and below $0.10$,
where a false acceptance is hardest to detect (the realised error of a candidate set varies around that value);
under scenario~E the wrong labels are the largest-dollar texts among those the design can
leave unread (inclusion probability below one; the certainty texts are always read and are never candidates),
taken in order of size until they carry 6\% of those texts' dollars, the confidently wrong large text that unequal
weights make dangerous. In each of \csStressDraws{} re-draws of the design the dollar-measure p-values of the unsampled texts
are computed from the mislabelled PPS draws, weighted BH and Learn-then-Test are applied, and the realised dollar
error of the accepted set is recorded. The table gives, for each rule and $\alpha$, the mean realised error, the
mean dollars accepted, the probability of a non-empty set, the mean error given a non-empty set, the worst draw,
and the violation probability $P(\text{non-empty set and realised error} > \alpha)$ with a one-sided 95\%
Clopper--Pearson upper bound \parencite{clopperpearson1934}, computed per cell under that cell's simulation law
(an empty set is not a violation; for zero events in \csStressDraws{} independent draws the bound is
\csStressLttViolUpMax). The violation probability is the quantity the Learn-then-Test statement
$P(R(t^*) > \alpha) \le \delta$ is about; a mean error below $\alpha$ is not. Across the four truths and five
levels the largest violation probability of the Learn-then-Test rule is \csStressLttViolMax{} (upper bound
\csStressLttViolUpMax) and of the weighted BH step \csStressWbhViolMax{} (\csStressWbhViolUpMax). Under
scenarios D and~E the Learn-then-Test rule accepts nothing at $\alpha \le 0.05$: a 6\% true error, whether spread
over every text or concentrated on the large ones, raises the observed error at every threshold above what the
Hoeffding--Bentkus bound at the effective size can clear. The absence of violations is a property of this design
and these truths, not of the rule, and the all-null and boundary truths, which reject everything, say more about
the rule's conservatism than about its safety when a large set is accepted; the unequal-weight counterexample of
Section~\ref{sec:selective} describes a case in which the same rule violates with probability 0.15. The
super-uniformity of the null p-values under the dollar measure (each null text weighted by its dollars) is
$P(p \le 0.05) = \csStressSUAFive$, $P(p \le 0.10) = \csStressSUATen$ and $P(p \le 0.20) = \csStressSUATwenty$
under scenario~A, and \csStressSUNullFive, \csStressSUNullTen, \csStressSUNullTwenty{} all-null. The equal-dollar
example of Section~\ref{sec:selective}: with the test text entered at $d_i$, $P(p \le 0.01) = \csExUncorrPle$ and
the BH false discovery rate at $\alpha = 0.05$ is \csExUncorrFDR{}; with the likelihood-ratio weight,
\csExCorrPle{} and \csExCorrFDR. Table~\ref{tab:two_unit} enumerates the two-unit cross-fitting example exactly:
the frame-fold
predictor has bias \csTwoUnitFixedBias{} under one-of-two sampling and \csTwoUnitPoissonBias{} under Poisson
sampling, the sampled/unsampled switch \csTwoUnitSwitchFixedBias{} and \csTwoUnitSwitchPoissonBias{}, and the fixed
predictor none under either. The committed PPS routine's empirical inclusion frequencies on a 30-unit frame with
8 draws match their targets to within \csPpsMaxDev{} over 20,000 draws, which validates the first-order
probabilities and nothing about the variance.

\begin{table}[H]\centering\footnotesize
\caption{Fixed-score stress tests on the committed design over \csStressDraws{} re-draws, panel 1: a truth drawn
from the real scores (A) and an all-null truth. Mean realised dollar error with its Monte Carlo standard error,
mean dollars accepted, probability of a non-empty set, mean error given a non-empty set, the worst draw, and the
violation probability $P$(non-empty and error $> \alpha$) with its one-sided 95\% Clopper--Pearson upper bound,
per cell. Panel 2 (truths D and E) is Table~\ref{tab:stress_b}.}\label{tab:stress}
\adjustbox{max width=\textwidth}{\input{tables/t_stress_a}}
\end{table}

\begin{table}[H]\centering\footnotesize
\caption{Fixed-score stress tests, panel 2: 6\% errors independent of confidence (D) and the largest candidate
texts wrong (E, 6\% of the candidate dollars); columns as in Table~\ref{tab:stress}.}\label{tab:stress_b}
\adjustbox{max width=\textwidth}{\input{tables/t_stress_b}}
\end{table}

\begin{table}[H]\centering\small
\caption{The two-unit cross-fitting example enumerated exactly: $d = (1,1)$, $y = (1,0)$, $\pi = (\tfrac12,
\tfrac12)$, one unit per fold, $T = 1$. The frame-fold predictor gives every unit the model trained outside the
unit's assigned fold, whether or not the unit is drawn; the switch gives a drawn unit that same model and an
undrawn unit the average of the two fold models.}\label{tab:two_unit}
\input{tables/t_two_unit}
\end{table}

\begin{table}[H]\centering\small
\caption{Coverage simulation with the ridge strength re-chosen inside every draw by the nested inner
cross-validation (truth~A, \csRetuneDraws{} re-draws, frame-fold cross-fit): the check that fixing the strength at
its real-sample value in Table~\ref{tab:coverage} does not drive the result. The classic estimator is shown on
the same draws so that the common sampling fluctuation of the \csRetuneDraws{} draws can be read off.}\label{tab:coverage_retune}
\adjustbox{max width=\textwidth}{\input{tables/t_coverage_retune}}
\end{table}

\begin{table}[H]\centering\small
\caption{Super-uniformity of held-out null p-values under the dollar measure (research-or-not task): the
likelihood-ratio-weighted share of null fold observations with $p \le t$, and the unweighted share. The fold
observations reuse the same sample texts across the 20 repetitions, so the shares are descriptive.}\label{tab:superuniformity}
\input{tables/t_superuniformity}
\end{table}

\begin{figure}[H]\centering
\includegraphics[width=0.45\textwidth]{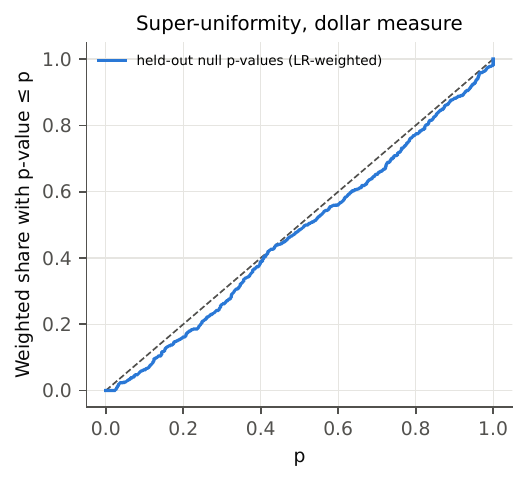}
\caption{Weighted empirical distribution of held-out null p-values under the dollar measure against the uniform.}\label{fig:superuniformity}
\end{figure}

\section{Additional tables}\label{app:tables}

\begin{table}[H]\centering\small
\caption{Intervals for the metrics of Table~\ref{tab:calibration} that it prints without one: ECE by texts and by
dollars and the text AUC, with 95\% intervals from the same \csBBoot{} within-stratum bootstrap resamples. For ECE
the decile cuts of the full sample are held fixed across resamples. ECE recomputed on a resample is biased upward,
so a percentile interval can lie above the value it accompanies.}\label{tab:calibration_intervals}
\input{tables/t_calibration_intervals}
\end{table}

\begin{table}[H]\centering\small
\caption{Three-class metrics (research / not research / unclear), design-weighted. The unclear status of the
reference is adjudicated on the \csNSplit{} texts the readers split on and is their majority status on the
\csNStatusOnlySplit{} texts where they differ between not research and unclear, which were not re-opened; a low
AUC for U therefore reflects that boundary of the reference as well as the models.}\label{tab:threeclass}
\input{tables/t_threeclass}
\end{table}

\begin{table}[H]\centering\small
\caption{Calibrated probabilities by vote-pattern stratum over the whole population. The last column is the
unweighted mean of $\hat p(R)$ over the stratum's texts; the expected research dollars weight each text by its
dollars, so the two need not agree.}\label{tab:stratum}
\adjustbox{max width=\textwidth}{\input{tables/t_stratum}}
\end{table}

\begin{table}[H]\centering\small
\caption{Rogan--Gladen correction of the classic total (\$B) for a reference with sensitivity $Se$ and specificity
$Sp$ relative to the codebook's intended research-or-not label.}\label{tab:rogan}
\input{tables/t_rogan}
\end{table}

\begin{table}[H]\centering\small
\caption{Selective labelling, three-class task (held-out error and population share accepted). A dash means that
no held-out fold accepted anything, so the error is undefined; where some folds accept nothing, such a fold counts
as zero error in the mean and in the range, which makes the figure an unconditional loss, and the column of
empty folds says how much of it that convention supplies; the next column is the mean error over the folds that
accept something, the error to expect given that the rule accepts at all; for the bootstrap-$p$ rule at
$\alpha = 0.01$ and $0.02$ that error is above $\alpha$ although the all-fold mean is below it. The population columns apply the rule
fitted on the whole sample, a different fitted rule from the fold-level ones, so its accepted share can be zero
where theirs is not. The three-way
reference is adjudicated only in part (Table~\ref{tab:threeclass}).}\label{tab:selective_three}
\input{tables/t_selective_three}
\end{table}

\begin{table}[H]\centering\small
\caption{In-pool replay with the cross-fitted calibrator (conditional diagnostic): RMSE (\$B) by policy and budget
(\csAcqReps{} replays).}\label{tab:acqcal}
\adjustbox{max width=\textwidth}{\input{tables/t_acquisition_cal}}
\end{table}

\begin{table}[H]\centering\small
\caption{In-pool replay, calibrated predictor: coverage of the 95\% interval, share of texts at the cap $q_i = 1$,
largest weight $m_i/q_i$ relative to $T^*$, and Kish effective sample size of the weights, by policy and budget.}\label{tab:acqdiag}
\adjustbox{max width=\textwidth}{\input{tables/t_acquisition_diag}}
\end{table}

\begin{table}[H]\centering\small
\caption{Greedy orders with the plug-in estimator, in-pool replay (not comparable with the design-based policies:
no inclusion probabilities, and the plug-in's bias is never estimated). Top: calibrated predictor; bottom: DS
posterior. Absolute error in \$B.}\label{tab:acqgreedy}
\adjustbox{max width=\textwidth}{\input{tables/t_acquisition_greedy_cal}}\\[4pt]
\adjustbox{max width=\textwidth}{\input{tables/t_acquisition_greedy_ds}}
\end{table}

\begin{table}[H]\centering\small
\caption{Cost-proxy experiment with the in-pool calibrated predictor (a conditional diagnostic, not a design with
every label charged): the designs at equal expected words read (the cost of a text is its word count; mean \csAdWordsMean{}
words, median \csAdWordsMedian), reference-label queries (texts read), words read and RMSE of the research total in the 651-text pool,
\csAdReps{} replays.}\label{tab:adaptive_cost}
\input{tables/t_adaptive_cost}
\end{table}

\begin{table}[H]\centering\small
\caption{Designs with every label charged, detail at $b = 100$ and $200$: RMSE of the research total (\$B) with its
Monte Carlo standard error from a bootstrap over the \csAdReps{} replays, bias (\$B), mean width of the 95\%
interval (\$B) and its coverage.}\label{tab:adaptive_detail}
\adjustbox{max width=\textwidth}{\input{tables/t_adaptive_detail}}
\end{table}

\begin{table}[H]\centering\small
\caption{The ridge strength of the replay's calibrator. The quarter-split two-phase design of
Table~\ref{tab:adaptive} (mixed measure in the second phase) at every strength of the grid the main calibrator is
tuned over, the same design with the strength chosen by cross-validation on its own first-phase labels (the
strongest penalty of the grid when fewer than two labels of a class were bought), and the classic design: RMSE of
the research total (\$B) with the coverage of the 95\% interval in parentheses, \csRidgeReps{} replays per cell on
random streams independent of Table~\ref{tab:adaptive}, and the reduction in RMSE against the classic design of
this table. $C$ is the inverse of the penalty, so a small $C$ is a strong penalty. The main replay fixes
$C = \csRidgeCMain$, the value the nested cross-validation chose on the full sample when reader~1 was the
reference; no setting was dropped from the table.}\label{tab:adaptive_ridge}
\input{tables/t_adaptive_ridge}
\end{table}

\begin{table}[H]\centering\small
\caption{Allocation for several questions (the national total and the five yearly totals) with the in-pool
calibrated predictor: RMSE and relative RMSE of each question's dw-PPI estimate under the total-only Neyman
allocation and under the allocation that minimises the largest variance ratio across questions; the variance
ratio is each question's anticipated variance relative to its own optimal allocation.}\label{tab:adaptive_multi}
\input{tables/t_adaptive_multi}
\end{table}

\begin{table}[H]\centering\small
\caption{Run-to-run and codebook-version stability from the committed dry runs (300 texts; earlier models, before
the panel was chosen).}\label{tab:dryruns}
\input{tables/t_dryruns}
\end{table}

\begin{table}[H]\centering\small
\caption{The sensitivity comparisons of Table~\ref{tab:invariance} inside one codebook version (design-weighted).
P0 is the repeat of the population wrapper under codebook 1.2, the version every variant and the no-context
reading inserted, with the model settings of the population run; the comparisons are on the texts every reading
of the model parsed. ``Texts into / out of research'' counts the sampled texts a variant labels research where P0
does not, and the reverse. One run per variant, so generation variability is inside every
comparison.}\label{tab:invariance_within}
\input{tables/t_invariance_within}
\end{table}

\begin{table}[H]\centering\small
\caption{Bias of each reading's research share against the reference, by tax year, in percentage points of the year's
dollars (design-weighted).}\label{tab:drift}
\input{tables/t_drift}
\end{table}

\begin{table}[H]\centering\small
\caption{Retrospective tie forecast, 2023--2024. Precision and recall at 10 are per institution over
\csLpInsts{} institutions; ECE uses bins cut at unweighted deciles of the predicted probability.}\label{tab:linkpred}
\adjustbox{max width=\textwidth}{\input{tables/t_linkpred}}
\end{table}

\paragraph{The second reading round.} Table~\ref{tab:readers_stratum} gives the readers' agreement on
research-or-not by sampling stratum and Table~\ref{tab:readers_machines} each machine reader's agreement with each
human reader. Reader~2's agreement with reader~1 was higher on the \csOccRestTwoN{} texts of the second occasion
(same status on \csOccRestTwoStatus, $\kappa = \csOccRestTwoKappa$) than on the first \csOccFirstTwoN{}
(\csOccFirstTwoStatus, $\kappa = \csOccFirstTwoKappa$), while reader~3's agreement with reader~1 is the same on
both subsets (\csOccFirstThreeStatus{} and \csOccRestThreeStatus); reader~2 said ``unclear'' less often on the
second occasion (\csOccRestTwoU{} against \csOccFirstTwoU). The \csNSplit{} texts on which the readers split on
research-or-not carry \csSplitDollars{} of the domain's design-weighted dollars; the largest are bare field names
(``medical research'', ``HIV'') and named centres or programmes. Each machine reader agrees with the adjudicated
reading on more texts than with any single reader (\csMachSonnetAdjT{} for Sonnet and \csMachMajorityAdjT{} for
the LLM majority, against \csMachSonnetOneT{} and \csMachMajorityOneT{} with reader~1); the adjudication saw no
machine output, and this is a statement about where it drew the line, not a validation of the machines.

\begin{table}[tbp]\centering\small
\caption{Share of texts on which pairs of readers, and all three, give the same answer on research or not, and on
which all three give the same status, by sampling stratum.}\label{tab:readers_stratum}
\input{tables/t_readers_stratum}
\end{table}

\begin{table}[tbp]\centering\small
\caption{Agreement of each machine reader, the LLM majority and the calibrated model (fitted to the reference,
thresholded at one half) with each human reader, the majority and the adjudicated reading on research or not, by
text and by design-weighted dollars, and on the texts where the three readers are unanimous. (The design-weighted text measure is dominated by two texts
of very large weight and is not shown.)}\label{tab:readers_machines}
\input{tables/t_readers_machines}
\end{table}

\begin{table}[tbp]\centering\small
\caption{The three language models differing together from a reading on research or not, under each reading of
the \csNSample{} sampled texts: the number of texts on which all three differ from it, their share of the sampled
texts and of design-weighted dollars; the design-weighted share of texts, the part of its mass carried by the two
heaviest such texts, and the share without those two; and the texts on which any of the three differs, with the
share of them on which all three do.}\label{tab:wrong_together}
\input{tables/t_wrong_together}
\end{table}

\paragraph{The later sections under each reader.} Table~\ref{tab:reference_rerun_sections} completes
Table~\ref{tab:reference_rerun}: the sensitivity study of Section~\ref{sec:invariance}, the institution totals of
Section~\ref{sec:downstream} and the designs with every label charged of Section~\ref{sec:adaptive} re-run with
each reader's labels as the reference and the calibrator refitted to that reader; the first column is the main
run under the adjudicated reading. The model answers of the
sensitivity runs are the same in every column; only the reference they are scored against changes. The
institution rows are for the
largest institution by domain dollars and for the twenty largest; the design rows are the RMSE ratios of
Table~\ref{tab:adaptive} under each reference, with the free-pilot reference policy recomputed under that reader.
Under reader~3 the moment estimate $\hat\kappa$ is truncated at zero, a boundary estimate, so $\gamma_g = 0$ and
the composite equals the calibrated plug-in; its working-model standard error is then the degenerate zero of a
plug-in, printed as a dash, not a measure of that plug-in's prediction error. The forecast, the regime analysis,
the stress tests and
the coverage simulation are not repeated under the readers: they use synthetic truths built from the calibrator
fitted to the reference, or machine labels, and are held fixed rather than being independent of the reference;
repeating them under each reader is optional sensitivity work.

\begin{table}[H]\centering\small
\caption{The sensitivity study, the institution totals and the designs with every label charged re-run under each
reference labelling (calibrator refitted to that labelling; the first column is the main run under the adjudicated
reading). Agreement is design-weighted by dollars or by texts on
the \csNSample-text sample; biases are in percentage points of the year's dollars; institution totals in \$B with
the direct estimate's design standard error and the composite's working-model standard error in parentheses
(a dash where $\hat\kappa$ is truncated at zero); design rows are RMSE ratios to the free-pilot reference policy
at 100 and 200 texts.}\label{tab:reference_rerun_sections}
\adjustbox{max width=\textwidth}{\input{tables/t_reference_rerun_sections}}
\end{table}

\section{Prompt wrappers of the sensitivity study}\label{app:prompts}
The sensitivity runs use the three panel models with the population run's settings (Claude Sonnet~5.5 with
reasoning effort low, Gemini 2.5~Flash with thinking off, GPT-5~mini with reasoning low); Flash was run under all
four variants, Sonnet and mini under P1 and P4, and every model under the no-context variant, the contrasting
contexts and a repeat of P0. The codebook (version~1.2) is inserted verbatim between the \texttt{=== CODEBOOK ===} markers of every wrapper,
the repeat of P0 included; the population run of 4~October had inserted version~1.1 into the same wrapper, so the
repeat differs from it in the codebook version as well as in the run. The
grant row block and the final instruction (``Return the JSON object described in the codebook's Output format and
nothing else'') are identical across wrappers. The original wrapper (P0) is the companion's, whose hash is in the
record. The four variants:
\begin{description}[leftmargin=1.2em,font=\normalfont\bfseries]
\item[P1] ``Classify the stated purpose of the grant row below using the codebook. Base the label only on the
  purpose text as filed; do not use what you know or can guess about the funder or the recipient. `U' (unclear) is
  a legitimate answer whenever the text does not settle the question.''
\item[P2] ``You are an annotator in a measurement study of grant purposes. Your task is to apply a fixed codebook
  to one grant row and report the labels it prescribes. The codebook is authoritative: follow its rules and worked
  cases even where your own judgement would differ. Judge the purpose text only.''
\item[P3] ``Label the grant purpose below. Rules:''
\item[P4] ``Work in three steps. (1) Split the purpose text into its atomic purposes. (2) Apply the codebook's
  rules to each atom. (3) Decide the status of the whole text from the atoms, following the codebook. Use only the
  purpose text; the funder, recipient, year and amount are context, not evidence of purpose. Answer `U' when the
  text does not settle the question.''
\end{description}
The no-context variant uses P0 with the grant row reduced to the purpose text. The contrasting-context runs use P0
with the context of the largest grant carrying the text and with the context of a randomly chosen grant of a
different funder carrying the same text.

%% file: tables/t_superuniformity.tex
\begin{tabular}{lrr}
\toprule
$t$ & $P(p \le t)$, dollar measure & unweighted share \\
\midrule
0.01 & 0.000 & 0.022 \\
0.02 & 0.000 & 0.022 \\
0.05 & 0.026 & 0.077 \\
0.10 & 0.063 & 0.117 \\
0.20 & 0.161 & 0.191 \\
0.30 & 0.260 & 0.251 \\
0.50 & 0.485 & 0.415 \\
\bottomrule
\end{tabular}

%% file: tables/t_rogan.tex
\begin{tabular}{lrrrr}
\toprule
Sensitivity $\backslash$ Specificity & 0.90 & 0.95 & 0.98 & 1.00 \\
\midrule
0.85 & 4.31 & 10.42 & 13.74 & 15.82 \\
0.90 & 4.04 & 9.81 & 12.96 & 14.94 \\
0.95 & 3.80 & 9.27 & 12.26 & 14.16 \\
1.00 & 3.59 & 8.78 & 11.64 & 13.45 \\
\bottomrule
\end{tabular}

%% file: tables/t_drift.tex
\begin{tabular}{lrrrrr}
\toprule
Tax year & Rules & Claude Sonnet 5.5 & Gemini 2.5 Flash & GPT-5 mini & Calibrated panel (OOF) \\
\midrule
2020 & -0.9 & 0.0 & +1.2 & +1.9 & +0.7 \\
2021 & -0.9 & -1.2 & -0.1 & +0.8 & +0.2 \\
2022 & -1.4 & -1.0 & +0.0 & +0.9 & -0.1 \\
2023 & -0.6 & -1.2 & +0.6 & +1.5 & +0.6 \\
2024 & -0.2 & -1.1 & -0.1 & +1.4 & +0.2 \\
\bottomrule
\end{tabular}

%% file: references.bib
@misc{irs_schi,
  author       = {{Internal Revenue Service}},
  title        = {Instructions for Schedule {I} ({Form} 990): Grants and Other Assistance to Organizations, Governments, and Individuals in the United States},
  year         = {2024},
  url          = {https://www.irs.gov/instructions/i990si}
}

@misc{irs_pf,
  author       = {{Internal Revenue Service}},
  title        = {Instructions for Form 990-{PF}: Return of Private Foundation or Section 4947(a)(1) Trust Treated as Private Foundation (2024)},
  year         = {2024},
  url          = {https://www.irs.gov/pub/irs-prior/i990pf--2024.pdf}
}

@misc{ipeds,
  author       = {{National Center for Education Statistics}},
  title        = {Integrated Postsecondary Education Data System: Institutional Characteristics, Directory File {HD2024}},
  year         = {2025},
  url          = {https://nces.ed.gov/ipeds/use-the-data}
}

@techreport{nsf26305,
  author       = {{National Center for Science and Engineering Statistics}},
  title        = {Universities Report 8.1\% Growth in {R\&D} Expenditures in {FY} 2024, Reaching Over \$117 Billion},
  type         = {InfoBrief},
  number       = {NSF 26-305},
  institution  = {U.S. National Science Foundation},
  year         = {2026},
  date         = {2026-02-10},
  url          = {https://ncses.nsf.gov/pubs/nsf26305}
}

@article{shekhtman2024,
  author  = {Shekhtman, Louis M. and Gates, Alexander J. and Barab{\'a}si, Albert-L{\'a}szl{\'o}},
  title   = {Mapping philanthropic support of science},
  journal = {Scientific Reports},
  volume  = {14},
  number  = {1},
  eid     = {9397},
  year    = {2024},
  doi     = {10.1038/s41598-024-58367-2},
  eprint = {2206.10661},
  eprinttype = {arxiv}
}

@article{raowu1988,
  author  = {Rao, J. N. K. and Wu, C. F. J.},
  title   = {Resampling inference with complex survey data},
  journal = {Journal of the American Statistical Association},
  volume  = {83},
  number  = {401},
  pages   = {231--241},
  year    = {1988},
  doi     = {10.1080/01621459.1988.10478591}
}

@article{clopperpearson1934,
  author  = {Clopper, C. J. and Pearson, E. S.},
  title   = {The use of confidence or fiducial limits illustrated in the case of the binomial},
  journal = {Biometrika},
  volume  = {26},
  number  = {4},
  pages   = {404--413},
  year    = {1934},
  doi     = {10.1093/biomet/26.4.404}
}

@online{tan2026source,
  author  = {Tan, Tao},
  title   = {Introducing {SOURCE} ({Searchable} {Open} {University} {Records} of {Charitable} {Expenditures})},
  organization = {American Enterprise Institute, Center for the Future of the American University},
  year    = {2026},
  month   = {6},
  url     = {https://cfau.aei.org/source/},
  urldate = {2026-10-09},
  note    = {Announcement post of 24 June 2026 (https://cfau.aei.org/introducing-source/); technical appendix at https://github.com/aeicfau/source (GPL-3.0); grant and funder counts quoted from the resource page as accessed}
}

@article{angelopoulos2023ppi,
  author  = {Angelopoulos, Anastasios N. and Bates, Stephen and Fannjiang, Clara and Jordan, Michael I. and Zrnic, Tijana},
  title   = {Prediction-powered inference},
  journal = {Science},
  volume  = {382},
  number  = {6671},
  pages   = {669--674},
  year    = {2023},
  doi     = {10.1126/science.adi6000}
}

@article{angelopoulos2023ppipp,
  author  = {Angelopoulos, Anastasios N. and Duchi, John C. and Zrnic, Tijana},
  title   = {{PPI++}: Efficient prediction-powered inference},
  journal = {The Annals of Applied Statistics},
  volume  = {20},
  number  = {3},
  pages   = {2235--2252},
  year    = {2026},
  doi     = {10.1214/26-AOAS2215},
  eprint = {2311.01453},
  eprinttype = {arxiv}
}

@article{zrnic2024cross,
  author  = {Zrnic, Tijana and Cand{\`e}s, Emmanuel J.},
  title   = {Cross-prediction-powered inference},
  journal = {Proceedings of the National Academy of Sciences},
  volume  = {121},
  number  = {15},
  eid     = {e2322083121},
  year    = {2024},
  doi     = {10.1073/pnas.2322083121}
}

@inproceedings{zrnic2024active,
  author    = {Zrnic, Tijana and Cand{\`e}s, Emmanuel J.},
  title     = {Active statistical inference},
  booktitle = {Proceedings of the 41st International Conference on Machine Learning},
  series    = {Proceedings of Machine Learning Research},
  volume    = {235},
  pages     = {62993--63010},
  publisher = {PMLR},
  year      = {2024},
  eprint = {2403.03208},
  eprinttype = {arxiv}
}

@inproceedings{gligoric2025,
  author    = {Gligori{\'c}, Kristina and Zrnic, Tijana and Lee, Cinoo and Cand{\`e}s, Emmanuel J. and Jurafsky, Dan},
  title     = {Can unconfident {LLM} annotations be used for confident conclusions?},
  booktitle = {Proceedings of the 2025 Conference of the Nations of the Americas Chapter of the Association for Computational Linguistics: Human Language Technologies (Volume 1: Long Papers)},
  publisher = {Association for Computational Linguistics},
  pages     = {3514--3533},
  year      = {2025},
  doi       = {10.18653/v1/2025.naacl-long.179},
  eprint = {2408.15204},
  eprinttype = {arxiv}
}

@article{dawidskene1979,
  author  = {Dawid, A. Philip and Skene, Allan M.},
  title   = {Maximum likelihood estimation of observer error-rates using the {EM} algorithm},
  journal = {Journal of the Royal Statistical Society, Series C (Applied Statistics)},
  volume  = {28},
  number  = {1},
  pages   = {20--28},
  year    = {1979},
  doi     = {10.2307/2346806}
}

@inproceedings{hovy2013mace,
  author    = {Hovy, Dirk and Berg-Kirkpatrick, Taylor and Vaswani, Ashish and Hovy, Eduard},
  title     = {Learning whom to trust with {MACE}},
  booktitle = {Proceedings of the 2013 Conference of the North American Chapter of the Association for Computational Linguistics: Human Language Technologies},
  publisher = {Association for Computational Linguistics},
  pages     = {1120--1130},
  year      = {2013}
}

@article{jincandes2023,
  author  = {Jin, Ying and Cand{\`e}s, Emmanuel J.},
  title   = {Selection by prediction with conformal p-values},
  journal = {Journal of Machine Learning Research},
  volume  = {24},
  number  = {244},
  pages   = {1--41},
  year    = {2023}
}

@article{bh1995,
  author  = {Benjamini, Yoav and Hochberg, Yosef},
  title   = {Controlling the false discovery rate: a practical and powerful approach to multiple testing},
  journal = {Journal of the Royal Statistical Society, Series B},
  volume  = {57},
  number  = {1},
  pages   = {289--300},
  year    = {1995},
  doi     = {10.1111/j.2517-6161.1995.tb02031.x}
}

@article{bh1997weighted,
  author  = {Benjamini, Yoav and Hochberg, Yosef},
  title   = {Multiple hypotheses testing with weights},
  journal = {Scandinavian Journal of Statistics},
  volume  = {24},
  number  = {3},
  pages   = {407--418},
  year    = {1997},
  doi     = {10.1111/1467-9469.00072}
}

@inproceedings{tibshirani2019,
  author    = {Tibshirani, Ryan J. and Barber, Rina Foygel and Cand{\`e}s, Emmanuel J. and Ramdas, Aaditya},
  title     = {Conformal prediction under covariate shift},
  booktitle = {Advances in Neural Information Processing Systems 32},
  pages     = {2530--2540},
  year      = {2019}
}

@article{horvitzthompson1952,
  author  = {Horvitz, Daniel G. and Thompson, Donovan J.},
  title   = {A generalization of sampling without replacement from a finite universe},
  journal = {Journal of the American Statistical Association},
  volume  = {47},
  number  = {260},
  pages   = {663--685},
  year    = {1952},
  doi     = {10.1080/01621459.1952.10483446}
}

@article{neyman1934,
  author  = {Neyman, Jerzy},
  title   = {On the two different aspects of the representative method: the method of stratified sampling and the method of purposive selection},
  journal = {Journal of the Royal Statistical Society},
  volume  = {97},
  number  = {4},
  pages   = {558--625},
  year    = {1934},
  doi     = {10.2307/2342192}
}

@book{sarndal1992,
  author    = {S{\"a}rndal, Carl-Erik and Swensson, Bengt and Wretman, Jan},
  title     = {Model Assisted Survey Sampling},
  publisher = {Springer},
  address   = {New York},
  year      = {1992},
  doi       = {10.1007/978-1-4612-4378-6}
}

@article{gilardi2023,
  author  = {Gilardi, Fabrizio and Alizadeh, Meysam and Kubli, Ma{\"e}l},
  title   = {{ChatGPT} outperforms crowd workers for text-annotation tasks},
  journal = {Proceedings of the National Academy of Sciences},
  volume  = {120},
  number  = {30},
  eid     = {e2305016120},
  year    = {2023},
  doi     = {10.1073/pnas.2305016120}
}

@article{ziems2024,
  author  = {Ziems, Caleb and Held, William and Shaikh, Omar and Chen, Jiaao and Zhang, Zhehao and Yang, Diyi},
  title   = {Can large language models transform computational social science?},
  journal = {Computational Linguistics},
  volume  = {50},
  number  = {1},
  pages   = {237--291},
  year    = {2024},
  doi     = {10.1162/coli_a_00502}
}

@article{adamicadar2003,
  author  = {Adamic, Lada A. and Adar, Eytan},
  title   = {Friends and neighbors on the {Web}},
  journal = {Social Networks},
  volume  = {25},
  number  = {3},
  pages   = {211--230},
  year    = {2003},
  doi     = {10.1016/S0378-8733(03)00009-1}
}

@misc{koomullil2026companion,
  author = {Koomullil, George},
  title  = {Who Funds University Research? Grants reported to {US} colleges and universities in 13.9 million {IRS} {Form} 990 grant records, tax years 2020--2024, compared with federal and foreign funding data},
  year   = {2026},
  note   = {Report and data release, version 2026.10. \url{https://ascendr.ai/resources/funds/}}
}

@misc{koomullil2026record,
  author = {Koomullil, George},
  title  = {Counting Under Disagreement: replication record (calibration sample, population aggregates, prompts and analysis code)},
  year   = {2026},
  note   = {Zenodo. \url{https://doi.org/10.5281/zenodo.24430517}}
}

@misc{liu2026interprompt,
  author = {Liu, Jingyuan},
  title  = {What is actually being annotated? {Inter-prompt} reliability as a measurement problem in {LLM}-based social science labeling},
  year   = {2026},
  eprint = {2604.16413},
  eprinttype = {arxiv}
}

@misc{li2026calibrate,
  author = {Li, Yanran},
  title  = {Calibrate, don't curate: label-efficient estimation from noisy {LLM} judges},
  year   = {2026},
  eprint = {2605.09702},
  eprinttype = {arxiv}
}

@misc{huang2025selective,
  author = {Huang, Huipeng and Liao, Wenbo and Xi, Huajun and Zeng, Hao and Zhao, Mengchen and Wei, Hongxin},
  title  = {Model-agnostic selective labeling with provable statistical guarantees},
  year   = {2025},
  eprint = {2510.14581},
  eprinttype = {arxiv}
}

@article{angelopoulos2021ltt,
  author  = {Angelopoulos, Anastasios N. and Bates, Stephen and Cand{\`e}s, Emmanuel J. and Jordan, Michael I. and Lei, Lihua},
  title   = {Learn then test: calibrating predictive algorithms to achieve risk control},
  journal = {The Annals of Applied Statistics},
  volume  = {19},
  number  = {2},
  pages   = {1641--1662},
  year    = {2025},
  doi     = {10.1214/24-AOAS1998},
  eprint = {2110.01052},
  eprinttype = {arxiv}
}

@inproceedings{angelopoulos2024crc,
  author    = {Angelopoulos, Anastasios N. and Bates, Stephen and Fisch, Adam and Lei, Lihua and Schuster, Tal},
  title     = {Conformal risk control},
  booktitle = {The Twelfth International Conference on Learning Representations},
  year      = {2024},
  eprint = {2208.02814},
  eprinttype = {arxiv}
}

@article{bates2023outliers,
  author  = {Bates, Stephen and Cand{\`e}s, Emmanuel J. and Lei, Lihua and Romano, Yaniv and Sesia, Matteo},
  title   = {Testing for outliers with conformal p-values},
  journal = {The Annals of Statistics},
  volume  = {51},
  number  = {1},
  pages   = {149--178},
  year    = {2023},
  doi     = {10.1214/22-AOS2244}
}

@article{jincandes2023weighted,
  author  = {Jin, Ying and Cand{\`e}s, Emmanuel J.},
  title   = {Model-free selective inference under covariate shift via weighted conformal p-values},
  journal = {Biometrika},
  volume  = {113},
  number  = {1},
  eid     = {asaf066},
  year    = {2026},
  doi     = {10.1093/biomet/asaf066},
  eprint = {2307.09291},
  eprinttype = {arxiv}
}

@inproceedings{chen2026judge,
  author    = {Chen, Yiqun T. and Lu, Sizhu and Li, Sijia and Guo, Moran and Li, Shengyi},
  title     = {Efficient inference for noisy {LLM}-as-a-judge evaluation},
  booktitle = {Proceedings of the 43rd International Conference on Machine Learning},
  series    = {Proceedings of Machine Learning Research},
  volume    = {306},
  pages     = {15836--15860},
  publisher = {PMLR},
  year      = {2026},
  eprint = {2601.05420},
  eprinttype = {arxiv}
}

@inproceedings{egami2023dsl,
  author    = {Egami, Naoki and Hinck, Musashi and Stewart, Brandon M. and Wei, Hanying},
  title     = {Using imperfect surrogates for downstream inference: design-based supervised learning for social science applications of large language models},
  booktitle = {Advances in Neural Information Processing Systems 36},
  pages     = {68589--68601},
  year      = {2023},
  doi       = {10.52202/075280-3000},
  eprint = {2306.04746},
  eprinttype = {arxiv}
}

@article{prasadrao1990,
  author  = {Prasad, N. G. N. and Rao, J. N. K.},
  title   = {The estimation of the mean squared error of small-area estimators},
  journal = {Journal of the American Statistical Association},
  volume  = {85},
  number  = {409},
  pages   = {163--171},
  year    = {1990},
  doi     = {10.1080/01621459.1990.10475320}
}

@misc{egamishin2026dmm,
  author = {Egami, Naoki and Shin, Sooahn},
  title  = {Debiased inference for {AI}-generated data without gold-standard labels: identification via multiple imperfect measurements},
  year   = {2026},
  eprint = {2608.18294},
  eprinttype = {arxiv}
}

@inproceedings{fisch2024stratppi,
  author    = {Fisch, Adam and Maynez, Joshua and Hofer, R. Alex and Dhingra, Bhuwan and Globerson, Amir and Cohen, William W.},
  title     = {Stratified prediction-powered inference for effective hybrid evaluation of language models},
  booktitle = {Advances in Neural Information Processing Systems 37},
  pages     = {111489--111514},
  year      = {2024},
  doi       = {10.52202/079017-3541},
  eprint = {2406.04291},
  eprinttype = {arxiv}
}

@article{genovese2006weighted,
  author  = {Genovese, Christopher R. and Roeder, Kathryn and Wasserman, Larry},
  title   = {False discovery control with p-value weighting},
  journal = {Biometrika},
  volume  = {93},
  number  = {3},
  pages   = {509--524},
  year    = {2006},
  doi     = {10.1093/biomet/93.3.509}
}

@article{miao2023popinf,
  author  = {Miao, Jiacheng and Miao, Xinran and Wu, Yixuan and Zhao, Jiwei and Lu, Qiongshi},
  title   = {Assumption-lean and data-adaptive post-prediction inference},
  journal = {Journal of Machine Learning Research},
  volume  = {26},
  number  = {179},
  pages   = {1--31},
  year    = {2025},
  eprint = {2311.14220},
  eprinttype = {arxiv}
}

@article{paun2018annotation,
  author  = {Paun, Silviu and Carpenter, Bob and Chamberlain, Jon and Hovy, Dirk and Kruschwitz, Udo and Poesio, Massimo},
  title   = {Comparing {Bayesian} models of annotation},
  journal = {Transactions of the Association for Computational Linguistics},
  volume  = {6},
  pages   = {571--585},
  year    = {2018},
  doi     = {10.1162/tacl_a_00040}
}

@article{rogangladen1978,
  author  = {Rogan, Walter J. and Gladen, Beth},
  title   = {Estimating prevalence from the results of a screening test},
  journal = {American Journal of Epidemiology},
  volume  = {107},
  number  = {1},
  pages   = {71--76},
  year    = {1978},
  doi     = {10.1093/oxfordjournals.aje.a112510}
}

@misc{verga2024juries,
  author = {Verga, Pat and Hofstatter, Sebastian and Althammer, Sophia and Su, Yixuan and Piktus, Aleksandra and Arkhangorodsky, Arkady and Xu, Minjie and White, Naomi and Lewis, Patrick},
  title  = {Replacing judges with juries: evaluating {LLM} generations with a panel of diverse models},
  year   = {2024},
  eprint = {2404.18796},
  eprinttype = {arxiv}
}

@article{wang2020postpred,
  author  = {Wang, Siruo and McCormick, Tyler H. and Leek, Jeffrey T.},
  title   = {Methods for correcting inference based on outcomes predicted by machine learning},
  journal = {Proceedings of the National Academy of Sciences},
  volume  = {117},
  number  = {48},
  pages   = {30266--30275},
  year    = {2020},
  doi     = {10.1073/pnas.2001238117}
}

@inproceedings{whitehill2009glad,
  author    = {Whitehill, Jacob and Ruvolo, Paul and Wu, Tingfan and Bergsma, Jacob and Movellan, Javier},
  title     = {Whose vote should count more: optimal integration of labels from labelers of unknown expertise},
  booktitle = {Advances in Neural Information Processing Systems 22},
  pages     = {2035--2043},
  year      = {2009}
}

@article{wolpert1992stacking,
  author  = {Wolpert, David H.},
  title   = {Stacked generalization},
  journal = {Neural Networks},
  volume  = {5},
  number  = {2},
  pages   = {241--259},
  year    = {1992},
  doi     = {10.1016/S0893-6080(05)80023-1}
}

@article{denishaddou2026ppe,
  author  = {Denis, Nicholas and Haddou, Mohammed},
  title   = {Prediction-powered estimation: unbiased model-assisted estimation},
  journal = {Journal of Official Statistics},
  volume  = {42},
  number  = {3},
  pages   = {371--394},
  year    = {2026},
  doi     = {10.1177/0282423X261451320}
}

@misc{mozer2026difference,
  author = {Mozer, Reagan},
  title  = {{PPI} is the difference estimator: recognizing the survey sampling roots of prediction-powered inference},
  year   = {2026},
  eprint = {2603.19160},
  eprinttype = {arxiv}
}

@misc{klugerbates2026multiwave,
  author = {Kluger, Dan M. and Bates, Stephen},
  title  = {M-estimation under two-phase multiwave sampling with applications to prediction-powered inference},
  year   = {2026},
  eprint = {2602.16933},
  eprinttype = {arxiv}
}

@misc{yeyoga2026allocation,
  author = {Ye, Zikun and Yoganarasimhan, Hema},
  title  = {Rectification difficulty and optimal sample allocation in {LLM}-augmented surveys},
  year   = {2026},
  eprint = {2604.17267},
  eprinttype = {arxiv}
}

@techreport{nces1999bootstrap,
  author      = {Kaufman, Steven},
  title       = {A bootstrap variance estimator for systematic {PPS} sampling},
  institution = {National Center for Education Statistics, U.S. Department of Education},
  address     = {Washington, DC},
  type        = {Working Paper},
  number      = {NCES 98-12},
  year        = {1998},
  note        = {Dated October 1998; released by NCES 31 August 1999. ERIC ED434133},
  url         = {https://nces.ed.gov/pubsearch/pubsinfo.asp?pubid=9812}
}

@book{kish1965,
  author    = {Kish, Leslie},
  title     = {Survey Sampling},
  publisher = {John Wiley \& Sons},
  address   = {New York},
  year      = {1965}
}
